\documentclass{IEEEtran}
\pdfoutput=1
\usepackage{amsmath}
\usepackage{float}
\usepackage{subcaption}
\usepackage[ruled]{algorithm2e}
\usepackage{verbatim,multirow}
\usepackage{hyperref}
\usepackage[export]{adjustbox}
\usepackage{array}
\newcolumntype{H}{>{\setbox0=\hbox\bgroup}c<{\egroup}@{}}
\usepackage{graphics}
\usepackage{mathtools}  
\usepackage{makecell}   

\usepackage{color, colortbl}
\usepackage{standalone}
\usepackage{pdfpages}
\usepackage{titling}

\definecolor{LightCyan}{rgb}{0.88,1,1}
\usepackage[table,x11names]{xcolor}
\newcommand{\rebuttal}[1]{\textcolor{black}{#1}}
\newcommand{\rebuttalii}[1]{\textcolor{black}{#1}}
\newcommand{\rebuttaliitable}{\color{black}}

\usepackage{amsthm}
\newcommand{\beginsupplement}{%
        \setcounter{table}{0}
        \renewcommand{\thetable}{S\arabic{table}}%
        \setcounter{figure}{0}
        \renewcommand{\thefigure}{S\arabic{figure}}%
        \setcounter{section}{0}
        \renewcommand{\thesection}{S\arabic{section}}
     }
 
\theoremstyle{definition}
\newtheorem{definition}{Definition}[section]

\title{Interpretable Rule Discovery Through Bilevel Optimization of Split-Rules of Nonlinear Decision Trees for Classification Problems}
\author{Yashesh~Dhebar\thanks{Yashesh Dhebar is with the Department of Mechanical Engineering at Michigan State University, East Lansing, MI 48824, USA, e-mail: dhebarya@egr.msu.edu},~\IEEEmembership{Member,~IEEE} ~and Kalyanmoy~Deb
\thanks{Kalyanmoy Deb is with the Department
of Electrical and Computer Engineering, Michigan State University, East Lansing,
MI, 48824, USA, e-mail: kdeb@egr.msu.edu,
~COIN Lab website: \url{https://coin-lab.org}},~\IEEEmembership{Fellow,~IEEE}}

\begin{document}
\date{}
\maketitle
\begin{abstract}
For supervised classification problems involving design, control, other practical purposes, users are not only interested in finding a highly accurate classifier, but they also demand that the obtained classifier be easily interpretable. While the definition of interpretability of a classifier can vary from case to case, here, by a humanly interpretable classifier we restrict it to be expressed in simplistic mathematical terms.
As a novel approach, we represent a classifier as an assembly of simple mathematical rules using a non-linear decision tree (NLDT). Each conditional (non-terminal) node  of the tree represents a non-linear mathematical rule (split-rule) involving features in order to partition the dataset in the given conditional node into two non-overlapping subsets. This partitioning is intended to minimize the impurity of the resulting child nodes. By restricting the structure of split-rule at each conditional node and depth of the decision tree, the interpretability of the classifier is assured. 
The non-linear split-rule at a given conditional node is obtained using an evolutionary bilevel optimization algorithm, in which while the upper-level focuses on arriving at an interpretable structure of the split-rule, the lower-level achieves the most appropriate weights \rebuttalii{(coefficients)} of individual constituents of the rule to minimize the net impurity of two resulting child nodes. 
The performance of the proposed algorithm is demonstrated on a number of controlled test problems, existing benchmark problems, and industrial problems. Results \rebuttalii{on two to 500-feature problems} are encouraging and open up further scopes of applying the proposed approach to more challenging and complex classification tasks.
\end{abstract}


\IEEEpeerreviewmaketitle

\section{Introduction}
In a classification task, usually a set of labelled data involving a set of features and its belonging to a specific class are provided. The task in a classification \rebuttalii{task} is to design an algorithm which works on the given dataset and arrives at one or more classification rules (expressed as a function of problem features) which are able to make predictions with maximum classification accuracy. A classifier can be expressed in many different ways suitable for the purpose of the application. \rebuttalii{In a quest to make the classifier interpretable, classifier representation through a decision-list or a decision-tree (DT) had been a popular choice.}
In decision trees, the classification rule-set is represented in an inverted tree based structure where  non-terminal conditional nodes comprise of conditional statements, \rebuttalii{mostly involving a single feature,} and the terminal leaf-nodes have a class label associated with them. A datapoint traverses through the DT by following the rules at  conditional nodes and lands at a particular leaf-node, \rebuttalii{marking its class}. \rebuttalii{In the artificial neural network (ANN) approach, an implicit relationship among features is captured through a multi-layered network, while in the support-vector-machine (SVM) approach, a complex mathematical relationship of associated support vectors is derived. One common aspect of these methods is that they use an optimization method of arriving at a suitable DT, ANN or SVM based on maximizing the classification accuracy.}


Due to their importance in many practical problems involving design, control, identification, and other machine learning related tasks, researchers have spent a lot of \rebuttalii{efforts} to develop efficient optimization-based classification algorithms \cite{kotsiantis2007supervised, bergstra2011algorithms, thornton2013auto}. While most algorithms are developed for classifying two-class data sets, the developed methods can be extended to multi-class classification problems as hierarchical two-class classification problems or by extending them to constitute a simultaneous multi-class classification algorithm. In this paper, we do not \rebuttalii{extend our approach to multi-class classification problems, except showing a proof-of-principle study.} 

In most classification problem solving tasks, maximizing classification accuracy (or minimizing classification error) on the labelled dataset is usually used as the sole objective. 
However, besides the classification accuracy, in many applications, users are also interested in finding an easily interpretable classifier for various practical reasons: (i) it helps to identify the most important rules which are responsible for the classification task, (ii) it also helps to provide a more direct relationship of features to have a better insight for the underlying classification task for knowledge enhancement and future developmental purposes. The definition of an easily interpretable classifier largely depends on the context, but the existing literature represents them as a linear, polynomial, or {\em posynomial\/} function involving only a few features. 

In this paper, we propose a number of novel ideas \rebuttalii{towards evolving a highly accurate and easily interpretable classifier.} First, instead of a DT, we propose a nonlinear decision tree (NLDT) as a classifier, in which every non-terminal conditional node will represent a nonlinear function of features to express a split-rule. Each split-rule will split the data into two non-overlapping subsets. Successive hierarchical splits of these sub-sets are carried out and the tree is allowed to grow until one of the termination criteria is met. We argue that flexibility of allowing nonlinear split-rule at conditional nodes (instead of a single-variable based rule which is found in tradition ID3 based DTs \cite{breiman2017classification}) will result in a more compact DT (having a fewer nodes). \emph{Second}, to derive the split-rule at a given conditional-node, a dedicated bilevel-optimization algorithm is applied, \rebuttalii{so that learning of the structure of the rule and associated coefficients are treated as two hierarchical and inter-related optimization tasks}. 
\emph{Third}, our proposed methodology is customized for  classification problem solving, \rebuttalii{so that our overall bilevel optimization algorithm is computationally efficient}. \emph{Fourth}, we emphasize \rebuttalii{evolving} simplistic rule structures \rebuttalii{through our customized} bilevel optimization method so that obtained rules are also interpretable. 

In the remainder of the paper, we provide a brief survey of the existing approaches of inducing DTs in Section~\ref{sec:past}. A detailed description about the problem of inducing interpretable DTs from optimization perspective and a high-level view on proposed approach is provided in  Section~\ref{sec:proposed_approach}. Next, we provide an in-depth discussion of the \rebuttalii{proposed} bilevel-optimization algorithm which is adopted to derive interpretable split-rules at each conditional node of NLDT in Section~\ref{sec:bilevel_split_rule}. 
Section~\ref{sec:pruning} provides a brief overview on the post-processing method which is used to prune the tree to simplify its topology.
Compilation of results on standard classification and engineering problems are given in section~\ref{sec:results_main}.
The paper ends with concluding remarks and some highlights on future work in Section~\ref{sec:conclusion}.

\section{Past Studies}
\label{sec:past}
There exist many studies involving machine learning and data analytics methods for discovering rules from data. Here, we provide a brief mention of some studies which are close to our proposed study. 

Traditional induction of DTs is done in axis-parallel fashion \cite{breiman2017classification}, wherein each split rule is of type $x_i \le \tau_i$ or $x_i \ge \tau_i$. Algorithms such as ID3, CART and C4.5 are among the popular ones in this category.
The work done in the past to generate \emph{non-axis-parallel} trees can be found in \cite{heath1993induction, murthy1994system, murthy1993oc1}, where the researchers rely on  randomized algorithms to search for multi-variate hyperplanes. Work done in \cite{cantu2000using, kretowski2004evolutionary} 
use evolutionary algorithms to induce oblique DTs. The idea of OCI \cite{murthy1993oc1} is extended in \cite{ittner1996non} to generate nonlinear quadratic split-rules in a DT. Bennett Et al. \cite{bennett1998support} uses SVM to generate linear/nonlinear split rules. 
However, these works do not address certain key practical considerations, such as the \emph{complexity} of the combined split-rules and handling of biased data. 

\rebuttalii{Next, we discuss some studies} which take the aspect of complexity of rule into consideration.
In \cite{nunez2002rule}, ellipsoidal and interval based rules are determined using the set of support-vectors and prototype points/vertex points. The authors there primarily focus at coming up with compact set of if-then-else rules which are comprehensible. Despite its intuitiveness, the approach proposed in \cite{nunez2002rule} doesn't result into comprehensible set of rules on high-dimensional datasets. Another approach suggested in \cite{fung2005rule} uses the decompositional based technique of deriving rules using the output of a linear-SVM classifier. The rules here are expressed as hypercubes. The method proposed is computationally fast, but it lacks in its scope  to address nonlinear decision boundaries and its performance is limited by the performance of a regular linear-SVM. On the other hand, this approach has a tendency to generate more rules if the data is distributed parallel to the decision boundary. 
The study conducted in \cite{craven1996extracting} uses a trained neural-network as an oracle to develop a DT of \emph{at least m-of-n} type rule-sets (similar to  the one described in ID2-of-3 \cite{murphy1991id2}). The strength of this approach lies in its potential to scale up. 
However, its accuracy on unseen dataset usually falls by about $3\%$ from the corresponding oracle neural-network counterpart. Authors in \cite{johansson2004truth} use a pedagogical technique to evolve comprehensible rules using genetic-programming. The proposed algorithm G-REX considers the \emph{fidelity} (i.e. how closely can the evolved AI agent mimic the behaviour of the oracle NN) as the primary objective and the compactness of the rule expression is penalized using a parameter to evolve interpretable set of \emph{fuzzy rules} for a classification problem. The approach is good enough to produce comprehensible rule-set, but it needs tweaking and fine tuning of the penalty parameter. A nice summary of the above-mentioned methods is provided in \cite{martens2007comprehensible}. Ishibuchi et al. in \cite{ishibuchi2001three} implemented a three-objective strategy to evolve fuzzy set of rules using a multi-objective genetic algorithm. The objectives considered in that research were the classification accuracy, number of rules in the evolved rule-set and the total number of antecedent conditions. In \cite{jin2008pareto}, an  \rebuttalii{ANN} is used as a final classifier and a multi-objective optimization approach is employed to find simple and accurate classifier. The simplicity is expressed by the number of nodes. Hand calculations are then used to analytically express the \rebuttalii{ANN} with a mathematical function. This procedure however becomes intractable when the evolved \rebuttalii{ANN} has a large number of nodes.
 
Genetic programming (GP) based methods have been found efficient in deriving relevant features from the set of original features which are then used to generate the classifier \rebuttalii{or interpretable rule sets \cite{nguyen2016surrogate, nag2015multiobjective, luna2014use, lensen2020genetic, guo2006breast, muharram2005evolutionary}}. In some studies, the entire classifier is encoded with a genetic-representation and the genome is then evolved using GP. Some works conducted in this regard also simultaneously consider complexity of the classifier \cite{shirasaka1998automatic, haruyama2002designing, kuo2007applying, bot2000improving}, but those have been limited to axis-parallel or oblique splits. Application of GP to induce nonlinear multivariate decision tree can be found in \cite{marmelstein1998pattern, mugambi2003polynomial}. Our approach of inducing nonlinear decision tree is conceptually similar to the idea discussed in  \cite{marmelstein1998pattern}, where the DT is induced in the top-down way and at each internal conditional node, the nonlinear split-rule is derived using a GP. Here, the fitness of a GP solution is expressed as a weighted-sum of misclassification-rates and generalizability term. However, the \emph{interpretability} aspect of the final split-rule does not get captured anywhere in the fitness assignment and authors do not report the complexity of the finally obtained classifiers. \rebuttalii{Authors in \cite{cano2013interpretable} use a three-phase evolutionary algorithm to create a composite rule set of axis parallel rules for each class. Their algorithm is able to generate interpretable rules but struggles to achieve respectable classification accuracy on datasets which are not separable by axis-parallel boundaries. In \cite{bot2000application}, GP is used to evolve DTs with linear split-rules, however the work does not focus on improving the interpretability of the evolved DT classifier. Eggermont et al. \cite{eggermont2004genetic} induce the DT using GP by allowing multiple splits through their clustering and refined atomic (internal nodes) representations and conduct a global evolutionary search to estimate the values of attribute thresholds. A multi-layered fitness approach is adopted with first priority given to the overall classification accuracy followed by the  complexity of DT. Here too, splits occur based on the box conditions and hence it becomes difficult to address classification datasets involving non-linear (or non axis-parallel) decision boundaries. In \cite{tan2002mining}, a decision list is generated using a GP by incorporating concept-mapping and covering techniques. Similar to \cite{cano2013interpretable}, a dedicated rule-set for each class is generated and this method too is limited to work well on datasets which can be partitioned through boxed boundaries.
An interesting application of GP to derive interpretable classification rules is reported in \cite{de2002discovering}. Here, a dedicated rule of type \emph{IF-ANTECEDENT-THEN-DECISION} is generated for each class in the dataset using a GP and a procedure is suggested to handle intermediate cases. The suggested algorithm shows promising results for classification accuracy but the study does not report statistics regarding the complexity of the evolved ruleset.}

{In our proposed approach, we attempt to evolve nonlinear DTs which are robust to different biases in the dataset and simultaneously target in evolving nonlinear yet simpler polynomial split-rules at each conditional node of NLDT with a dedicated bilevel genetic algorithm (shown pictorially in Figure~\ref{fig:sample_DT}). In oppose to the method discussed in \cite{marmelstein1998pattern}, where the fitness of GP individual doesn't capture the notion of complexity of rule expression, the bilevel-optimization proposed in our work deals with the aspects of \emph{interpretability} and \emph{performance} of split-rule in a logically hierarchical manner (conceptually illustrated in Figure~\ref{fig:bilevel_clip_art}). The results indicate that the proposed bilevel algorithm for evolving nonlinear split-rules eventually generates simple and comprehensible classifiers with high or comparable accuracy on all the problems used in this study.}

\section{Proposed Approach}
\label{sec:proposed_approach}

\subsection{Classifier Representation Using Nonlinear Decision Tree}
\label{sec:NLDT}

{
An interpretable classifier for our proposed method is represented in the form of a nonlinear decision tree (NLDT), as depicted in Figure~\ref{fig:sample_DT}. 
\begin{figure}[hbt]
    \centering
    \includegraphics[width = \linewidth]{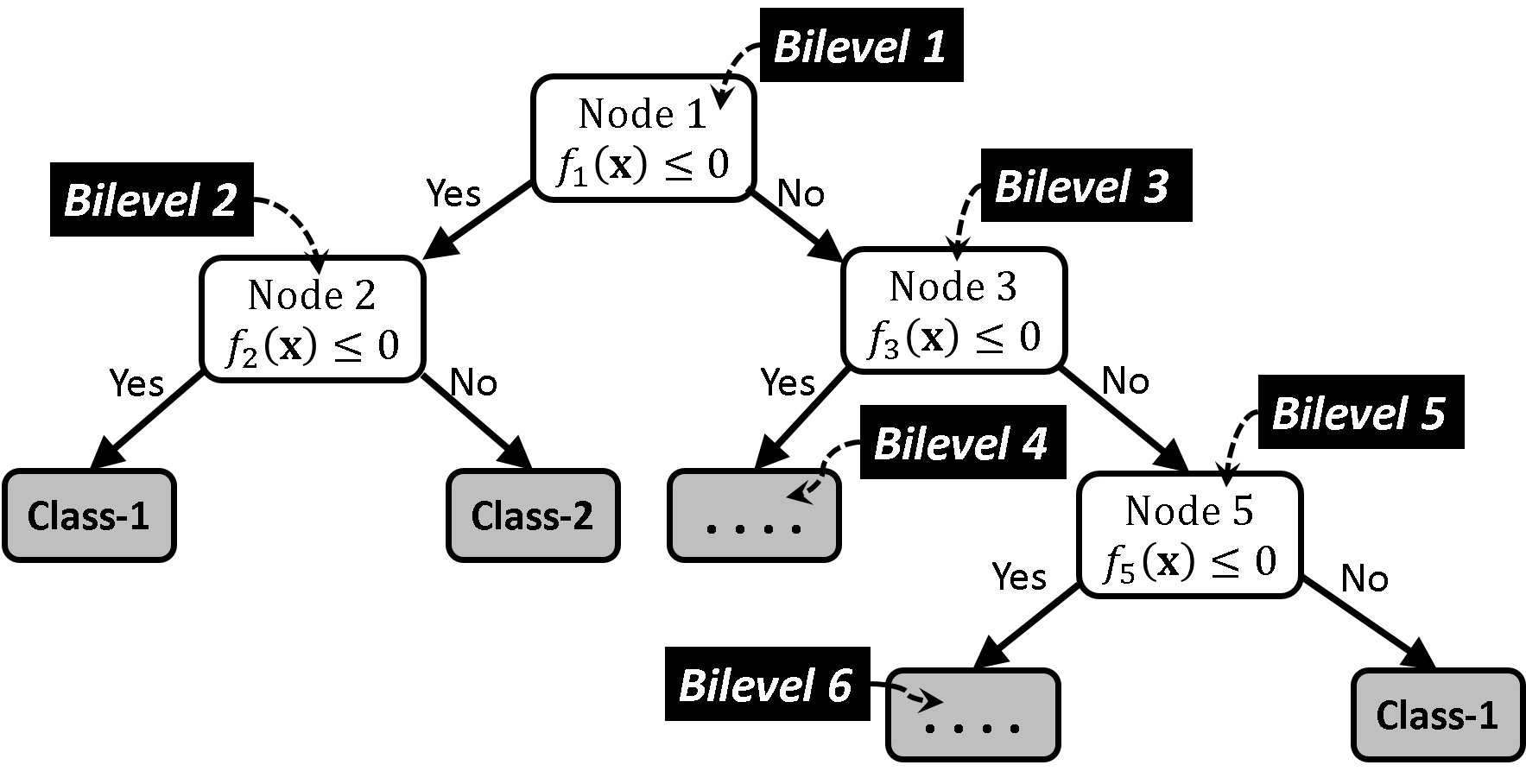}
    \caption{\rebuttalii{An illustration of a Nonlinear Decision Tree (NLDT) classifier for a two class problem. For a given conditional node $i$, the split-rule function $f_i(\mathbf{x})$ is evolved using a dedicated bilevel-optimization procedure.}}
    \label{fig:sample_DT}
\end{figure}
The decision tree (DT) comprises of conditional (or non-terminal) nodes and terminal leaf-nodes. Each conditional-node of the DT has a rule ($f_i(\mathbf{x})\leq 0$) associated to it, where
$\mathbf{x}$ represents the feature vector. 
To make the DT more interpretable, two aspects are considered:}
{
\begin{enumerate}
    \item Simplicity of split-rule $f_i(\mathbf{x})$ at each conditional nodes (see Figure~\ref{fig:bilevel_clip_art}) and
    \item Simplicity of the topology of overall DT, which is computed by  counting total number of conditional-nodes in the DT.
\end{enumerate}
}

Under an ideal scenario, the \emph{simplest split-rule} will involve just one attribute (or feature) and can be expressed as \rebuttalii{$f_i(\mathbf{x}): x_k - \tau \le 0$, for $i$-th rule.} Here, the split occurs \rebuttalii{solely based on the $k^{th}$ feature ($x_k$). In most complex problems, such a simple split rule involving a single feature may not lead to a small-depth DT, thereby making the overall classifier un-interpretable.} 
DTs induced using algorithms such as ID.3 and C4.5 fall under this category. In the present work, we refer to these trees \rebuttalii{with classification and regression trees (CART)} \cite{breiman2017classification}.

On the other extreme, a \emph{topologically-simplest} tree will correspond to the DT involving only one conditional node, but the associated rule \rebuttalii{may be too complex} to interpret. SVM based classifiers fall in this category, wherein decision-boundary is expressed in form of a complicated mathematical equation. Another way to represent a classifier is through an \rebuttalii{ANN}, which when attempted to express analytically, will resort into a complicated nonlinear function $f(\mathbf{x})$ without any easy interpretability.

In this work, we propose a novel and compromise approach to above two extreme cases so that the resulting DT is not too deep and associated split-rule function at each conditional node $f_i(\mathbf{x})$ assume a nonlinear form with controlled complexity and is easily interpretable. \rebuttalii{The development of a nonlinear DT is similar in principle to the search of a linear DT in CART, but this task requires the use of an efficient nonlinear optimization method for evolving nonlinear rules ($f_i(\mathbf{x})$), for which we exploit recent advances in bilevel optimization.} 
The \rebuttalii{proposed} bilevel approach is discussed in Section~\ref{sec:bilevel_split_rule} \rebuttalii{and a} high-level perspective of the optimization process is provided in Figure~\ref{fig:sample_DT}. During the training phase, NLDT is induced using a recursive algorithm as shown in Algorithm~\ref{algo:NLDT_Induction}. Brief description about subroutines used in Algorithm~\ref{algo:NLDT_Induction} is provided below and relevant pseudo codes for \emph{ObtainSplitRule} subroutine and an \emph{evaluator} for upper-level GA are provided in Algorithm~\ref{algo:obtain_split_rule} and \ref{algo:upper_level_eval} respectively. For brevity, details of the termination criteria is provided in the Supplementary Section.
\begin{itemize}
    \item \emph{ObtainSplitRule(Data)}
    \begin{itemize}
        \item {\bf Input}: $\left(N\times(D+1) \right)$-matrix representing the dataset,  where the last column indicates the class-label.
        \item {\bf Output}: Non linear-split rule $f(\mathbf{x}) \le 0$.
        \item {\bf Method}: Bilevel optimization is used to determine $f(\mathbf{x})$. Details regarding this are discussed in Sec \ref{sec:bilevel_split_rule}.
    \end{itemize}
    \item \emph{SplitNode(Data, SplitRule)}
    \begin{itemize}
        \item {\bf Input}: Data, Split Rule $f(\mathbf{x}) \le 0$.
        \item{\bf Output}:  LeftNode and RightNode which are node-data-structures, where LeftNode.Data are the datapoints in input Data satisfying $f(\mathbf{x})\le 0$ while the RightNode.Data are the datapoints in the input Data violating the split rule.
    \end{itemize}
\end{itemize}

\begin{algorithm}[hbt]
\SetAlgoLined
\KwIn{Dataset}
\SetKwProg{Func}{Function}{:}{}
\Func{UpdatedNode = InduceNLDT(Node, depth)}
{
Node.depth = depth\;
 \eIf {TerminationSatisfied(Node)}{
 Node.Type = `leaf'\;
 }
 {
Node.Type = `conditional'\;
Node.SplitRule = \emph{ObtainSplitRule}(Node.Data)\;

[LeftNode, RightNode] = \emph{SplitNode}(Node.Data, Node.SplitRule)\;

Node.LeftNode = \emph{InduceNLDT}(LeftNode, depth + 1)\;
Node.RightNode = \emph{InduceNLDT}(RightNode, depth + 1)\;
 }
UpdatedNode = Node\;
}
\tcp{NLDT Induction Algorithm}
RootNode.Data = Dataset\;
Tree = InduceNLDT(RootNode, 0)
\caption{Pseudo Code to Recursively Induce NLDT.}
\label{algo:NLDT_Induction}
\end{algorithm}

\begin{algorithm}[hbt]
\caption{\emph{ObtainSplitRule} subroutine. Implements a \emph{Bilevel} optimization algorithm of determine a split-rule. The UpperLevel searches the space of Block Matrix $\mathbf{B}$ and modulus-flag $m$. Constraint violation value for an upper level individuals comes by executing lower-level GA (LLGA) as shown in Algorithm~\ref{algo:upper_level_eval}.}
\label{algo:obtain_split_rule}
\KwIn{Data}
\SetKwInOut{Output}{Output}
\SetKwInOut{Return}{return}
\SetKwInOut{Initialize}{Initialize}
\SetKwFunction{Upper}{UpperLevel}
\SetKwFunction{Lower}{LowerLevel}
\SetKwFunction{Selection}{Selection}
\SetKwFunction{Crossover}{Crossover}
\SetKwFunction{Mutation}{Mutation}
\SetKwFunction{MergeAndSelect}{MergeAndSelect}
\SetKwFunction{ObtainBestInd}{ObtainBestInd}
\SetKwFunction{MaxActiveTerms}{MaxActiveTerms}
\Output{SplitRule \tcp{Split rule $f(\mathbf{x})$}}
\SetKwProg{Fn}{Function}{:}{}
\Fn{SplitRule = ObtainSplitRule(Data)}
{
\Initialize{$P_U$ \tcp{Upper Level population}}
\tcp{Execute LLGA (Algorithm~\ref{algo:upper_level_eval})}
$P_U$ = \emph{EvaluateUpperLevelPop}($P_U$)\;
    
\tcp{Upper Level GA Loop}
\For{gen = 1:MaxGen}
{
$P_U^{Parent}$ = \emph{Selection}($P_U$)\;
$P_U^{Child}$ = \emph{Crossover}($P_U^{Parent}$)\;
$P_U$ = \emph{Mutation}($P_U^{Child}$)\;
\tcp{Execute LLGA (Algorithm~\ref{algo:upper_level_eval})}
$P_U$ = \emph{EvaluateUpperLevelPop}($P_U$)\;
\tcp{Elite preservation}
$P_U$ = \emph{SelectElite}($P_U^{Parent}$, $P_U$)\;
\If{TerminationConditionSatisfied}
{break\;}
}
\tcp{Extract best solution in $P_U$}
SplitRule = \emph{ObtainBestInd}($P_U$)\;
}
\end{algorithm}

\begin{algorithm}[hbt]
\caption{\emph{EvaluateUpperLevelPop} subroutine. A dedicated \emph{LowerLevel} optimization is executed for each upper level population member.}
\label{algo:upper_level_eval}
\SetKwFunction{Lower}{LowerLevel}
\SetKwFunction{MaxActiveTerms}{MaxActiveTerms}
\KwIn{$P_U$ \tcp{Upper Level population}}
\SetKwInOut{Output}{Output}
\SetKwInOut{Return}{return}
\Output{$P_U'$ \tcp{Evaluated Population}}
\SetKwProg{Fn}{Function}{:}{}
\Fn{$P_U'$ = EvaluateUpperLevelPop($P_U$)}
{
    \For{i = 1:PopSize}
    {
    \tcp{Execute LLGA (see Section~\ref{sec:lower_level_ga})}
        [$F_L$, $\mathbf{w}$, $\mathbf{\Theta}$] = \emph{LLGA}(Data, $\mathbf{B}$, $m$)\;
        [$P_U$[i].$\mathbf{w}$, $P_U$[i].$\mathbf{\Theta}$] = [$\mathbf{w}$, $\mathbf{\Theta}$]\;
        \tcp{Constraint and Fitness Value}
        $P_U$[i].CV = $F_L - \tau_I$\;
        $P_U$[i].$F_U$ = \emph{MaxActiveTerms}($\mathbf{B}$)\;
    }
    $P_U' = P_U$
}   

\end{algorithm}

\subsection{\rebuttalii{Split-Rule Representation}}
\label{sec:split_rule_representation}
In this paper, we restrict the expression of \emph{split-rule} at a conditional node of the decision tree operating on the feature vector $\mathbf{x}$ to assume the following structure:
\begin{align}
\text{Rule}:  & \quad f(\mathbf{x},\mathbf{w},\mathbf{\Theta}, \mathbf{B}) \le 0, 
\label{eq:deci_boundary_our}
\end{align}
where $f(\mathbf{x},\mathbf{w},\mathbf{\Theta}, \mathbf{B})$ can be expressed in two different forms depending on whether a {\tt modulus} operator $m$ is sought or not:
\begin{align}
\small
f(\mathbf{x},\mathbf{w},\mathbf{\Theta}, \mathbf{B})= 
\begin{cases}
\theta_1 + w_1B_1 + \ldots + w_pB_p, & \text{if $m = 0$,}\\
    \left| \theta_1 + w_1B_1+ \ldots + w_pB_p\right| - \left|\theta_2\right|, & \text{if $m = 1$.}
\end{cases}
\label{eq:split_rule}
\end{align}
Here, $w_i$'s are coefficients or weights of several power-laws ($B_i$'s), $\theta_i$'s are biases, $m$ is the \emph{modulus-flag} which indicates the presence or absence of the \emph{modulus} operator, $p$ is a user-specified parameter to indicate the maximum number of \emph{power-laws} ($B_i$) which can exist in the expression of $f(\mathbf{x})$, and $B_i$ represents a power-law rule of type:
\begin{equation}
    B_i = x_1^{b_{i1}} \times x_2^{b_{i2}} \times \ldots \times x_d^{b_{id}},
    \label{eq:power_rule}
\end{equation}
$\mathbf{B}$ is a block-matrix of exponents $b_{ij}$, given as follows:
\begin{equation}
    \mathbf{B} = 
    \begin{bmatrix}
    b_{11}       & b_{12} & b_{13} & \dots & b_{1d} \\
    b_{21}       & b_{22} & b_{23} & \dots & b_{2d} \\
     \vdots & \vdots & \vdots & \ddots & \vdots \\
    b_{p1}       & b_{p2} & b_{p3} & \dots & b_{pd}
    \end{bmatrix}.
    \label{eq:B_matrix}
\end{equation}
Exponents $b_{ij}$'s are allowed to assume values from a specified discrete set $\mathbf{E}$. In this work, we set $p = 3$ and $\mathbf{E} = \{-3,-2,-1,0,1,2,3\}$ to make the rules interpretable, however they can be changed by the user. The parameters $w_i$ and $\theta_i$ are real-valued variables in $[-1,1]$. The feature vector $\mathbf{x}$ is a data-point in a $d$-dimensional space. Another user-defined parameter $a_{\max}$ controls the maximum number of variables that can be present in each power-law $B_i$. The default is $a_{\max} = d$ (i.e. dimension of the feature space).

\section{Bilevel Approach for Split-Rule Identification}
\label{sec:bilevel_split_rule}
\begin{figure}[hbt]
    \centering
    \includegraphics[width = 0.8\linewidth]{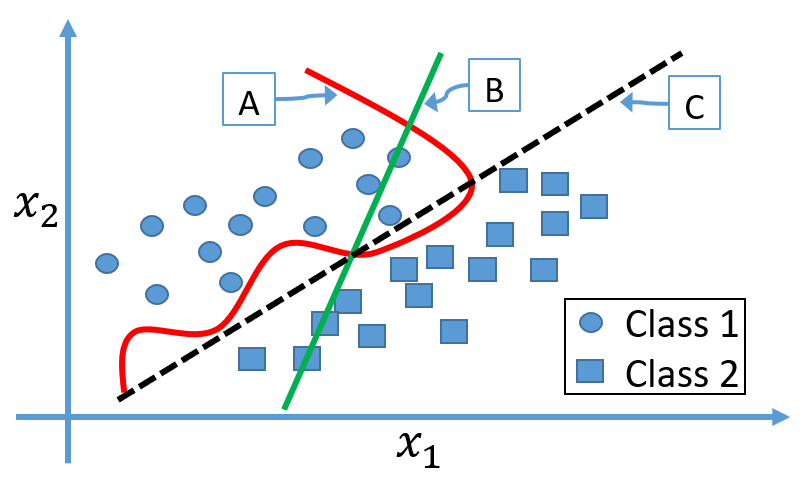}
    \caption{\rebuttalii{Illustrative sketch representing different split-rules on a 2D data. While split-rule $A$ may be overly complex and un-interpretable, split-rules $B$ and $C$ have similar simple and interpretable structures, but $C$ has better coefficients to make it more accurate than $B$.}}
    \label{fig:bilevel_clip_art}
\end{figure}

{\subsection{The Hierarchical Objectives to derive split-rule}}
{
Here, we illustrate the need of formulating the problem of split-rule identification as a bilevel-problem using Figure~\ref{fig:bilevel_clip_art}.}

{The geometry and shape of split-rules is defined by exponent terms $b_{ij}$ appearing in its expression (Eq.~\ref{eq:split_rule} and \ref{eq:power_rule}) while the orientation and location of the split-rule in the feature-space is dictated by the values of coefficients $w_i$ and biases $\theta_i$ (Eq.~\ref{eq:split_rule}).} Thus, the above optimization task of estimating split-rule $f(\mathbf{x})$ involves two types of variables: 
\begin{enumerate}
    \item $B$-matrix representing exponents of $B$-terms (i.e. $b_{ij}$ as shown in Eq.~\ref{eq:B_matrix}) and the modulus flag $m$ indicating the presence or absence of a modulus operator in the expression of $f(\mathbf{x})$, and
    \item weights $\mathbf{w}$ and biases $\mathbf{\Theta}$ in each rule function $f(\mathbf{x})$.
\end{enumerate}
Identification of a good structure for $\mathbf{B}$ terms and value of $m$ is a more difficult task, compared to the weight and bias identification. We argue that if both types of variables ($\mathbf{B}$, $m$, $\Theta$, $\mathbf{w}$) are concatenated in a \rebuttalii{single-level} solution, a good $(\mathbf{B}, m)$ combination may be associated with a 
not-so-good $(\mathbf{w}$, $\mathbf{\Theta})$ combination {(like split-rule $B$ in Fig.~\ref{fig:bilevel_clip_art})}, thereby making the whole solution vulnerable to deletion during optimization. It may be better to separate the search of a good structure of $(\mathbf{B}, m)$ combination from the weight-bias search at the same level, and search for the best weight-bias combination for every $(\mathbf{B}, m)$ pair as a detailed task. This hierarchical structure of the variables motivates us to employ a {\em bilevel\/} optimization approach \cite{sinha2017review} to handle \rebuttalii{the two types of} variables. The upper level optimization algorithm searches the space of $\mathbf{B}$ and $m$. Then, for each $(\mathbf{B}, m)$ pair, the lower level optimization algorithm is invoked to determine the optimal values of $\mathbf{w}$ and $\mathbf{\Theta}$.  {Referring to Fig.~\ref{fig:bilevel_clip_art}, the upper-level will search for the structure of $f(\mathbf{x})$ which might have a nonlinearity (like in rule $A$) or might be linear (like rule $B$). Then, for each upper-level solution (for instance a solution corresponding to a linear rule structure), the lower-level will adjust its weights and biases to determine its optimal location $C$}.

The bi-level optimization problem can be then formulated as shown below:
\begin{equation}
    \label{eq:bilevel_formulation}
\hspace{-1ex}\begin{array}{rl}
\text{Min.}&   
    F_U(\mathbf{B},m,\mathbf{w}^{\ast}, \mathbf{\Theta}^{\ast}), \\
    \text{s.t.} & 
 (\mathbf{w}^{\ast}, \mathbf{\Theta}^{\ast}) \in {\rm argmin}\left\{F_L(\mathbf{w}, \mathbf{\Theta})|_{(\mathbf{B}, m)} \big| \right. \\
 & \quad F_L(\mathbf{w}, \mathbf{\Theta})|_{(\mathbf{B},m)} \leq \tau_I,\\ 
 & \quad \left. -1\leq w_i \leq 1, \ \forall i,\ \mathbf{\Theta} \in [-1,1]^{m+1}\right\},\\
 & m \in \{0,1\},\ b_{ij} \in \{-3,-2,-1,0,1,2,3\},
\end{array}
\end{equation}
where the upper level objective $F_U$ is the quantification of the simplicity of the split-rule, and, the lower-level objective $F_L$ quantifies quality of split resulting due to split-rule $f(\mathbf{x}) \le 0$. An upper-level solution is considered feasible only if it is able to partition the data within some acceptable limit which is set by a parameter $\tau_I$. A more detailed explanation regarding the upper level objective $F_U$ and the lower-level objective $F_L$ is provided in next sections. A pseudo code of the bilevel algorithm to obtain split-rule is provided in Alogrithm~\ref{algo:obtain_split_rule} and the pseudo code provided in Algorithm~\ref{algo:upper_level_eval} gives on overview on the evaluation of population-pool in upper level GA. 

\subsubsection{{Computation of $F_L$}}
The impurity of a node in a decision tree is defined by the distribution of data present in the node. \rebuttalii{\emph{Entropy} and \emph{Gini}-score are popularly used to quantify the impurity of  data distribution.} In this work, we use  \emph{gini-score} to gauge the impurity of a node, \rebuttalii{however, \emph{entropy} measure could also be used. For a binary classification task, \emph{gini}-score ranges from 0-0.5, while \emph{entropy} ranges from 0-1. Hence, the value of $\tau_I$ (Eq.~\ref{eq:bilevel_formulation}) needs to be changed if we use \emph{entropy} instead of \emph{gini} while everything else stays same}. For a given parent node $P$ in a decision tree and two child nodes $L$ and $R$ resulting from it, the \emph{net-impurity of child nodes} ($F_L$) can be computed as follows:
\begin{equation}
\small
    F_L(\mathbf{w}, \mathbf{\Theta})|_{(\mathbf{B}, m)} =  \left(\frac{N_L}{N_P}\text{Gini}(L) + \frac{N_R}{N_P}\text{Gini}(R)\right)_{(\mathbf{w},\mathbf{\Theta},\mathbf{B},m)},
    \label{eq:delta_impurity}
\end{equation}
where $N_P$ is the total number of datapoints present in the parent node $P$, and $N_L$ and $N_R$ indicate the total number of points present in left ($L$) and right ($R$) child nodes, respectively. Datapoints in $P$ which satisfy the split rule $f_P(\mathbf{x}) \le 0$ (Eq~\ref{eq:deci_boundary_our}) go to left child node, while the rest go to the right child node. The objective $F_L$ of minimizing the net-impurity of child nodes favors the creation of purer child nodes (i.e. nodes with a low gini-score). 
 
\subsubsection{{Computation of $F_U$}}
The objective $F_U$ is \emph{subjective} in its form since it targets at dealing with a subjective notion of \emph{generating visually simple} equations of split rule. Usually, equations with more variables and terms seem visually complicated. Taking this aspect into consideration, $F_U$ is set as \emph{the total number of non-zero exponent terms} present in the overall equation of the split rule (Eq.~\ref{eq:deci_boundary_our}). Mathematically, this can be represented with the following equation:
\begin{align}
    \label{eq:simplification_metric}
    & F_U(\mathbf{B},m,\mathbf{w}^{\ast}, \mathbf{\Theta}^{\ast}) = \sum_{i=1}^{p}\sum_{j=1}^{d}g(b_{ij}),
\end{align}
\rebuttalii{where $g(\alpha) = 1$, if $\alpha \ne 0$; zero, otherwise.} 
Here, we use only $\mathbf{B}$ to define $F_U$, but another more comprehensive simplicity term involving presence or absence of modulus operators and relative optimal weight/bias values of the rule can also be used.

\subsection{Upper Level Optimization (ULGA)}
A genetic algorithm (GA) is implemented to explore the search space of power rules $B_i$'s and modulus flag $m$ in the upper level. The genome is represented as a tuple $(\mathbf{B}, m)$ wherein $\mathbf{B}$ is a matrix as shown in Equation~\ref{eq:B_matrix} and $m$ assumes a Boolean value of 0 or 1.

The upper level GA focuses at estimating a simple equation of the split-rule within a desired value of \emph{net impurity} ($F_L$) of child nodes. Thus, the optimization problem for upper level is formulated as a single objective constrained optimization problem as shown in Eq.~\ref{eq:bilevel_formulation}.
The constraint function  $F_L(\mathbf{w}, \mathbf{\Theta})|_{(\mathbf{B},m)}$ is evaluated at the lower level of our bilevel optimization framework. The threshold value $\tau_I$ indicates the desired value of net-impurity of the resultant child nodes (Eq.~\ref{eq:delta_impurity}). In our experiments, we set $\tau_I$ to be 0.05. As mentioned before, a solution satisfying this constraint implies creation of purer child nodes. Minimization of objective function $F_U$ should result in a simplistic structure of the rule and the optimization will also reveal the key variables needed in the overall expression.

\subsubsection{Custom Initialization for Upper Level GA}
Minimization of objective $F_U$ (given in Eq~\ref{eq:simplification_metric}) requires to have less number of active (or non-zero) exponents in the expression of split-rule  (Eq~\ref{eq:deci_boundary_our}). To facilitate this, the population is initialized with a restriction of having only one active (or non-zero) exponent in the expression of split-rule, i.e., any-one of the $b_{ij}$'s in the block matrix $\mathbf{B}$ is set to a non-zero value from a user-specified set $\mathbf{E}$ and the rest of the elements of matrix $\mathbf{B}$ are set to zero. Note here that only $2d$ number of unique individuals ($d$ individuals with $m = 0$ and $d$ individuals with $m = 1$) can exist which satisfy the above mentioned restriction. If the population size for upper level GA exceeds $2d$, then remaining individuals are initialized with two non-zero active-terms.
As the upper level GA progresses, incremental enhancement in rule complexity is realized through crossover and mutation operations, which are described next. 

\subsubsection{Ranking of Upper Level Solutions}
The binary-tournament selection operation and $(\mu + \lambda)$ survival selection strategies are implemented for the upper level GA. Selection operators use the following hierarchical ranking criteria to perform selection:
\begin{definition}
For two individuals $i$ and $j$ in the upper level, $\textit{rank}(i)$ is better than $\textit{rank}(j)$, when any of the following is true:
\begin{itemize}
    \item $i$ and $j$ are both infeasible AND $F_L(i) < F_L(j)$,
    \item $i$ is feasible (i.e. $F_L(i) \le \tau_I$) AND $j$ is infeasible (i.e. $F_L(j) > \tau_I$),
    \item $i$ and $j$ both are feasible AND
         $F_U(i) < F_U(j)$,
    \item $i$ and $j$ both are feasible AND  $F_U(i) = F_U(j)$ \\ AND  $F_L(i) < F_L(j)$,
    \item $i$ and $j$ both are feasible AND  $F_U(i) = F_U(j)$ \\ AND  $F_L(i) = F_L(j)$ AND $m(i) < m(j)$.
\end{itemize}
\end{definition}

\subsubsection{Custom Crossover Operator for Upper Level GA}
Before crossover operation, population members are clustered according to their $m$-value -- all individuals with $m=0$ belong to one cluster and all individuals with $m = 1$ belong to another cluster. The crossover operation is then restricted on individuals belonging to the same cluster. 
The crossover operation in the upper level takes two block matrices from the parent population ($\mathbf{B_{P_1}}$ and $\mathbf{B_{P_2}}$) to create two \emph{child} block matrices ($\mathbf{B_{C_1}}$ and $\mathbf{B_{C_2}}$). 
First, rows of block matrices of participating parents are rearranged in descending order of the magnitude of their corresponding coefficient values (i.e. weights $w_i$ of Eq.~\ref{eq:split_rule}). 
Let the parent block matrices be represented as $\mathbf{B_{P_1}'}$ and $\mathbf{B_{P_2}'}$ after this rearrangement. The crossover operation is then conducted element wise on each row of $\mathbf{B_{P_1}'}$ and $\mathbf{B_{P_2}'}$. For better understanding of the cross-over operation, a psuedo code is provided in Algorithm~\ref{algo:crossover_upperlevel}.

\IncMargin{1em}
\begin{algorithm}
\SetKwData{Left}{left}\SetKwData{This}{this}\SetKwData{Up}{up}
\SetKwFunction{Union}{Union}\SetKwFunction{FindCompress}{FindCompress}
\SetKwInOut{Input}{input}\SetKwInOut{Output}{output}
\SetKwFunction{SortRows}{SortRows}
\SetKwFunction{Size}{Size}
\Input{Block matrices $\mathbf{B_{P_1}}$ and $\mathbf{B_{P_2}}$, and weight vectors $\mathbf{w_{P_1}}$ and $\mathbf{w_{P_2}}$.}
\Output{Child block matrices $\mathbf{B_{C_1}}$ and $\mathbf{B_{C_2}}$.}
\BlankLine
$\mathbf{B_{P_1}'}$ = \SortRows{$\mathbf{B_{P_1}}$,$\mathbf{w_{P_1}}$}\; 
$\mathbf{B_{P_2}'}$ = \SortRows{$\mathbf{B_{P_2}}$, $\mathbf{w_{P_2}}$}\;
$n_{rows}$ = \Size{$\mathbf{B_{P_1}}$,$1$}\;
$n_{cols}$ = \Size{$\mathbf{B_{P_1}}$,$2$}\;
\For{$i \leftarrow 1$ \KwTo $n_{rows}$}
{
\For{$j\leftarrow 1$ \KwTo $n_{cols}$}{\label{forins}
$r$ = rand()\;\tcp{random no. between 0 and 1.}
{
\uIf{$r \le 0.5$}{\label{lt}
$\mathbf{B_{C_1}}(i,j) = \mathbf{B_{P_1}}(i,j)$\;
$\mathbf{B_{C_2}}(i,j) = \mathbf{B_{P_2}}(i,j)$\;
}
\Else{\label{ut}
$\mathbf{B_{C_1}}(i,j) = \mathbf{B_{P_2}}(i,j)$\;
$\mathbf{B_{C_2}}(i,j) = \mathbf{B_{P_1}}(i,j)$\;
}}
}
}
\caption{Crossover operation in upper level GA.}
 \label{algo:crossover_upperlevel}
\end{algorithm}\DecMargin{1em}

\subsubsection{Custom Mutation Operator for Upper Level GA}
\begin{figure}[hbt]
    \centering
    \includegraphics[width = 0.8\linewidth]{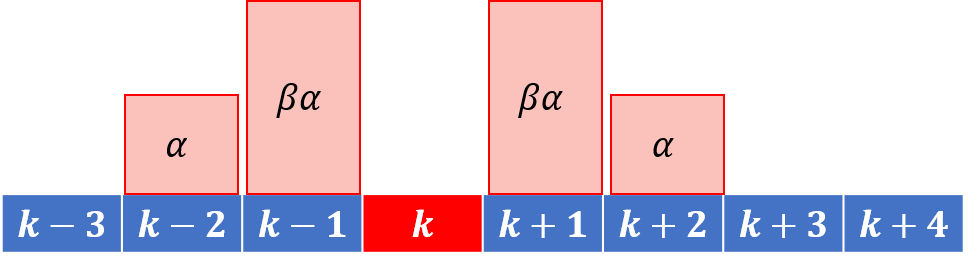}
    \caption{Mutation operation for upper level GA.}
    \label{fig:mutation_operation}
\end{figure}

For a given upper level solution with block matrix $\mathbf{B}$ and modulus flag $m$, the probability with which $b_{ij}$'s and $m$ gets mutated is controlled by parameter $p_{mut}^U$. From the experiments, value of $p_{mut}^U = 1/d$ is found to work well.
The mutation operation then changes the value of exponents $b_{ij}$ and the modulus flag $m$. Let the domain of $b_{ij}$ be given by $\mathbf{E}$, which is a sorted set of finite values. Since $\mathbf{E}$ is sorted, these ordinal values can be accessed using an integer-id $k$, with \emph{id-value} of $k = 1$ representing the smallest value and $k = n_e$ representing the largest value in set $\mathbf{E}$. In our case, $\mathbf{E} = [-3,-2,-1,0,1,2,3]$ (making $n_e = 7$).  The mechanism with which the mutation operator mutates the value of $b_{ij}$ for any arbitrary sorted array $\mathbf{E}$ is illustrated in Fig.~\ref{fig:mutation_operation}.
Here, the \emph{red} tile indicates the \emph{id-value} ($k$) of $b_{ij}$ in array $\mathbf{E}$. The \emph{orange} shaded vertical bars indicate the probability distribution for mutated-values. The $b_{ij}$ can be mutated to either of $k-2, k-1, k+1$, or $k+2$ id-values with a probability of $\alpha$, $\beta\alpha$, $\beta\alpha$, and $\alpha$ respectively. The parameter $\beta$ is preferred to be greater than one and is supplied by the user. \rebuttalii{The parameter $\alpha$ can then be computed as $\alpha = \frac{1}{2(1 + \beta)}$ by equating the sum of probabilities to $1$.} In our experiments, we have set $\beta = 3$.
The value of the modulus flag $m$ is mutated randomly to either 0 or 1 with $50\%$ probability.

In order to bias the search to create simpler rules (i.e. split-rules with a small number of non-zero $b_{ij}$s), we introduce a parameter $p_{zero}$. The value of parameter $p_{zero}$ indicates the probability with which a variable $b_{ij}$ participating in mutation is set to \emph{zero}.
In our case, we use $p_{zero} = 0.75$, \rebuttalii{making} $b_{ij} \rightarrow 0$ with a net probability of $p_{mut}^U\times p_{zero}$.

Duplicates from the the child-population are then detected and changed so that the entire child-population comprises of unique individuals. This is done to ensure creation of novel rules and preservation of diversity.  
Details about this procedure can be found in the supplementary document.

($\mu + \lambda$) survival selection operation is then applied on the combined child and parent population. The selected elites then go to the next generation as parents and the process repeats. Parameter setting for upper-level-GA is provided in the supplementary document.

\subsection{Lower Level Optimization (LLGA)}
\label{sec:lower_level_ga}
For a given population member of the upper level GA (with block matrix $\mathbf{B}$ and modulus flag $m$ as variables), the lower level optimization determines the coefficients $w_i$ (Eq.~\ref{eq:split_rule}) and biases $\theta_i$ such that $F_L(\mathbf{w},\mathbf{\Theta})|_{(\mathbf{B},m)}$ (Eq.~\ref{eq:delta_impurity}) is minimized.
The lower level optimization problem can be stated as below:
\begin{eqnarray}
\label{eq:lower_level_optim}
\mbox{Minimize:} & \quad F_L(\mathbf{w},\mathbf{\Theta})|_{(\mathbf{B},m)},\\
    & \quad \mathbf{w} \in [-1,1]^p,
\nonumber    & \mathbf{\Theta} \in [-1,1]^{m+1}.
\end{eqnarray}

\subsubsection{Custom Initialization for Lower Level GA}
\begin{figure}[hbt]
    \centering
    \includegraphics[width = 0.65\linewidth]{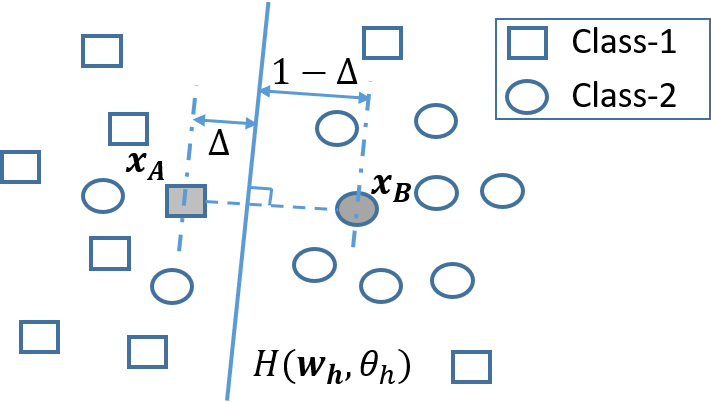}
    \caption{Mixed dipole ($\mathbf{x_A}$, $\mathbf{x_B}$) and a hyperplane $H(\mathbf{w_h},\theta_h)$.}
    \label{fig:dipole}
\end{figure}

In a bilevel optimization, the lower level problem must be solved for every upper level variable vector, thereby requiring a computationally fast algorithm for the lower level problem. Instead of creating every population member randomly, we use the mixed dipole concept 
 \cite{bobrowski2000induction, kretowski2004evolutionary,krketowski2005global} 
 to facilitate faster convergence towards optimal (or near-optimal) values of $\mathbf{w}$ and $\mathbf{\Theta}$. Let $\mathbf{x_A}$ and $\mathbf{x_B}$ be the data-points belonging to Class-1 and Class-2, respectively. Then, weights $\mathbf{w_h}$ and bias $\theta_h$ corresponding to the hyperplane which separates $\mathbf{x_A}$ and $\mathbf{x_B}$, and is orthogonal to vector $\mathbf{x_A - x_B}$, can be given by the following equation:
\begin{align}
    \mathbf{w_h} = & \quad \mathbf{x_A} - \mathbf{x_B},\\
    \theta_h = & \quad \rebuttalii{\Delta\times (\mathbf{w_h}\cdot\mathbf{x_B}) + (1 - \Delta)\times (\mathbf{w_h}\cdot\mathbf{x_A}),}
\end{align}
where $\Delta$ is a random number between 0 and 1. This is pictorially represented in Fig.~\ref{fig:dipole}.

The dipole pairs ($\mathbf{x_A}$, $\mathbf{x_B}$) for all individuals in the initial population pool are chosen randomly from the provided training dataset. The variables
$\mathbf{w}$ and $\mathbf{\Theta}$ for an individual are then initialized as mentioned below:
\begin{align}
   \mathbf{w} = &\quad \mathbf{w_h}, \\
    \theta_1  = &\quad \theta_h, \\
    \theta_2  = &\quad \text{min}(\Delta, 1 - \Delta) \quad \mbox{if $m = 1$}.
\end{align}
This method of initialization is found to be more efficient than randomly initializing $\mathbf{w}$ and $\Theta$. This is because, for the regions beyond the convex-hull bounding the training dataset, the landscape of $F_L$ is flat which makes any optimization algorithm to stagnate.

\subsubsection{Selection, Crossover, and Mutation for Lower Level GA} 
The binary tournament selection, SBX crossover \cite{deb1995simulated} and polynomial mutation \cite{deb2001multi} were used to create the offspring solutions. $(\mu + \lambda)$ survival selection strategy is then adopted to preserve elites.

\subsubsection{Termination Criteria for Lower level GA}
The lower level GA is terminated after a maximum of 50 generations were reached or when the change in the lower level objective value ($F_L$) is less than $0.01\%$ in the past $10$ consecutive generations. 

\rebuttalii{Other parameter values} for lower-level GA can be found in the supplementary document. 
Results from the ablation-studies on LLGA and the overall bilevel-GA are provided in supplementary document for conceptually understanding the efficacy of the proposed approach.

\section{{Overall Tree Induction and Pruning}}
\label{sec:pruning}
Once the split-rule for a conditional node is determined using the above bilevel optimization procedure, the resulting child nodes are checked with respect to the following termination criteria:
{
\begin{itemize}
    \item Depth of the Node $>$ Maximum allowable depth,
    \item Number of datapoints in node $< N_{min}$, and 
    \item Node Impurity $\le \tau_{min}$.
\end{itemize}}
If any of the above criteria is satisfied, then that node is declared as a terminal leaf node. Otherwise, the node is flagged as an additional conditional node. It then undergoes another split (using a split-rule which is derived by running the proposed bilevel-GA on the data present in the node) and the process is repeated. This process is illustrated in Algorithm~\ref{algo:NLDT_Induction}. This eventually produces the main nonlinear decision tree $NLDT_{main}$.
The fully grown decision tree then undergoes pruning to produce the pruned decision tree $NLDT_{pruned}$.

During the pruning phase, splits after the root-node are systematically removed until the training accuracy of the pruned tree does not fall below a pre-specified threshold value of $\tau_{prune} = 3
\%$ (set here after trial-and-error runs). This makes the resultant tree topologically less complex and provides a better generalizability.
In subsequent sections, we provide results on final NLDT obtained after pruning (i.e. $NLDT_{pruned}$), unless otherwise specified.

\section{Results}
\label{sec:results_main}
This section summarizes results obtained on four customized test problems, two real-world classification problems,  three real-world  optimization problems, \rebuttalii{eight multi-objective problems involving 30-500 features and one multi-class problem}. For each experiment, 50 runs are performed with random training and testing sets. The dataset is split into training and testing sets with a ratio of 7:3, respectively. Mean and standard deviation of \emph{training and testing accuracy-scores} across all 50 runs are evaluated to gauge the performance of the classifier. Statistics about the number of {conditional nodes (given by {\#Rules})} in NLDT, average number of active terms in the split-rules of decision tree (i.e. $F_U$/Rule) and \emph{rule length} (which gives \emph{total number} of active-terms appearing in the entire NLDT) is provided to quantify the simplicity and interpretability of classifier {(lesser this number, simpler is the classifier)}. The comparison is made against classical CART tree solution \cite{breiman2017classification} and SVM \cite{vapnik2013nature} results. For SVM, the Gaussian kernel is used and along with overall accuracy-scores; statistics about the number of support vectors (which is also equal to the rule-length) is also reported. It is to note that the decision boundary computed by SVM can be expressed with a single equation, however the length of the equation usually increases with the number of support vectors \cite{vert2004primer}. 

For each set of experiments, best scores are highlighted in bold and Wilcoxon signed-rank test is performed on overall testing accuracy scores to check the statistical similarity with other classifiers in terms of classification  accuracy. Statistically similar classifiers (which will have their $p$-value greater than $0.05$) are italicized.

\subsection{Customized Datasets: DS1 to DS4}
The compilation of results of 50 runs on datasets DS1-DS4 {(see supplementary for details regarding customized datasets)} is presented in Table~\ref{tab:ds1_ds4_results}. The results clearly indicate the superiority of the proposed nonlinear, bilevel based decision tree approach over classical CART and SVM based classification methods.
\begin{table*}[hbt]
\setcellgapes{2pt}
\makegapedcells
    \centering
    \caption{Results on DS1 to DS4 datasets. NLDT is not only most accurate, it also involves a few features.}
    \begin{tabular}{|c|c|c|c|cHH|HHcH|c|c|}\hline
    \multirow{2}{*}{\textbf{Dataset}} & \multirow{2}{*}{\textbf{Method}}  &  \multirow{2}{*}{\textbf{Training Accuracy}} & \multirow{2}{*}{\textbf{Testing Accuracy}} &
    \multirow{2}{*}{\textbf{p-value}} & 
    \textbf{Training} & \textbf{Training} & \textbf{Testing Error} & \textbf{Testing Error} & \multirow{2}{*}{\textbf{\#Rules}} & \multirow{2}{*}{\textbf{n-vars}} & \multirow{2}{*}{\textbf{${\bf F_U}$/Rule}} & \multirow{2}{*}{\textbf{Rule Length}}\\
    & & &  &  & {\bf Type 1} & {\bf Type 2} & {\bf Type 1} & {\bf Type 2} & & & & \\\hline
    \multirow{3}{*}{DS1} &  \rebuttalii{NLDT} & $99.78 \pm 0.51 $ & $ {\bf 99.55 \pm 1.08} $ & -- & $ 0\pm 0$ & $ 0.44 \pm 1.03 $ & $ 0.07 \pm 0.47 $ & $ 0.83 \pm 1.79$ & {\bf 1.0 $\pm$ 0.0} & $2.0 \pm 0.0$ & $ 2.3 \pm 0.6$ & ${\bf 2.3 \pm 0.6} $\\\cline{2-13}
    & CART & $97.99 \pm 0.96 $ & $90.32 \pm 4.06 $ & 7.34e-10 & $1.46 \pm 1.56 $ & $2.56 \pm 1.33 $ & $9.57 \pm 6.84 $ & $9.80 \pm 4.90$ & $14.5 \pm 1.7$ & $1$ & {\bf 1.0 $\pm$ 0.0} & $14.5 \pm 1.7$ \\\cline{2-13}
    & SVM & $98.67 \pm 0.31 $ & $98.50 \pm 1.09 $ &  1.29e-05 & $0\pm 0$ & $2.66 \pm 0.63 $ & $0\pm 0$ & $3 \pm 2.18$ & {\bf 1.0 $\pm$ 0.0} & $2$ & $71.5 \pm 3.3$ & $71.5 \pm 3.3$\\\hline\hline
    
   \multirow{3}{*}{DS2} & \rebuttalii{NLDT} & $99.80 \pm 0.40 $ & ${\bf 99.44 \pm 0.87 }$ & -- & $2.14 \pm 6.33 $ & $0.10 \pm 0.23 $ & $6.00 \pm 12.94 $ & $0.28 \pm 0.60$ & {\bf 1.0 $\pm$ 0.0} & $2.0 \pm 0.0$ & $2.3 \pm 0.7$ & ${\bf 2.3 \pm 0.7}$\\\cline{2-13}
    & CART & $98.80 \pm 0.50 $ & $95.43 \pm 1.50$ &  5.90e-10 & $12.00 \pm 8.95 $ & $0.66 \pm 0.45 $ & $56.00 \pm 20.69 $ & $2.00 \pm 1.39$ & $11.0 \pm 1.4$ & {\bf 1.0 $\pm$ 0.0} & {\bf 1.0 $\pm$ 0.0} & $11.0 \pm 1.4$ \\\cline{2-13}
    & SVM & $96.95 \pm 0.73 $ & $95.24 \pm 0.16 $ &  2.43e-10 & $64.00 \pm 15.34 $ & $0\pm 0$ & $99.67 \pm 2.36 $ & $0.02 \pm 0.12$ & {\bf 1.0 $\pm$ 0.0} & $2$ & $44.7 \pm 1.9$ & $44.7 \pm 1.9$\\\hline\hline
    
    \multirow{3}{*}{DS3} & \rebuttalii{NLDT} & $99.91 \pm 0.35 $ & ${\bf 99.77 \pm 0.67} $ & -- & $0.03 \pm 0.20 $ & $0.14 \pm 0.65 $ & $0.07 \pm 0.47 $ & $0.40 \pm 1.28$ & {\bf 1.0 $\pm$ 0.0} & $2.0 \pm 0.0$ & $2.2 \pm 0.5$ & ${\bf 2.2 \pm 0.5} $\\\cline{2-13}
    & CART & $99.42 \pm 0.57 $ & $95.00 \pm 2.35 $ &  7.00e-10 & $0.40 \pm 0.68 $ & $0.76 \pm 0.80 $ & $5.20 \pm 3.74 $ & $4.80 \pm 3.49$ & $11.5 \pm 1.3$ & $1$ & {\bf 1.0 $\pm$ 0.0} & $11.5 \pm 1.3$ \\\cline{2-13}
    & SVM & $98.96 \pm 0.42 $ & $98.38 \pm 1.15 $ &  7.16e-09 & $0\pm 0$ & $2.07 \pm 0.84 $ & $0\pm 0$ & $3.23 \pm 2.29$ & {\bf 1.0 $\pm$ 0.0} & $2$ & $62.1 \pm 3.4$ & $62.1 \pm 3.4$\\\hline\hline
    
    \multirow{3}{*}{DS4} & \rebuttalii{NLDT} & $99.34 \pm 1.21 $ & ${\bf 98.88 \pm 1.65} $ & -- & $0.19 \pm 0.70 $ & $1.13 \pm 2.11 $ & $0.53 \pm 1.45 $ & $1.70 \pm 2.51$ & $1.2 \pm 0.4$ & $2.0 \pm 0.0$ & $2.6 \pm 0.6$ & ${\bf 3.1 \pm 1.4}$\\\cline{2-13}
    & CART & $96.98 \pm 1.19 $ & $88.68 \pm 3.60 $ & 7.31e-10 & $1.99 \pm 1.36 $ & $4.06 \pm 2.00 $ & $10.33 \pm 4.45 $ & $12.30 \pm 5.38$ & $31.3 \pm 4.2$ & $1$ & {\bf 1.0 $\pm$ 0.0} & $31.3 \pm 4.2$ \\\cline{2-13}
    & SVM & $98.19 \pm 0.44 $ & $97.28 \pm 1.19 $ & 1.31e-05 & $0\pm 0$ & $3.63 \pm 0.89 $ & $0\pm 0$ & $5.43 \pm 2.38$ & {\bf 1.0 $\pm$ 0.0} & $2$ & $89.8 \pm 3.3$ & $89.8 \pm 3.3$\\\hline
    \end{tabular}
    \label{tab:ds1_ds4_results}
\end{table*}
Bilevel approach finds a single rule with two to three appearances of variables in the rule, whereas CART requires 11 to 31 rules involving a single variable per rule, and SVM requires only one rule but involving 44 to 90 appearances of variables in the rule. Moreover, the bilevel approach achieves this with the best accuracy. Thus, the classifiers obtained by bilevel approach are more simplistic and simultaneously more accurate.

\subsection{Breast Cancer Wisconsin Dataset}
\begin{figure}[hbt]
    \centering
    \includegraphics[width = 0.9\linewidth]{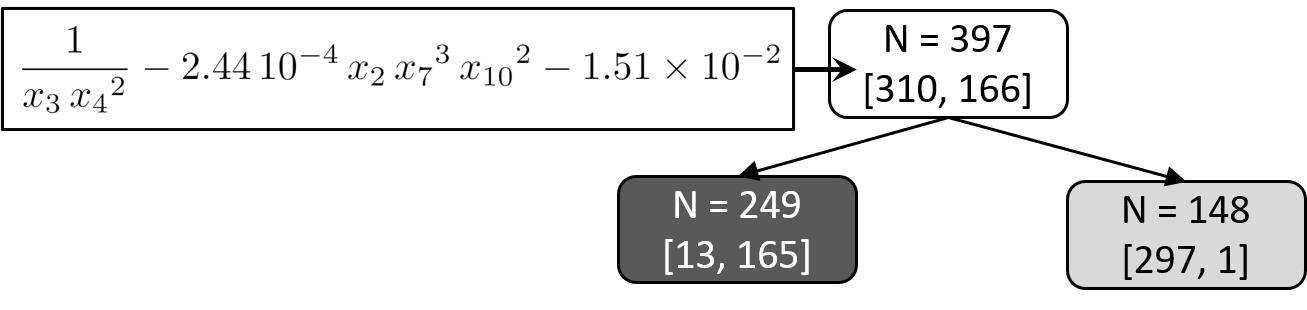}
    \caption{NLDT classifier for breast cancer Wisconsin dataset.}
    \label{fig:tree_breast_cancer}
\end{figure}

This problem was originally proposed in 1991. It has two classes: \emph{benign} and \emph{malignant}, with 458 (or 65.5$\%$)  data-points belonging to \emph{benign} class and 241 (or 34.5$\%$) belonging to \emph{malignant} class. Each data-point is represented with 10 attributes. 
Results are tabulated in Table~\ref{tab:results_breast_cancer}. The bilevel method and SVM had similar performance (\emph{p-value}$>0.05$), but SVM requires about 90 variable appearances in its rule, compared to only about 6 variable appearances in the single rule obtained by the bilevel approach. The proposed approach outperformed the classifiers generated using  techniques proposed in \cite{nunez2002rule, martens2007comprehensible, johansson2004truth, fung2005rule, craven1996extracting} in terms of both: \emph{accuracy} and \emph{comprehensibility/compactness}. The NLDT classifier obtained by a specific run of the bilevel approach has five variable appearances and is presented in Fig.~\ref{fig:tree_breast_cancer}.
\begin{table*}[hbt]
\setcellgapes{2pt}
\makegapedcells
    \centering
    \caption{Results on breast cancer Wisconsin dataset. Despite having a high accuracy, SVM's rule involves about 90 repeated entries of features.}
    \label{tab:results_breast_cancer}
    \begin{tabular}{|c|c|c|cHH|HHc|Hc|c|}\hline
     \multirow{2}{*}{\textbf{Method}}  &  \multirow{2}{*}{\textbf{Training Accuracy}} & \multirow{2}{*}{\textbf{Testing Accuracy}} & \multirow{2}{*}{\textbf{p-value}} & \textbf{Training} & \textbf{Training} & \textbf{Testing Error} & \textbf{Testing Error} & \multirow{2}{*}{\textbf{\#Rules}} & \multirow{2}{*}{\textbf{n-vars}} & \multirow{2}{*}{\textbf{${\bf F_U}$/Rule}} & \multirow{2}{*}{\textbf{Rule Length}}\\
     & &  &  & {\bf Type 1} & {\bf Type 2} & {\bf Type 1} & {\bf Type 2} & & & & \\\hline
\rebuttalii{NLDT} & $98.07 \pm 0.39 $ & $\mathit {96.50 \pm 1.16}$ & $\mathit{0.308}$ & $2.75 \pm 0.57 $ & $0.41 \pm 0.43 $ & $3.78 \pm 1.66 $ & $3 \pm 2.40$ & ${\bf 1.0} \pm {\bf 0.0}$ & $5.3 \pm 1.1$ & $6.4 \pm 1.7$ & ${\bf 6.4 \pm 1.7 }$\\\hline
CART & $98.21 \pm 0.49 $ & $94.34 \pm 1.92 $ & 8.51e-09 & $1.61 \pm 0.58 $ & $2.11 \pm 1.34 $ & $4.28 \pm 1.59 $ & $8.22 \pm 4.25$ & $11.6 \pm 2.4$ & $1$ & {\bf 1.00 $\pm$ 0.00} & $11.6 \pm 2.4$\\\hline
SVM & $97.65 \pm 0.39 $ & ${\bf 96.64 \pm 1.16 }$ & -- & $2.41 \pm 0.44 $ & $2.24 \pm 0.66 $ & $3.66 \pm 1.55 $ & $2.81 \pm 1.81$ & {\bf 1.00 $\pm$ 0.00} & $10$ & $89.4 \pm 14.8$ & $89.4 \pm 14.8$ \\\hline
    \end{tabular}
\end{table*}

\begin{figure}[hbt]
    \centering
    \includegraphics[width = 0.9\linewidth]{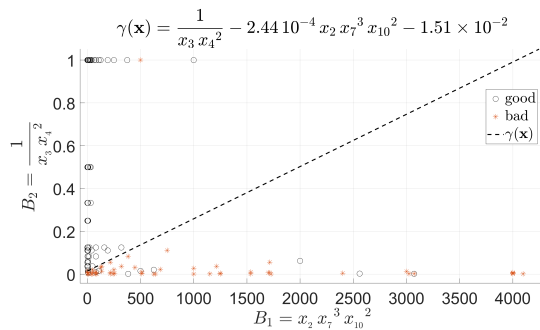}
    \caption{B-space plot for Wisconsin breast cancer dataset.}
    \label{fig:breast_cancer_b_space}
\end{figure}

Figure~\ref{fig:breast_cancer_b_space} provides B-Space visualization of decision-boundary obtained by the bilevel approach, which is able to identify two nonlinear $B$-terms involving variables to split the data linearly to obtain a high accuracy. 

\subsection{Wisconsin Diagnostic Breast Cancer Dataset (WDBC)}
\begin{figure}[hbt]
    \centering
    \includegraphics[width = 0.95\linewidth]{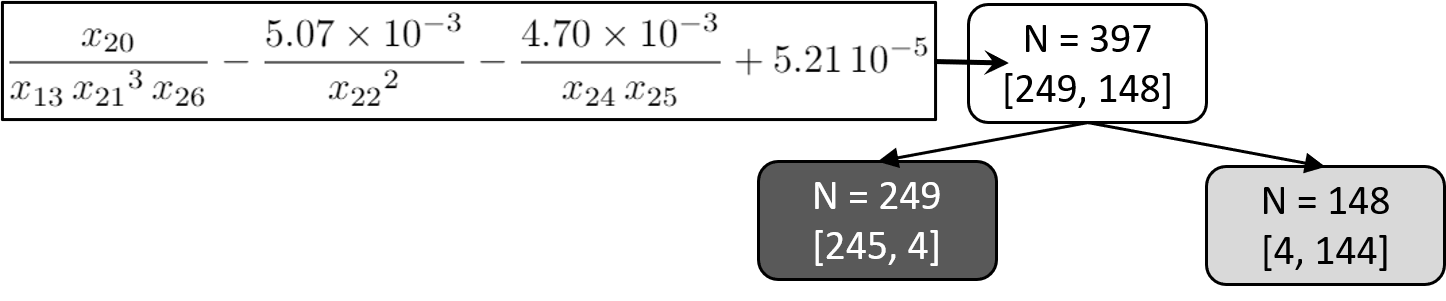}
    \caption{NLDT classifier for WDBC dataset.}
    \label{fig:tree_wdbc}
\end{figure}

This dataset is an extension to the dataset of the previous section. It has 30 features with total 356 datapoints belonging to \emph{benign} class and 212 to \emph{malign} class. Results shown in Table~\ref{tab:wdbc_results} indicate that the bilevel-based NLDT is able to outperform standard CART and SVM algorithms. 
The NLDT generated by a run of the bilevel approach requires seven out of 30 variables (or features) and is shown in Fig.~\ref{fig:tree_wdbc}. It is almost as accurate as that obtained by SVM and is more interpretable (seven versus about 107 variable appearances).
\begin{table*}[hbt]
\setcellgapes{2pt}
\makegapedcells
    \centering
    \caption{Results on WDBC dataset. Despite having a high accuracy, SVM's rule involves about 107 repeated entries of features.}
    \label{tab:wdbc_results}
 \begin{tabular}{|c|c|c|cHH|HHc|Hc|c|}\hline
     \multirow{2}{*}{\textbf{Method}}  &  \multirow{2}{*}{\textbf{Training Accuracy}} & \multirow{2}{*}{\textbf{Testing Accuracy}} & \multirow{2}{*}{\textbf{p-value}} & \textbf{Training} & \textbf{Training} & \textbf{Testing Error} & \textbf{Testing Error} & \multirow{2}{*}{\textbf{\#Rules}} & \multirow{2}{*}{\textbf{n-vars}} & \multirow{2}{*}{\textbf{${\bf F_U}$/Rule}}  & \multirow{2}{*}{\textbf{Rule Length}}\\
    & &  &  & {\bf Type 1} & {\bf Type 2} & {\bf Type 1} & {\bf Type 2} & & & & \\\hline
    \rebuttalii{NLDT} & $98.24 \pm 0.64 $ & $96.20 \pm 1.49 $ & 4.65e-05 & $0.80 \pm 0.71 $ & $3.39 \pm 1.41 $ & $2.58 \pm 1.75 $ & $5.84 \pm 3.09$ & ${\bf 1.0} \pm {\bf 0.0}$ & $8.2 \pm 3.3$ & $9.2 \pm 4.1$ & ${\bf 9.2 \pm 4.1}$\\\hline
    CART & $98.76 \pm 0.60 $ & $92.11 \pm 2.07 $ &  1.09e-09 & $0.86 \pm 0.78 $ & $1.88 \pm 1.25 $ & $6.90 \pm 2.79 $ & $9.56 \pm 3.57$ & $10.8 \pm 2.1$ & $1$ & {\bf 1.00 $\pm$ 0.00} & $10.8 \pm 2.1$\\\hline
    SVM & $98.65 \pm 0.37 $ & ${\bf 97.39 \pm 1.37 }$ & -- & $0.17 \pm 0.24 $ & $3.34 \pm 0.89 $ & $1.70 \pm 1.30 $ & $4.13 \pm 2.27$ & {\bf 1.00 $\pm$ 0.00} & $30$ & $106.7 \pm 6.6$ & $106.7 \pm 6.6$\\\hline
    \end{tabular}
\end{table*}

\begin{figure}[hbt]
    \centering
    \includegraphics[width = 0.9\linewidth]{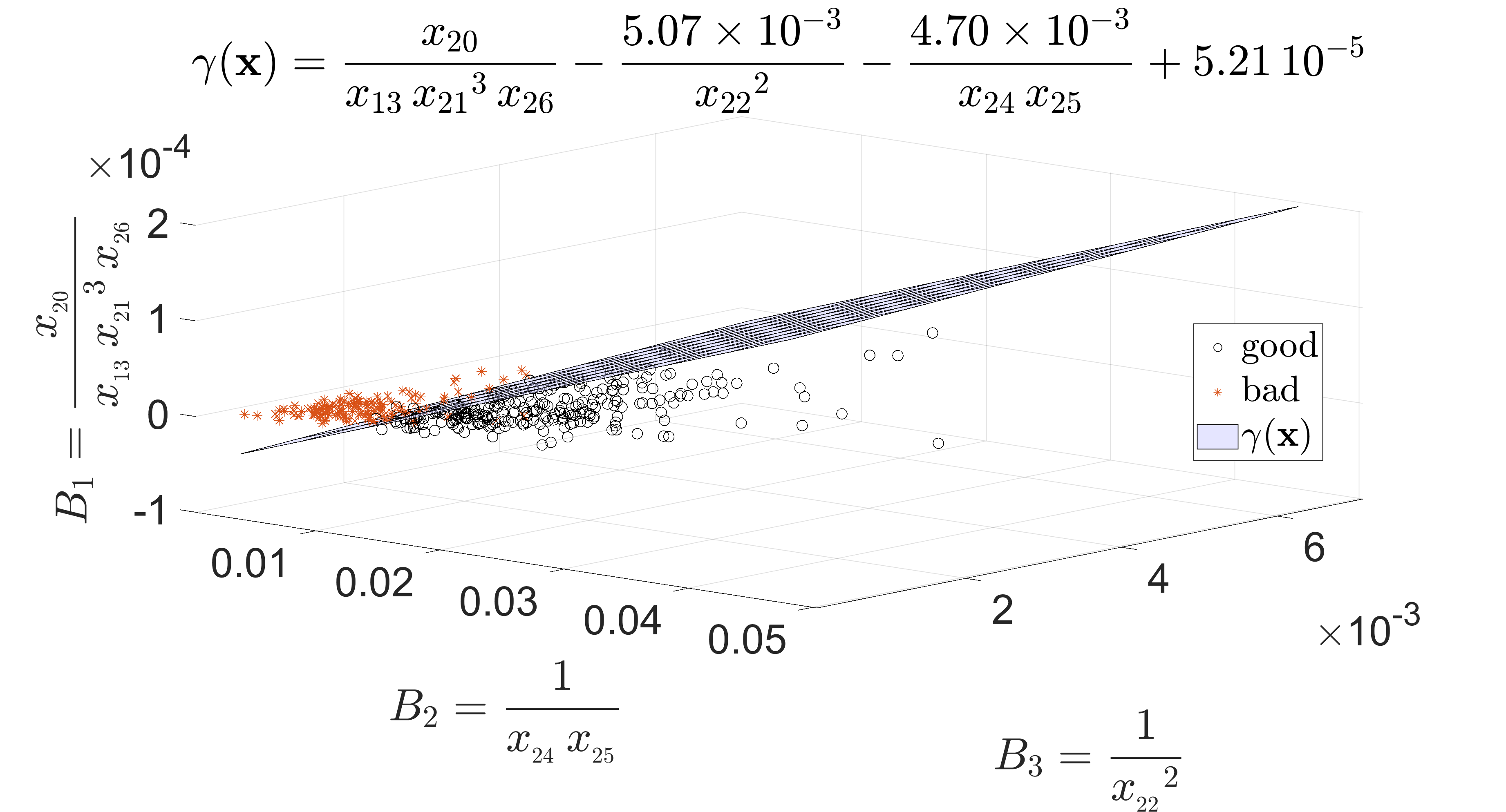}
    \caption{B-space plot for WDBC dataset.}
    \label{fig:wdbc_b_space}
\end{figure}

The $B$-space plot (Fig.~\ref{fig:wdbc_b_space}) shows an efficient discovery of $B$-functions to linearly classify the supplied data with a high accuracy. 

\subsection{Real-World Auto-Industry Problem (RW-problem)}
\begin{figure}[hbt]
    \centering
    \includegraphics[width = \linewidth]{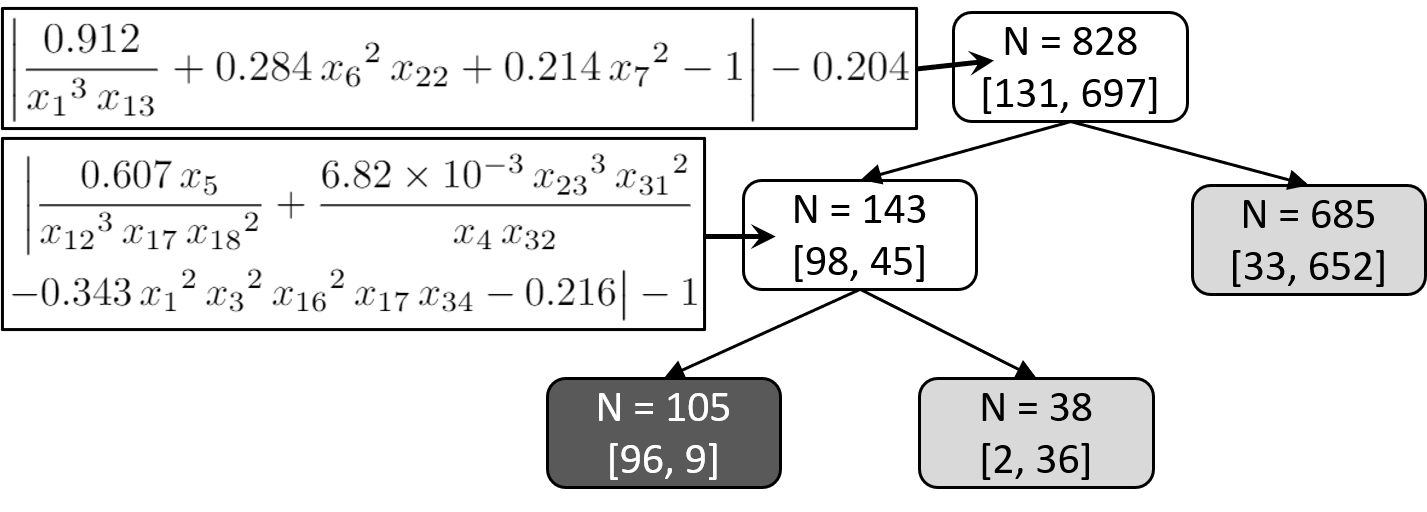}
    \caption{NLDT classifier for the auto-industry problem. The first split-rule uses five  variables and the second one uses 12.}
    \label{fig:auto_original_tree}
\end{figure}

This real-world engineering design optimization problem has 36 variables, eight constraints, and one objective function. The dataset created was highly biased, with 188 points belonging to the \emph{good-class} and 996 belonging to the \emph{bad-class}. Results obtained using the bilevel GA are shown in  Table~\ref{tab:auto_original}. The proposed algorithm is able to achieve near $90\%$ accuracy scores requiring only two split rules. The best performing NLDT has the testing-accuracy of $93.82\%$ and is shown in Fig.~\ref{fig:auto_original_tree}.

SVM performed the best in terms of accuracy, but the resulting classifier is complicated with about 241 variable appearances in the rule. However, bilevel GA requires only about two rules, each having about only 10 variable appearances per rule to achieve slightly less-accurate classification. CART requires about 30 rules with a deep DT, making the classifier difficult to interpret easily.    

\begin{table*}[hbt]
\setcellgapes{2pt}
\makegapedcells
    \centering
    \caption{Results on the real-world auto-industry problem.}
    \begin{tabular}{|c|c|c|cHH|HHc|Hc|c|}\hline
     \multirow{2}{*}{\textbf{Method}}  &  \multirow{2}{*}{\textbf{Training Accuracy}} & \multirow{2}{*}{\textbf{Testing Accuracy}} & \multirow{2}{*}{\textbf{p-value}} & \textbf{Training} & \textbf{Training} & \textbf{Testing Error} & \textbf{Testing Error} & \multirow{2}{*}{\textbf{\#Rules}} & \multirow{2}{*}{\textbf{n-vars}} & \multirow{2}{*}{\textbf{${\bf F_U}$/Rule}}  & \multirow{2}{*}{\textbf{Rule Length}}\\
    &  &  &  & {\bf Type 1} & {\bf Type 2} & {\bf Type 1} & {\bf Type 2} & & & & \\\hline
    \rebuttalii{NLDT} & $94.36 \pm 1.47 $ & $89.93 \pm 2.04 $ & 4.35e-09 & $25.01 \pm 5.45 $ & $2.01 \pm 1.37 $ & $36.25 \pm 7.23 $ & $5.08 \pm 2.11$ & $1.9 \pm 0.5$ & $8.3 \pm 2.7$ & $10.0 \pm 2.9$ & ${\bf 18.2 \pm 5.9}$\\\hline
    CART & $98.00 \pm 0.55 $ & $91.13 \pm 1.32 $ & 9.80e-08 & $7.63 \pm 2.95 $ & $0.94 \pm 0.50 $ & $26.88 \pm 7.56 $ & $5.43 \pm 1.45$ & $29.6 \pm 3.9$ & $1$ & {\bf 1.00 $\pm$ 0.00} & $29.6 \pm 3.9$\\\hline
    SVM & $94.98 \pm 0.73 $ & ${\bf 93.24 \pm 1.38} $ & -- & $19.22 \pm 3.80 $ & $2.35 \pm 0.38 $ & $24.07 \pm 5.57 $ & $3.46 \pm 1.43$ & {\bf 1.00 $\pm$ 0.00} & $36$ & $240.8 \pm 9.4$ & $240.8 \pm 9.4$\\\hline
    \end{tabular}
    \label{tab:auto_original}
\end{table*}

\subsection{Results on Multi-Objective Optimization Problems}
\rebuttalii{Here we evaluate performance of NLDT} on two real-world multi-objective problems: \emph{welded-beam design} and \emph{2D truss design}, \rebuttalii{and modified versions of ZDT \cite{deb2002fast} and DTLZ \cite{deb2002scalable} problems -- \emph{m-ZDT} and \emph{m-DTLZ}. The data generation process of multi-objective datasets is provided in the supplementary document. Complexity of these datasets in controlled by two parameters $\tau_{rank}$ and $g_{ref}$, explanation to which is provided in supplementary document. Results for multi-objective problems are compiled in Table~\ref{tab:multi_objective_all}. Due to space limitations, we report scores of only testing accuracy, total number of rules and $F_U/\textit{Rule}$.} 


\subsubsection{Truss 2D}
\begin{figure}[hbt]
    \centering
    \begin{subfigure}[b]{\linewidth}
        \centering
        \includegraphics[width = 0.8\linewidth]{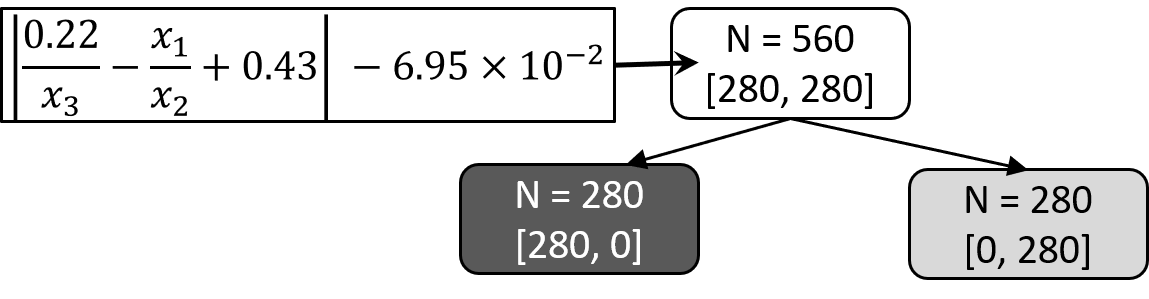}
        \caption{NLDT classifier requiring only one rule.}        \label{fig:truss_bilevel}
    \end{subfigure}%
    \hfill
    \vspace{10pt}
    \begin{subfigure}[b]{0.8\linewidth}
        \centering
        \includegraphics[width = 0.95\linewidth]{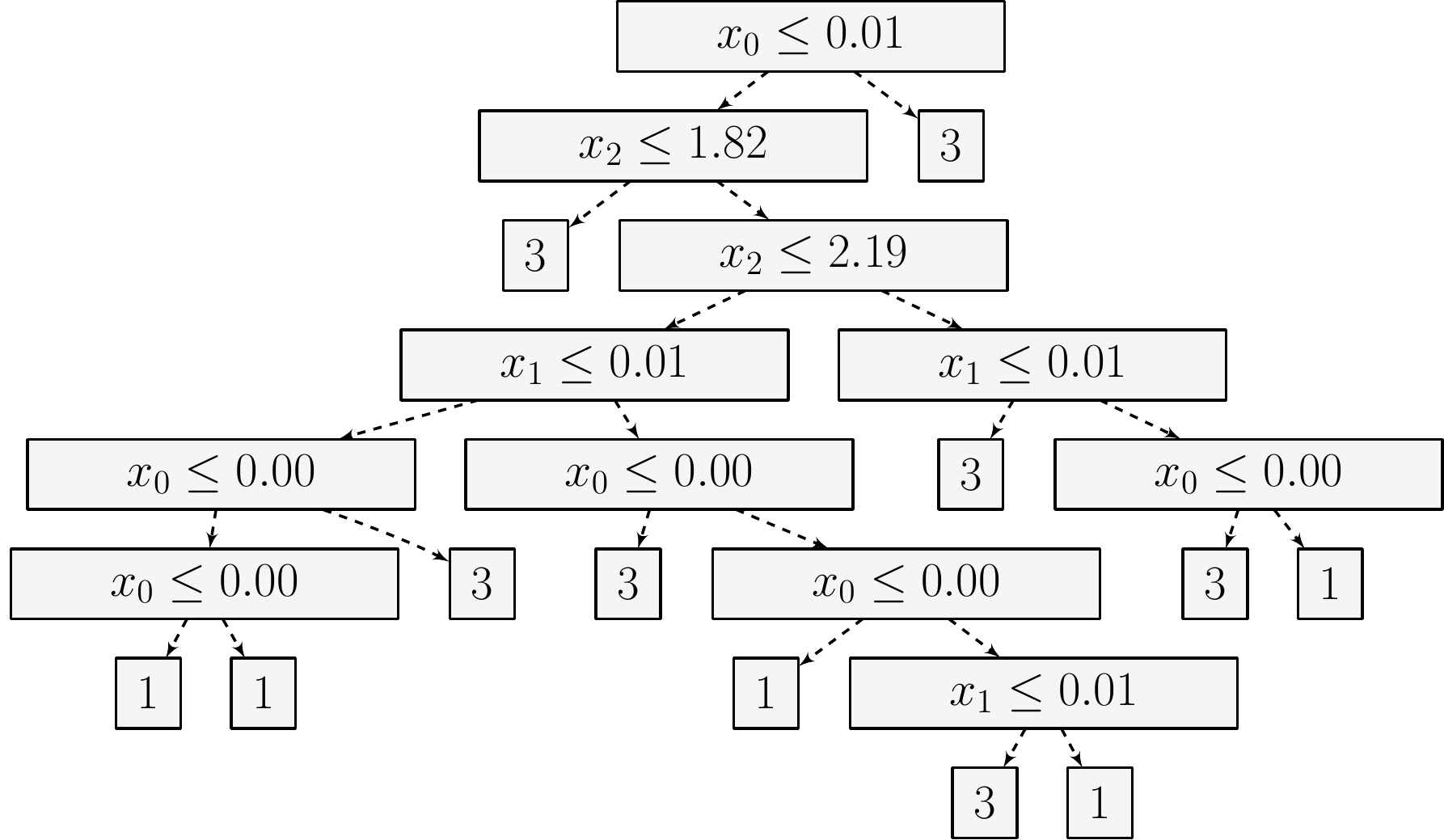}
        \caption{CART classifier \rebuttalii{requiring} 11 rules.}        \label{fig:truss_CART}
    \end{subfigure}%
    \caption{Comparison of bilevel and CART methods on truss problem with $\tau_{rank}=6$ dataset.}
    \label{fig:truss_trees}
\end{figure}

Truss 2D problem \cite{deb2006innovization} is a three-variable problem. Clearly, the bilevel NLDT has the best accuracy and fewer variable appearances (meaning easier intrepretability) compared to CART and SVM generated classifiers \rebuttalii{as shown in Table~\ref{tab:multi_objective_all}.} 
Fig.~\ref{fig:truss_bilevel} shows a 100\% correct classifier with a single rule obtained by our bilevel NLDT, whereas Fig.~\ref{fig:truss_CART} shows a typical CART classifier with 19 rules, making it difficult to interpret easily.

\subsubsection{Welded Beam Design Problem}
\begin{figure}[hbt]
    \centering
    \includegraphics[width = 0.7\linewidth]{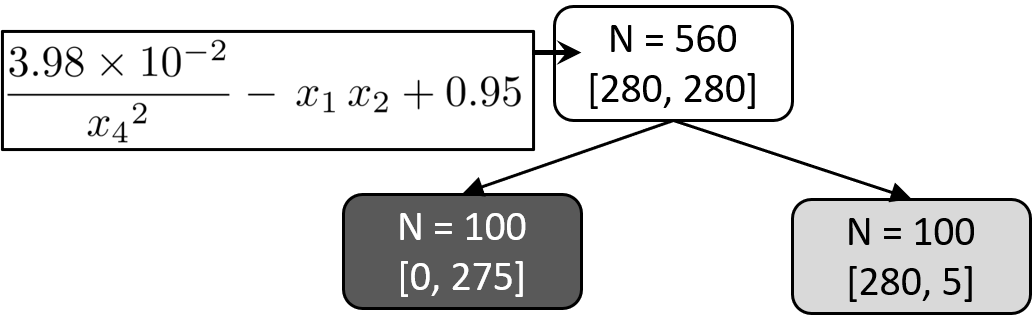}
    \caption{NLDT classifier requiring only one rule for $g_{ref} = 10$ for welded beam problem.}
    \label{fig:welded_beam_NLDT}
\end{figure}

This bi-objective optimization problem has four variables and four constraints \cite{deb2006innovization}. The NLDT (having a single rule) obtained with $g_{ref} = 10$ is shown in Fig.~\ref{fig:welded_beam_NLDT}. Interestingly, the bilevel NLDTs have similar accuracy to that of CART and SVM \rebuttalii{with a very low complexity as shown in Table~\ref{tab:multi_objective_all}.}

\subsubsection{m-ZDT and m-DTLZ Problems}
\rebuttalii{%
Experiments conducted on datasets involving 500 features (see Table~\ref{tab:multi_objective_all}) and for two and three-objective optimization problems confirm the scalability aspect of the proposed approach. In all these problems, traditional methods like CART and SVM find themselves difficult to conduct a proper classification task. The provision of allowing controlled non-linearity at each conditional-node provides the proposed NLDT approach with necessary flexibility to make an appropriate overall classification.}
Additional results and visualization of some of the obtained NLDT classifiers are provided in the supplementary document.

\begin{table*}[hbtp]
\setcellgapes{2pt}
\makegapedcells
    \centering
    \caption{Results on  multi-objective problems for classifying dominated and non-dominated solutions.}
    {\rebuttaliitable
    \begin{tabular}{|c|HcH|c|cH|}\hline
    \textbf{Method} & \textbf{Training Accuracy} & \textbf{Testing Acc.} & \textbf{p-value} & \textbf{\# Rules} & \textbf{$F_U$/Rule} & \textbf{Rule Length}\\\hline\hline
    \multicolumn{7}{|c|}{Welded Beam design with  $g_{ref}= 10$ and $\tau_{rank}=3$.}\\\hline
        NLDT & $99.39 \pm 0.38 $ & $98.58 \pm 1.13 $ & 3.42e-02 & ${\bf 1.0} \pm {\bf 0.0}$ & $3.9 \pm 1.0$ & ${\bf 3.9 \pm 1.0}$\\\hline
    CART & $99.46 \pm 0.27 $ & $97.72 \pm 1.04 $ & 4.63e-08 & $8.42 \pm 1.42$ & {\bf 1.00 $\pm$ 0.00} & $8.42 \pm 1.42$ \\\hline
    SVM & $99.46 \pm 0.19 $ & ${\bf 98.97 \pm 0.54 }$ & -- & {\bf 1.00 $\pm$ 0.00} &  $126.0 \pm 6.8$ & $126.0 \pm 6.8$\\\hline
    \multicolumn{7}{|c|}{m-ZDT1-2-30} \\\hline
    NLDT & $ 99.18 \pm  0.40 $ &  {\bf 98.96 $\pm$ 0.59} & NA & $ 1.80 \pm 0.60 $ & $ 4.47 \pm 2.21 $ & $ 7.64 \pm 3.50 $ \\\hline
    CART & $ 97.48 \pm  0.37 $ & $ 94.78 \pm 1.02 $ & NA & $ 26.42 \pm 1.89 $ & {\bf 1.00 $\pm$ 0.00} & $ 26.42 \pm 1.89 $ \\\hline
    SVM & $ 84.14 \pm  2.42 $ & $ 82.84 \pm 2.77 $ & NA & {\bf 1.00 $\pm$ 0.00} & $ 1192.38 \pm 11.34 $ & $ 1192.38 \pm 11.34 $ \\\hline\hline
    \multicolumn{7}{|c|}{m-ZDT2-2-30} \\\hline
    NLDT & $ 99.23 \pm  0.38 $ & {\bf 98.96 $\pm$ 0.57} & NA & $ 1.92 \pm 0.56 $ & $ 4.37 \pm 1.82 $ & $ 8.14 \pm 3.36 $ \\ \hline
    CART & $ 97.41 \pm  0.36 $ & $ 94.72 \pm 0.89 $ & NA & $ 27.80 \pm 2.19 $ & {\bf 1.00 $\pm$ 0.00} & $ 27.80 \pm 2.19 $ \\\hline
    SVM & $ 99.28 \pm  0.18 $ & $ 98.44 \pm 0.57 $ & NA & {\bf 1.00 $\pm$ 0.00} & $ 315.54 \pm 6.12 $ & $ 315.54 \pm 6.12 $ \\\hline\hline
    
    \multicolumn{7}{|c|}{m-DTLZ1-3-30} \\\hline
    NLDT & $ 97.21 \pm  2.52 $ & {\bf 96.65 $\pm$ 2.86} & NA & $ 3.12 \pm 0.59 $ & $ 6.14 \pm 2.18 $ & $ 19.10 \pm 7.03 $ \\\hline

    CART & $ 89.72 \pm  0.88 $ & $ 71.74 \pm 2.49 $ & NA & $ 93.88 \pm 3.23 $ & {\bf 1.00 $\pm$ 0.00} & $ 93.88 \pm 3.23 $ \\\hline
    SVM & $ 52.44 \pm  0.63 $ & $ 45.86 \pm 1.28 $ & NA & {\bf 1.00 $\pm$ 0.00} & $ 1381.56 \pm 8.25 $ & $ 1381.56 \pm 8.25 $ \\\hline\hline
    
    \multicolumn{7}{|c|}{m-DTLZ2-3-30} \\\hline
    NLDT & $ 97.76 \pm  1.88 $ & {\bf 97.22 $\pm$ 2.25} & NA & $ 3.02 \pm 0.62 $ & $ 5.81 \pm 1.95 $ & $ 17.50 \pm 6.73 $ \\\hline
    CART & $ 85.64 \pm  3.33 $ & $ 63.41 \pm 7.59 $ & NA & $ 100.82 \pm 6.11 $ & {\bf 1.00 $\pm$ 0.00} & $ 100.82 \pm 6.11 $ \\\hline
    SVM & $ 54.82 \pm  0.98 $ & $ 49.61 \pm 1.62 $ & NA & {\bf 1.00 $\pm$ 0.00} & $ 1367.42 \pm 8.44 $ & $ 1367.42 \pm 8.44 $ \\\hline\hline
    \end{tabular}}
    ~
    {\rebuttaliitable
    \begin{tabular}{|c|HcH|c|cH|}\hline
    \textbf{Method} & \textbf{Training Accuracy} & \textbf{Testing Acc.} & \textbf{p-value} & \textbf{\# Rules} & \textbf{$F_U$/Rule} & \textbf{Rule Length}\\\hline\hline
    \multicolumn{7}{|c|}{Truss 2D with $\tau_{rank} = 6$ and $g_{ref} = 1$.}
    \\ \hline
     NLDT & $99.77 \pm 0.72 $ & ${\bf 99.54} \pm {\bf 0.75}$ & -- &  $1.20 \pm 0.50$ & $2.90 \pm 0.50$ & ${\bf 3.30} \pm {\bf 0.90}$\\\hline
    CART & $99.34 \pm 0.32 $ & $98.33 \pm 1.10 $ & 6.04e-08 & $11.06 \pm 3.15$ & {\bf 1.00 $\pm$ 0.00} & $11.06 \pm 3.15$\\\hline
    SVM & $99.66 \pm 0.15 $ & $\mathit{99.46 \pm 0.50} $ & {\it 0.135} & {\bf 1.00 $\pm$ 0.00} & $62.5 \pm 2.90$ & $62.50 \pm 2.90$\\\hline
    
    \multicolumn{7}{|c|}{m-ZDT1-2-500} \\\hline
    NLDT & $ 99.20 \pm  0.29 $ & $ 98.93 \pm 0.60 $ & NA & $ 1.78 \pm 0.54 $ & $ 5.66 \pm 3.23 $ & $ 9.36 \pm 4.15 $ \\\hline
    CART & $ 98.76 \pm  0.27 $ & $ 93.48 \pm 0.94 $ & NA & $ 20.58 \pm 1.39 $ & {\bf 1.00 $\pm$ 0.00} & $ 20.58 \pm 1.39 $ \\\hline
    SVM & $\mathbf{100.00 \pm  0.00}$ & $ \mathbf{100.00 \pm 0.00}$ & NA &  $\mathbf{\bf 1.00 \pm 0.00}$ & $ 240.88 \pm 4.62 $ & $ 240.88 \pm 4.62 $ \\\hline\hline
    
    \multicolumn{7}{|c|}{m-ZDT2-2-500} \\\hline
    NLDT & $ 99.18 \pm  0.38 $ & $ 98.88 \pm 0.71 $ & NA & $ 1.80 \pm 0.69 $ & $ 5.30 \pm 2.45 $ & $ 8.88 \pm 3.98 $ \\\hline
    CART & $ 98.61 \pm  0.36 $ & $ 93.95 \pm 1.33 $ & NA & $ 20.98 \pm 1.98 $ & {\bf 1.00 $\pm$ 0.00} & $ 20.98 \pm 1.98 $ \\\hline
    SVM & {\bf 100.00 $\pm$ 0.00} & {\bf 100.00 $\pm$ 0.00} & NA & {\bf 1.00 $\pm$ 0.00} & $ 248.06 \pm 4.37 $ & $ 248.06 \pm 4.37 $ \\\hline\hline
    
    \multicolumn{7}{|c|}{m-DTLZ1-3-500} 
    \\\hline
    NLDT & $ 94.36 \pm  4.03 $ & {\bf 93.77 $\pm$ 4.24} & NA & $ 3.00 \pm 0.45 $ & $ 7.61 \pm 2.36 $ & $ 22.76 \pm 7.57 $ \\\hline
    CART & $ 83.23 \pm  3.52 $ & $ 58.26 \pm 7.37 $ & NA & $ 104.88 \pm 5.49 $ & {\bf 1.00 $\pm$ 0.00} & $ 104.88 \pm 5.49 $ \\\hline
    SVM & $ 51.56 \pm  0.52 $ & $ 47.13 \pm 1.19 $ & NA & {\bf 1.00 $\pm$ 0.00} & $ 1382.38 \pm 8.84 $ & $ 1382.38 \pm 8.84 $ \\\hline\hline
    
    \multicolumn{7}{|c|}{m-DTLZ2-3-500} 
    \\\hline
    NLDT & $ 95.89 \pm  3.83 $ & {\bf 95.33 $\pm$ 4.46} & NA & $ 3.04 \pm 0.56 $ & $ 7.28 \pm 3.16 $ & $ 22.14 \pm 10.45 $ \\\hline
    CART & $ 89.09 \pm  1.75 $ & $ 70.04 \pm 3.46 $ & NA & $ 96.08 \pm 3.93 $ & {\bf 1.00 $\pm$ 0.00} & $ 96.08 \pm 3.93 $ \\\hline
    SVM & $ 51.30 \pm  0.55 $ & $ 47.01 \pm 1.20 $ & NA & {\bf 1.00 $\pm$ 0.00} & $ 1382.62 \pm 8.46 $ & $ 1382.62 \pm 8.46 $ \\\hline
    
    \end{tabular}}
    
        \label{tab:multi_objective_all}
\end{table*}

\subsection{\rebuttalii{Multi-class Iris Dataset}}
\rebuttalii{Though the purpose of the current study was to develop an efficient algorithm for two-class problems, its extension to multi-class problem is possible.
The iris dataset is a popular multi-class dataset involving total three classes and four features, and is used here as a benchmark problem to check the feasibility of extending our method. The NLDT classifier, producing 94.8\% classification accuracy on test data and 98.6\% accuracy on training data, is presented in Figure~\ref{fig:iris_tree} (see supplementary document for more details). Notice that of the four features, feature~1 ($x_1$) is not required for the classification task. Statistics on results obtained across 50 runs is provided in the supplementary document. We are currently investigating further on a more efficient extension of our approach to multi-class problems.} 
\begin{figure}[hbt]
    \centering
    \includegraphics[width = 0.9\linewidth]{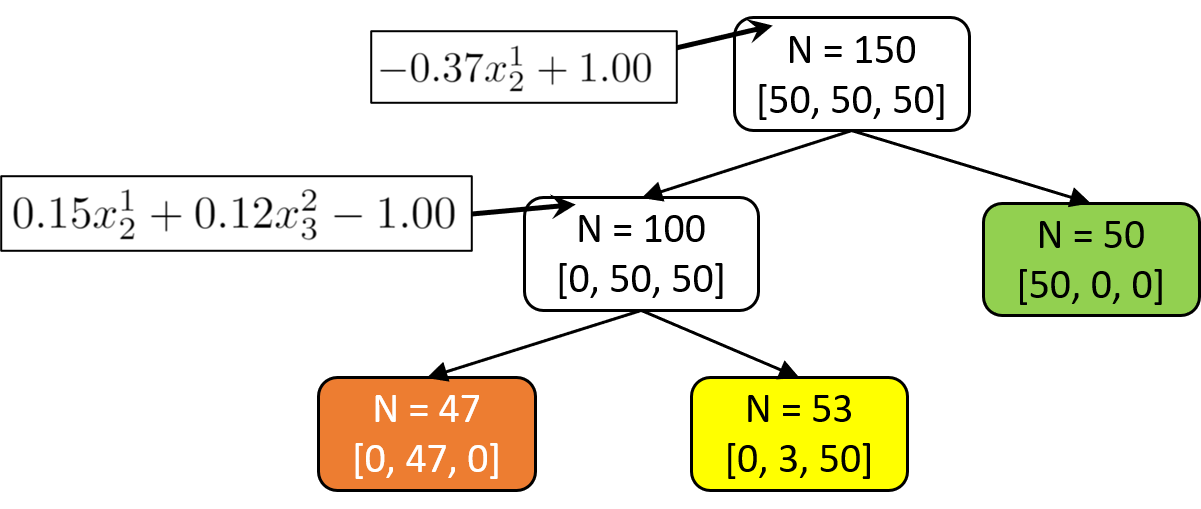}
    \caption{\rebuttalii{NLDT classifier for the three-class iris dataset. The orange node is for datapoints belonging to \emph{iris-versicolor}, yellow for \emph{iris-virginica} and green node indicates datapoints present in \emph{iris-setosa} class.}}
    \label{fig:iris_tree}
\end{figure}

\section{Conclusions and Future work}
\label{sec:conclusion}
In this paper, we have addressed the problem of generating \emph{interpretable} and accurate nonlinear decision trees for classification. Split-rules at the {conditional nodes of a decision tree} have been represented as a weighted sum of power-laws of feature variables. A provision of integrating \emph{modulus} operation in the split-rule has also been formulated, {particularly to handle sandwiched classes of datapoints. The proposed algorithm of deriving split-rule at a given conditional-node in a decision tree} has two levels of hierarchy, wherein the \emph{upper-level} is focused at evolving the power-law structures {and operates on a discrete variable space}, while the \emph{lower-level} is focused at deriving values of optimal weights and biases {by searching a continuous variable space} to minimize the net impurity of child nodes after the split. Efficacy of upper-level and lower-level evolutionary algorithm have been tested on customized datasets. The lower-level GA is able to robustly and efficiently determine weights and biases to minimize the net-impurity. 
A test conducted on imbalanced dataset has also revealed the superiority of the proposed lower-level GA to achieve a high classification accuracy without relying on any synthetic data. Ablation studies conducted on the upper-level GA {(see supplementary document)} also demonstrate the efficacy of the algorithm to obtain simpler split-rules without using any prior knowledge.
Results obtained on standard classification benchmark problems and a real-world single-objective problem has amply demonstrated the efficacy and usability of the proposed algorithm.

The classification problem has been extended to discriminate Pareto-data from non-Pareto data {in a multi-objective problem}. Experiments conducted on multi-objective engineering problems have resulted in determining simple polynomial rules which can assist in determining if the given solution is Pareto-optimal or not. These rules can then be used as design-principles for generating future optimal designs.
As further future studies, we plan to integrate non-polynomial and generic terms in the expression of the split-rules. The proposed bilevel framework can also be tested for \emph{regression} and \emph{reinforcement-learning} related tasks. 
The upper-level problem can be converted to a bi-objective problem for optimizing both $F_U$ and $F_L$ simultaneously, so that a set of trade-off solutions can be obtained in a single run. Nevertheless, this proof-of-principle study on the use of a customized bilevel optimization method for classification tasks is encouraging for us to launch such future studies.

\section*{Acknowledgment}
An initial study of this work was supported by General Motors Corporation, USA. 





%

\bibliographystyle{IEEEtran}
\bibliography{Arxiv_main}
\vspace{-1cm}
\begin{IEEEbiography}[{\includegraphics[width=1in,height=1.25in,clip,keepaspectratio]{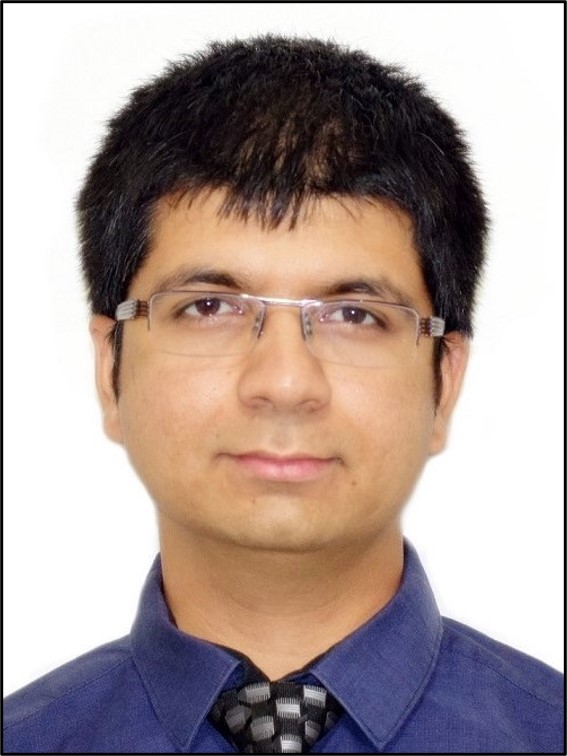}}]{Yasheh Dhebar}
is a PhD student in the Department of Mechanical Engineering at Michigan State University. He did his bachelors and masters in Mechanical Engineering from the Indian Institute of Technology Kanpur and graduated in 2015 with his thesis on modular robotics and their operation in cluttered environments. Currently he is actively pursuing his research in the field of explainable and interpretable artificial intelligence, machine learning and knowledge discovery. He has worked on applied optimization and machine learning projects with industry collaborators. 
\end{IEEEbiography}

\vspace{-1cm}
\begin{IEEEbiography}[{\includegraphics[width=1in,height=1.25in,clip,keepaspectratio]{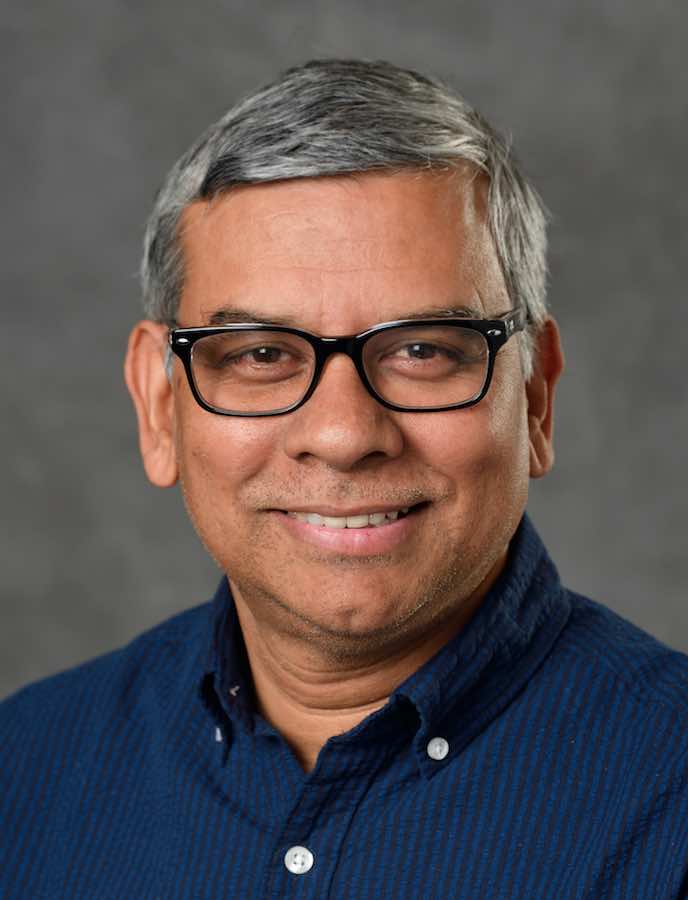}}]{Kalyanmoy Deb}
is Fellow, IEEE and the Koenig Endowed Chair Professor with the Department of Electrical and Computer Engineering, Michigan State University, East Lansing, Michigan, USA. He received his Bachelor’s degree in Mechanical Engineering from IIT Kharagpur in India, and his Master’s and Ph.D. degrees from the University of Alabama, Tuscaloosa, USA, in 1989 and 1991. He is largely known for his seminal research in evolutionary multi-criterion optimization. He has published over 535 international journal and conference research papers to date. His current research interests include evolutionary optimization and its application in design, modeling, AI, and machine learning. He is one of the top cited EC researchers with more than 144,000 Google Scholar citations.
\end{IEEEbiography}

\beginsupplement
\title{Supplementary Document}
\date{}
\author{}
\maketitle

\section*{\textbf{Supplementary Document}}
In this supplementary document, we provide additional information related to our original paper. 

\section{Summary of Supplementary Document}
In this document, we first provide details of parameter settings used for all our proposed algorithms, followed by explanation on the procedure to create customized datasets for benchmarking lower-level and upper-level GAs. Details of the duplicate update operator used in the upper level GA of our proposed bilevel optimization approach is mentioned next. Thereafter, we present additional results to support the algorithms and discussions of the original paper. 

\section{Parameter Setting}
\subsection{Parameter Setting for NSGA-II and NSGA-III}
We used the implementation of NSGA-II and NSGA-III Algorithm as provided in \emph{pymoo} \cite{pymoo}. \emph{pymoo} is an official python package for multiobjective optimization algorithms  and is developed under supervision of the original developer of NSGA-II and NSGA-III algorithm. Following parameter setting was adopted to create Pareto and non-Pareto datasets for two objective Truss and Welded beam problems:
\begin{itemize}
    \item Population Size = 500
    \item Maximum Generations = 1000
    \item Cross Over = SBX
    \item SBX $\eta_c$ = 15
    \item SBX probabililty = 0.9
    \item Mutation type = Polynomial Mutation
    \item Mutation $\eta_m$ = 20
    \item Mutation probability $p_m$ = $1/n_{var}$ (where $n_{var}$ is total number of variables)
\end{itemize}

\subsection{Parameter Setting for Upper Level GA}
This section provides the general parameter setting used for the upper level of our bilevel GA.
\begin{itemize}
    \item Population size = $10 \times d$, where $d$ is the dimension of the feature space of the original problem.
    \item Selection = Binary tournament selection.
    \item Crossover probability ($p_{xover}^U$) = 0.9
    \item Mutation parameters:
        \begin{itemize}
            \item Mutation probability ($p^U_{mut}$) = $\min(0.33, 1/d)$.
            \item Distribution parameter $\beta$ = 3 (any value $> 1$ will be good).
        \end{itemize}
    \item $p_{zero}$ = 0.75
    \item Maximum number of generations = 100
\end{itemize}

\subsection{Parameter Setting for Lower Level GA}
Following parameter setting is used for lower level GA:
\begin{itemize}
    \item Population size = 50,
    \item Maximum generations = 50,
    \item Variable range = $[-1, 1]$,
    \item Crossover type = Simulated Binary Crossover (SBX)
    \item Mutation type = Polynomial Mutation
    \item Selection type = Binary Tournament Selection
    \item Crossover probability = 0.9,
    \item Mutation probability = $1/n_{var}$,
    \item Number of variables $n_{var} = p + 1$, if $m = 0$ or $p + 2$ if $m = 1$, where $p$ is the total number of power-laws allowed and $m$ is the modulus flag,
    \item SBX and polynomial mutation parameters: $(\eta_c, \eta_m) = (2, 15)$.
\end{itemize}

\subsection{Parameter Setting for Inducing a Non-linear Decision Tree}
\begin{itemize}
    \item Number of power-laws per split-rule: $p = 3$,
    \item Allowable set of exponents: $\mathbf{E} = \{-3,-2,\dots,3\}$,
    \item Impurity Metric: Gini Score.
    \item Minimum Impurity to conduct a split: $\tau_{\min} =  0.05$
    \item Minimum number of points required in a node conduct a split: $N_{min} = 10$
    \item Maximum allowable Depth = 5
    \item Pruning threshold: $\tau_{prune} = 3\%$
\end{itemize}

\section{Creation of Customized 2D Datasets}
Datasets presented in this section were synthetically generated. Datapoints in $2D$ space were created using a reference curve $\xi(\mathbf{x})$ and were assigned to either the \emph{Class-1} or \emph{Class-2}. Two additional parameters $\delta$ and $\sigma$ controlled the location of a datapoint relative to the reference curve $\xi(\mathbf{x})$. Datapoints ($\mathbf{x^{(i)}}$) for a given dataset were generated using following steps:
\begin{itemize}
    \item First $n$ points falling on  curve $\xi(\mathbf{x}) = 0$ were initialized for reference. Lets denote these reference points on the curve with $\mathbf{x_r^{(i)}}$ (where $i = 1,2,\dots,n$). Thus, \[\xi(\mathbf{x_r^{(i)})} = 0, \qquad i = 1,2,\dots, n.\]
    \item  $n$ points ($\mathbf{x^{(i)}}$) for  a dataset were then generated using the following equation
    \begin{equation}
    \mathbf{x^{(i)}} = \mathbf{x_r^{(i)}} + \delta + \sigma r, \qquad i = 1, 2, \dots, n,
    \label{eq:data_point}
    \end{equation}
where $r$ is a random number between $0$ and $1$, $\delta$ represents the \emph{offset} and $\sigma$ indicates the amount of \emph{spread}.
\end{itemize}

\begin{figure*}[hbt]
    \centering
    \begin{subfigure}[b]{0.33\textwidth}
    \includegraphics[width = \textwidth]{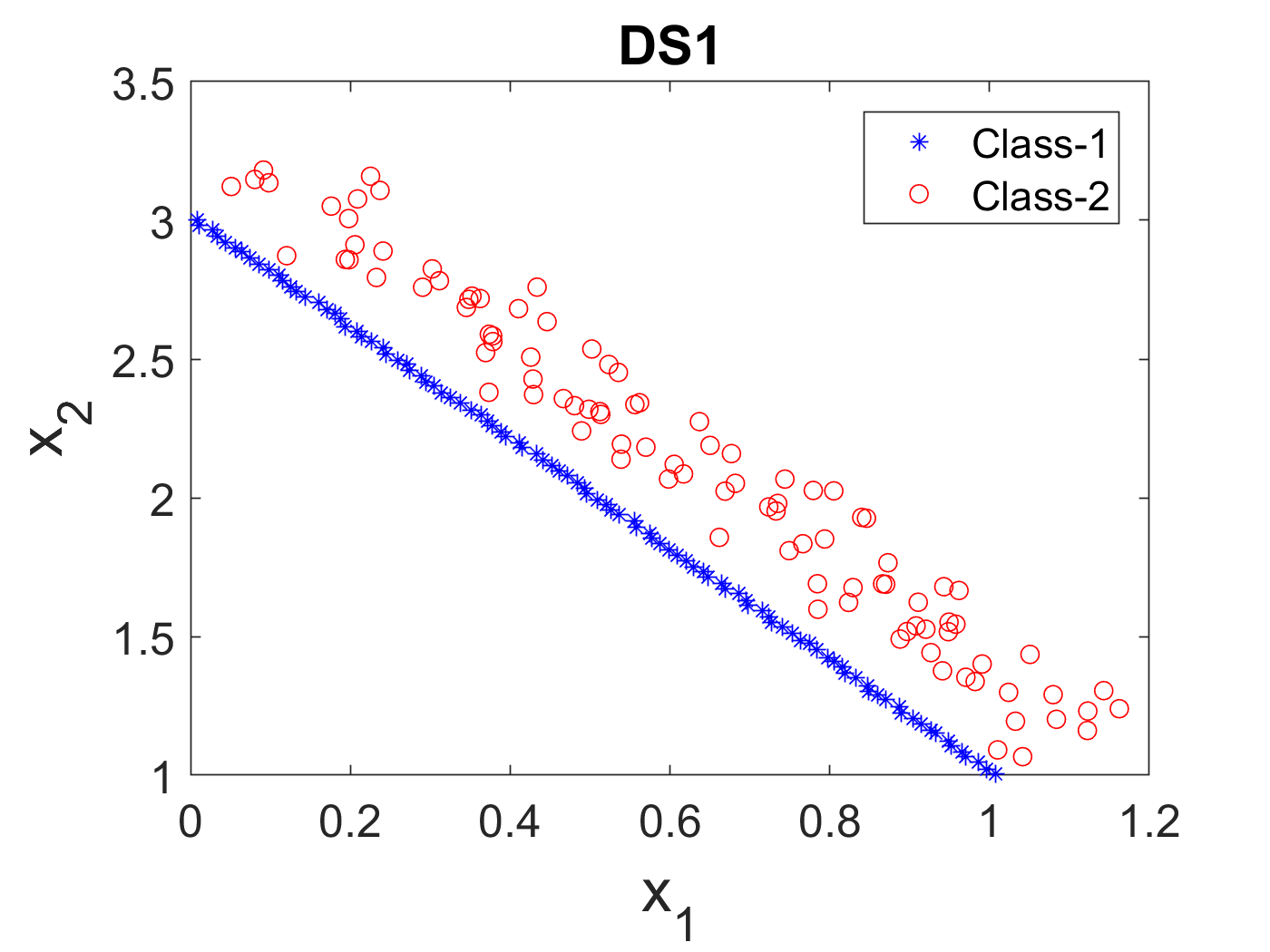}
    \caption{DS1 Dataset.}
    \end{subfigure}%
    ~
    \begin{subfigure}[b]{0.33\textwidth}
    \includegraphics[width = \textwidth]{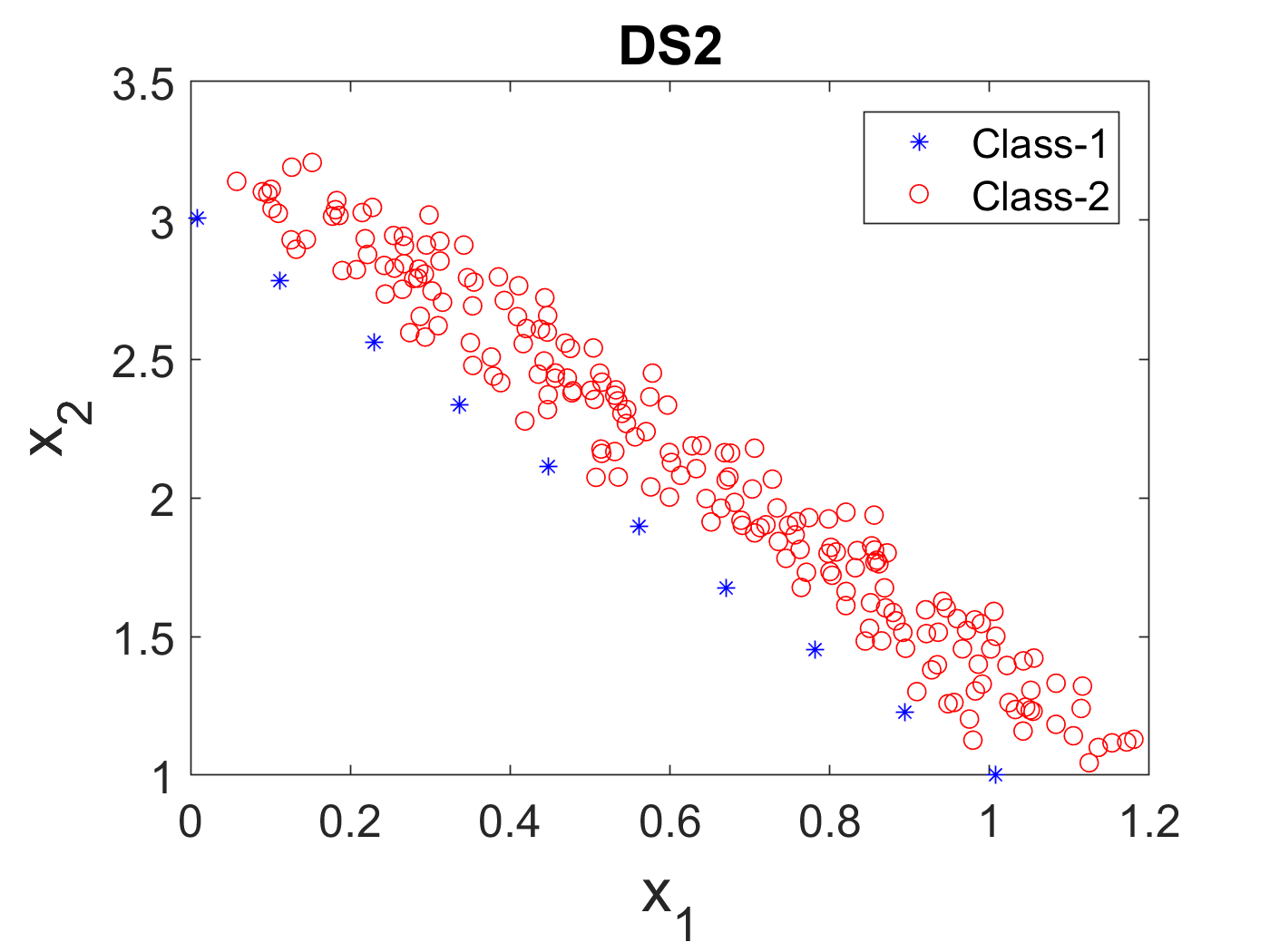}
    \caption{DS2 Dataset.}
    \end{subfigure}%
    \\
    \begin{subfigure}[b]{0.33\textwidth}
    \centering
    \includegraphics[width = \textwidth]{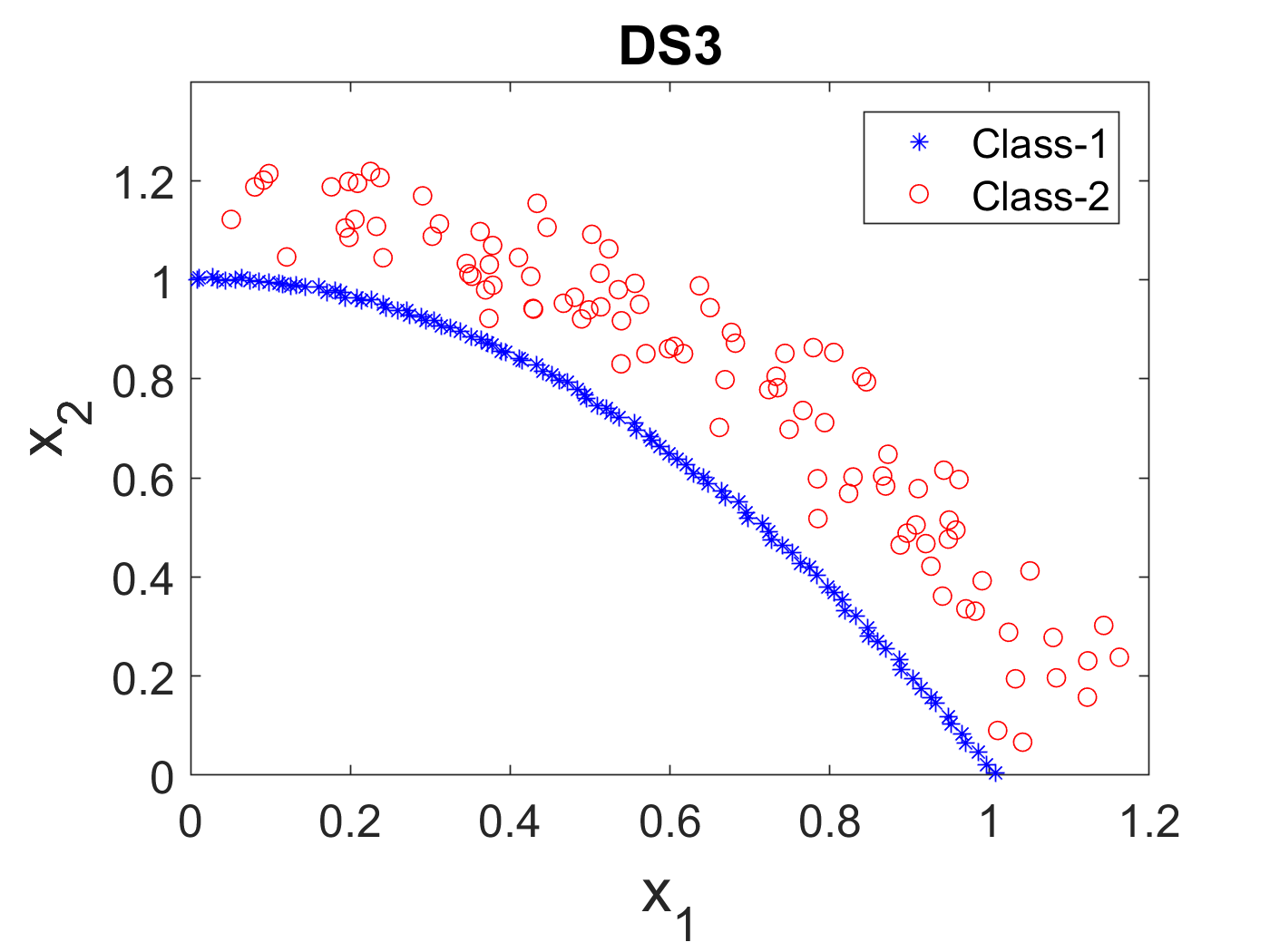}
    \caption{DS3 Dataset.}
    \end{subfigure}%
    ~
    \begin{subfigure}[b]{0.33\textwidth}
    \centering
    \includegraphics[width = \textwidth]{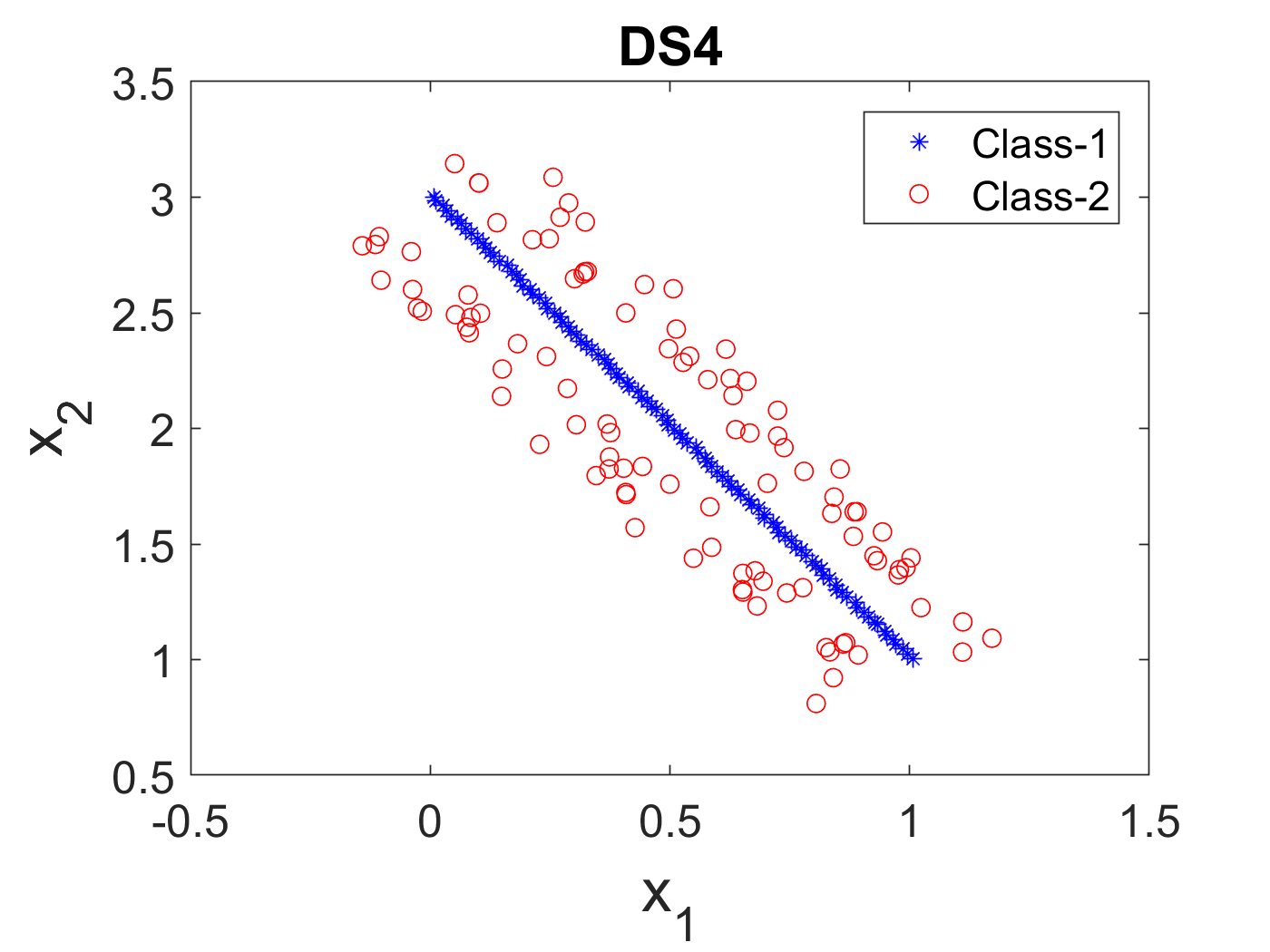}
    \caption{DS4 Dataset.}
    \label{fig:ds4_dataset}
    \end{subfigure}%
    \caption{Customized datasets.}
    \label{fig:customized_datasets}
\end{figure*}

Four datasets generated for conducting ablation studies are graphically represented in Fig.~\ref{fig:customized_datasets}.

Parameter setting which was used to generate these datasets is provided Table~\ref{tab:customized_datasets}. $N_A$ and $N_B$ indicate the number of datapoints belonging to \emph{Class-1} and \emph{Class-2}, respectively.

\begin{table}[hbt]
\setcellgapes{2pt}
\makegapedcells
\caption{Parameter setting to create customized datasets \emph{D1-D4}. For Class-1 data-points $(\delta, \sigma) = (0, 0.01)$.}
    \label{tab:customized_datasets}
    \centering
    \begin{tabular}{|c|c|c|c|c|}\hline
      \multirow{2}{*}{\bf Dataset} & \multirow{2}{*}{$\xi(\mathbf{x})$} & {\bf Class-1} & \multicolumn{2}{c|}{\bf Class-2}  \\\cline{3-5}
    &  & $N_A$ & $N_B$ & $(\delta, \sigma)$\\\hline
   DS1 & $2x_1 + x_2 - 3 = 0$ & 100 &  100 & $(0.025, 0.2)$\\\hline
   DS2 & $2x_1 + x_2 - 3 = 0$ & 10 &   200 & $(0.025, 0.2)$\\\hline
   DS3 & $x_1^2 + x_2 - 1 = 0$ & 100 &   100 & $(0.025, 0.2)$\\\hline
   \multirow{2}{*}{DS4} & \multirow{2}{*}{$2x_1 + x_2 - 3 = 0$} & \multirow{2}{*}{100} &   50 & $(0.025, 0.2)$ \\ \cline{4-5}
    & & & 50 & $(-0.015, -0.2)$\\\hline
    \end{tabular}
\end{table}

\section{Multi-objective Datasets}

\subsection{\rebuttalii{ZDT, Truss and Welded Beam Datasets}}
\rebuttalii{For each problem, NSGA-II \cite{deb2002fast} algorithm is first applied to evolve the population of $N$ individuals for $g_{max}$ generations. Population for each generation is stored. Naturally, population from later generations are closer to Pareto-front than the population of initial generations. We artificially separate the entire dataset into two classes -- good and bad -- using two parameters $g_{ref}$(indicating an intermediate generation for data collection) and $\tau_{rank}$(indicating the minimum non-dominated rank for defining the bad cluster). First, population members from $g_{ref}$ and $g_{max}$ generations are merged. Next, non-dominated sorting is executed on the merged population to determine their non-domination rank. Points belonging to same non-domination rank are further sorted based on their crowding-distance (from highest to lowest) \cite{deb2002fast}. For the good-class, top $N_A$ points from rank-1 front are chosen and for the bad-class, top $N_B$ points from $\tau_{rank}$ front onward are chosen. Increasing the $\tau_{rank}$ increases the separation between good and bad classes, while the $g_{ref}$ parameter has the inverse effect. }

\begin{figure*}[hbt]
    \centering
    
    \begin{subfigure}[b]{0.45\textwidth}
    \centering
    \includegraphics[width = \textwidth]{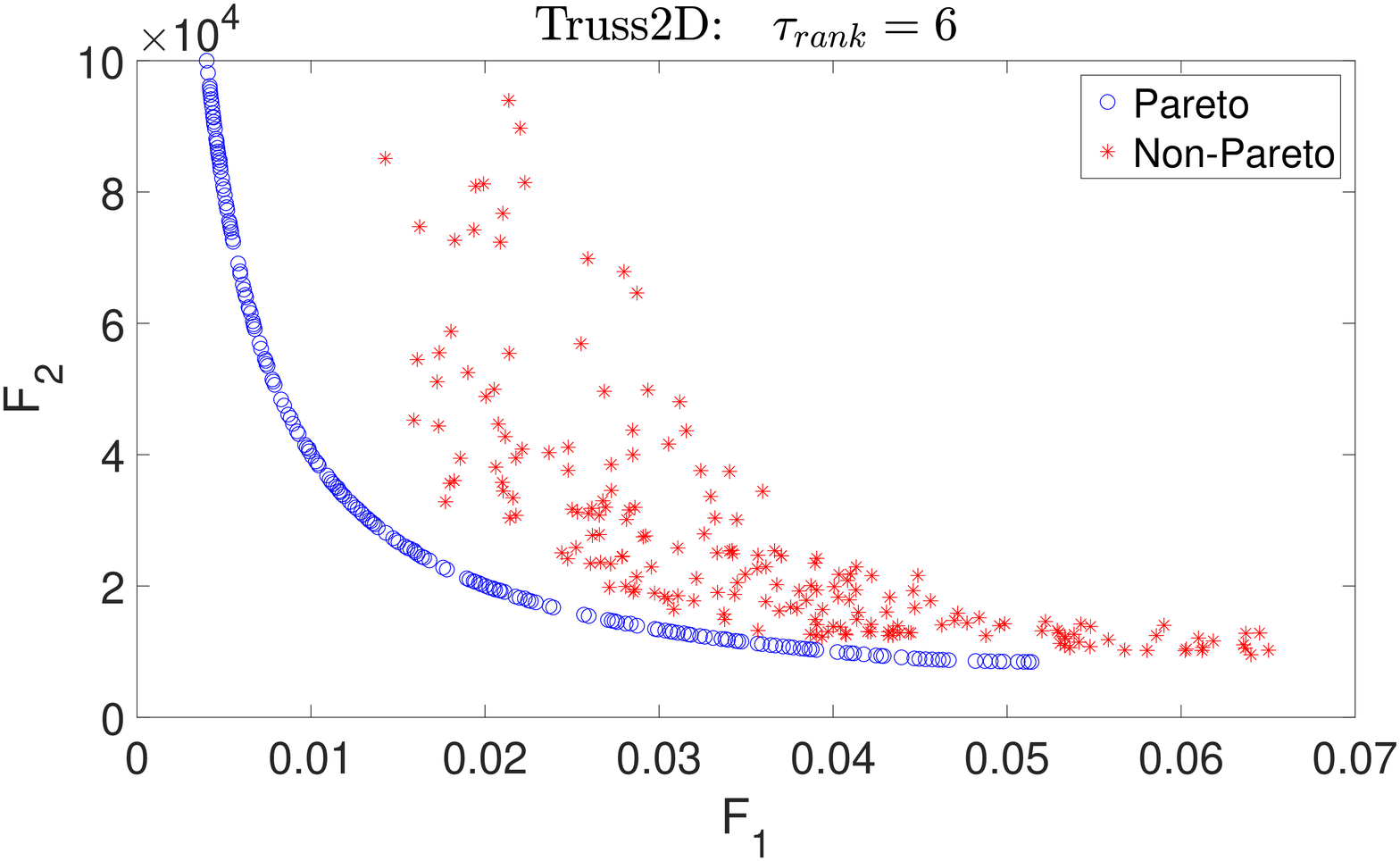}
    \caption{F-space plot with $\tau_{rank} = 6$.}
    \label{fig:truss_6f}
    \end{subfigure}%
    \hfill
    \begin{subfigure}[b]{0.45\textwidth}
    \centering
    \includegraphics[width = \textwidth]{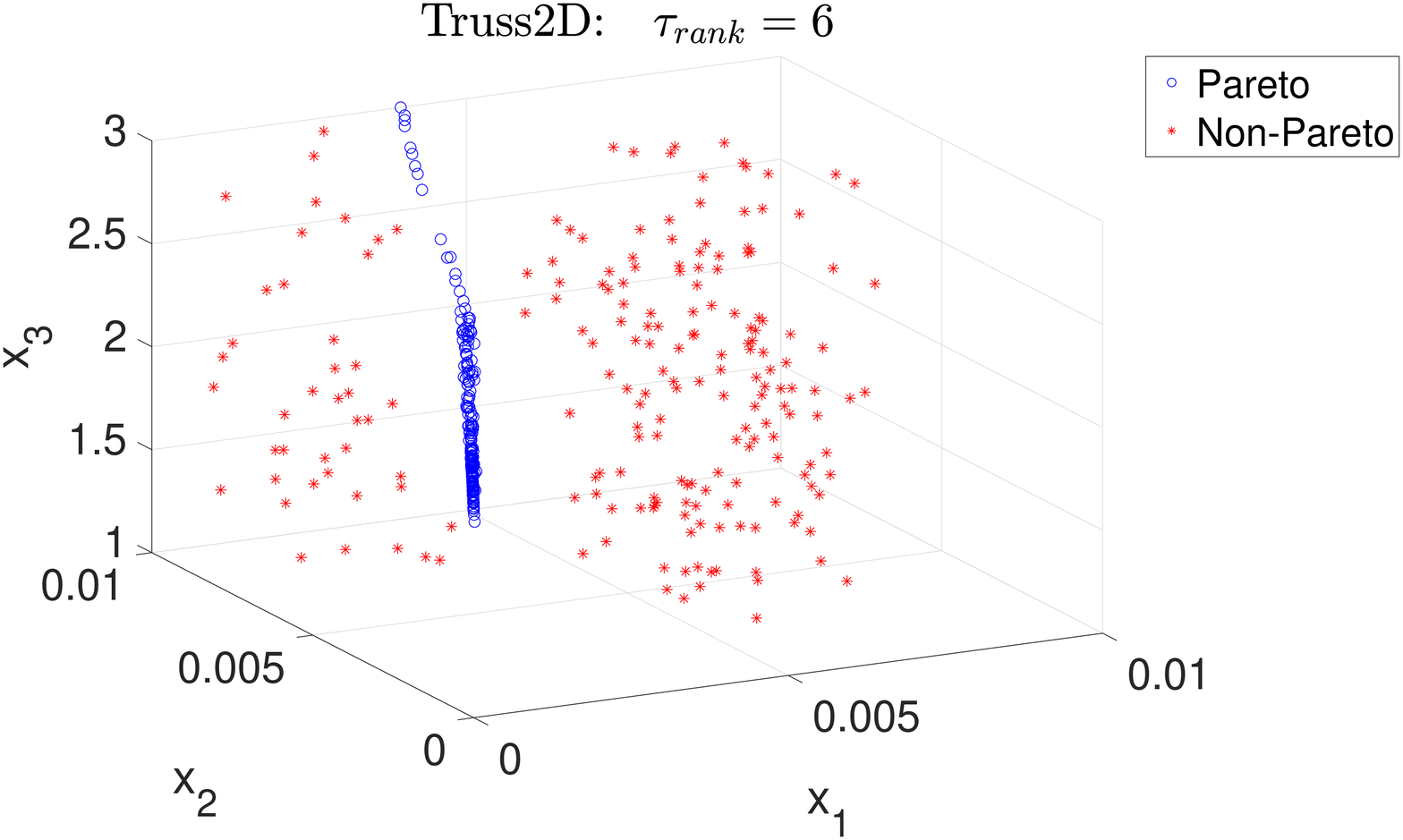}
    \caption{X-space plot with $\tau_{rank} = 6$.}
    \label{fig:truss_6x}
    \end{subfigure}%
    \hfill
        \begin{subfigure}[b]{0.45\textwidth}
        \centering
        \includegraphics[width = \textwidth]{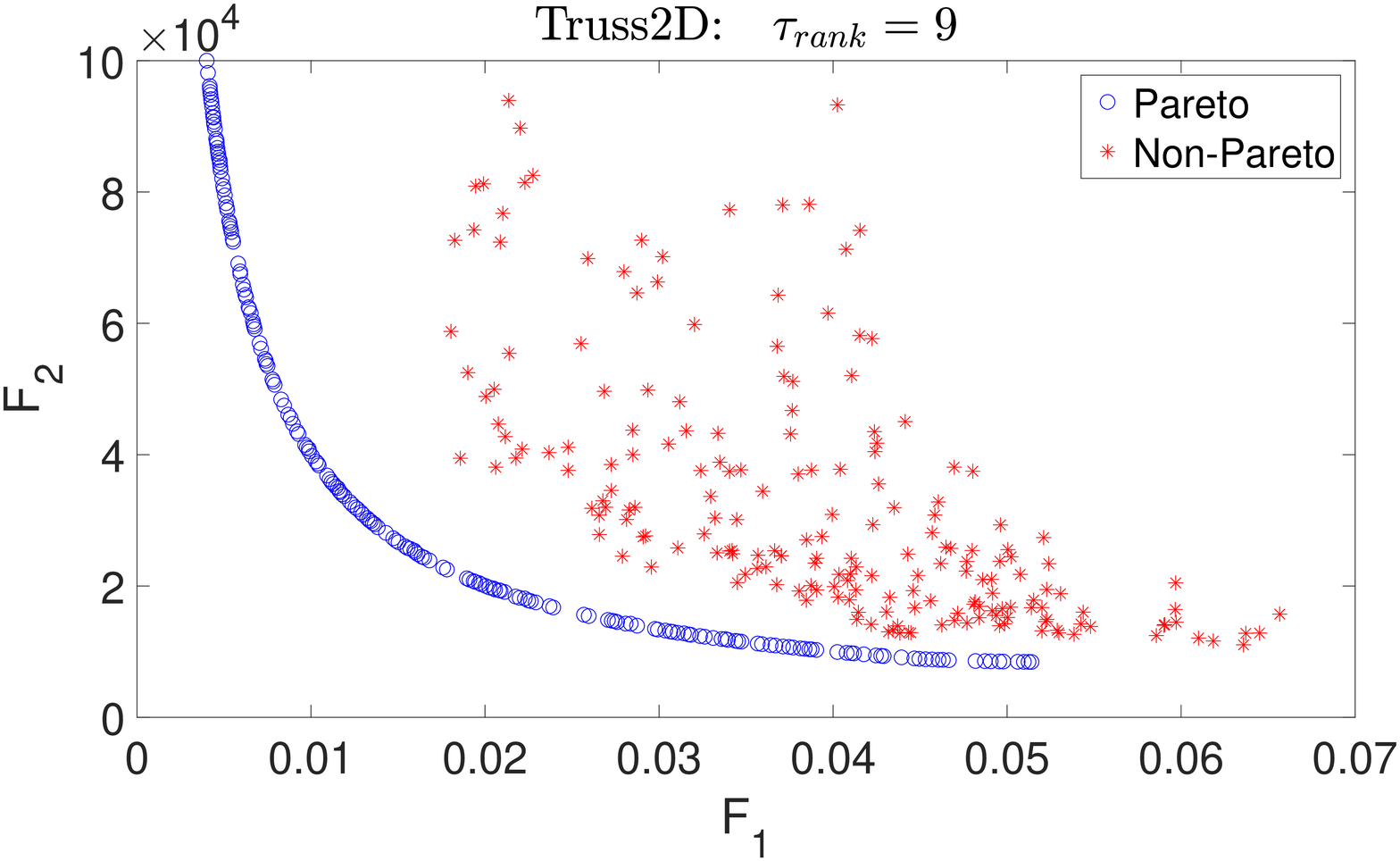}%
        \caption{F-space plot with $\tau_{rank} = 9$.}
    \end{subfigure}%
    \hfill
    \begin{subfigure}[b]{0.45\textwidth}
        \centering
        \includegraphics[width = \textwidth]{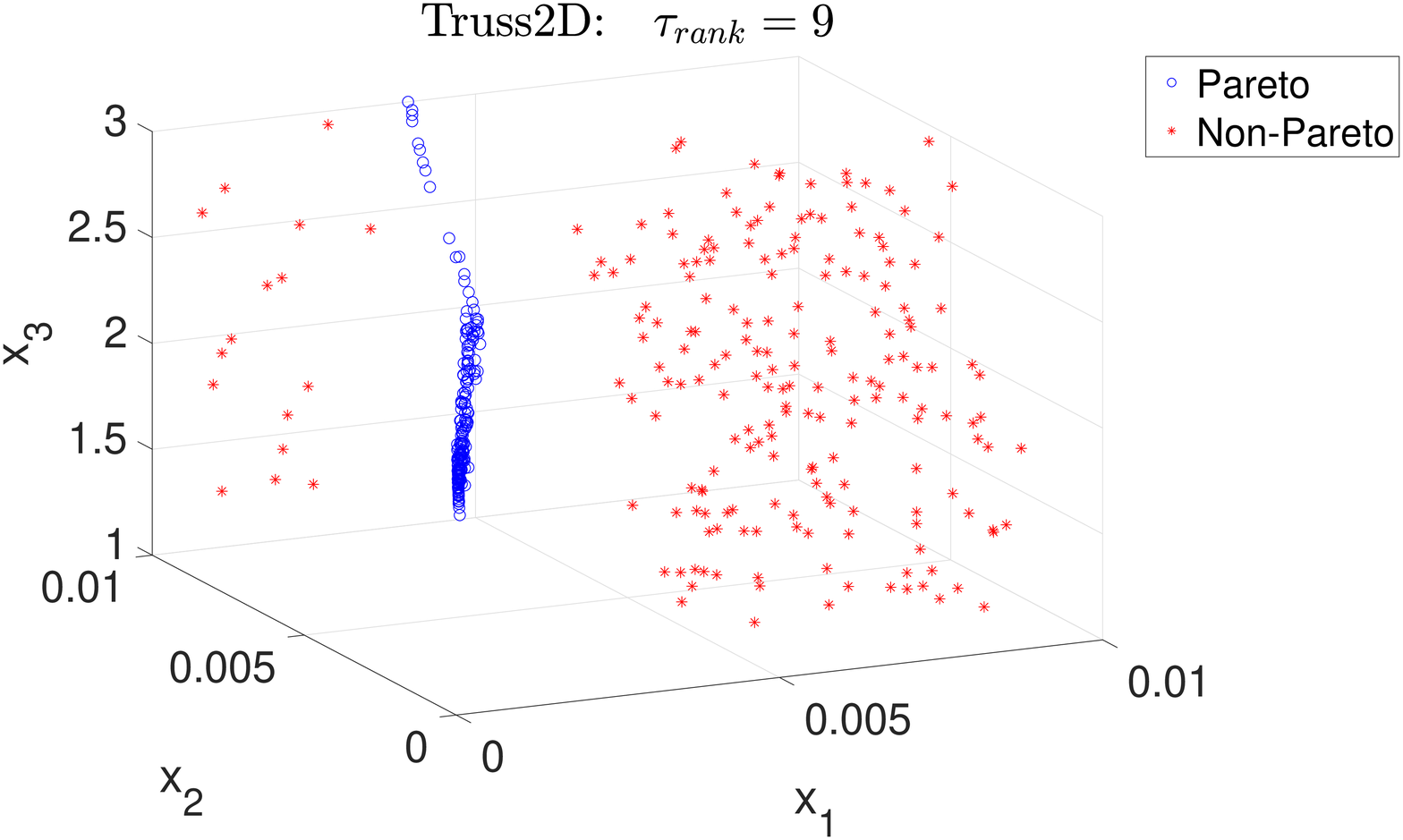}%
        \caption{X-space plot with $\tau_{rank} = 9$.}
        \label{fig:truss_9x}
    \end{subfigure}%
    \caption{Truss design problem data visualization. $g_{ref} = 1$ is kept fixed.}
    \label{fig:truss_data}
\end{figure*}

\begin{figure*}[hbt]
    \centering
    
    \begin{subfigure}[b]{0.4\textwidth}
    \centering
    \includegraphics[width = \linewidth]{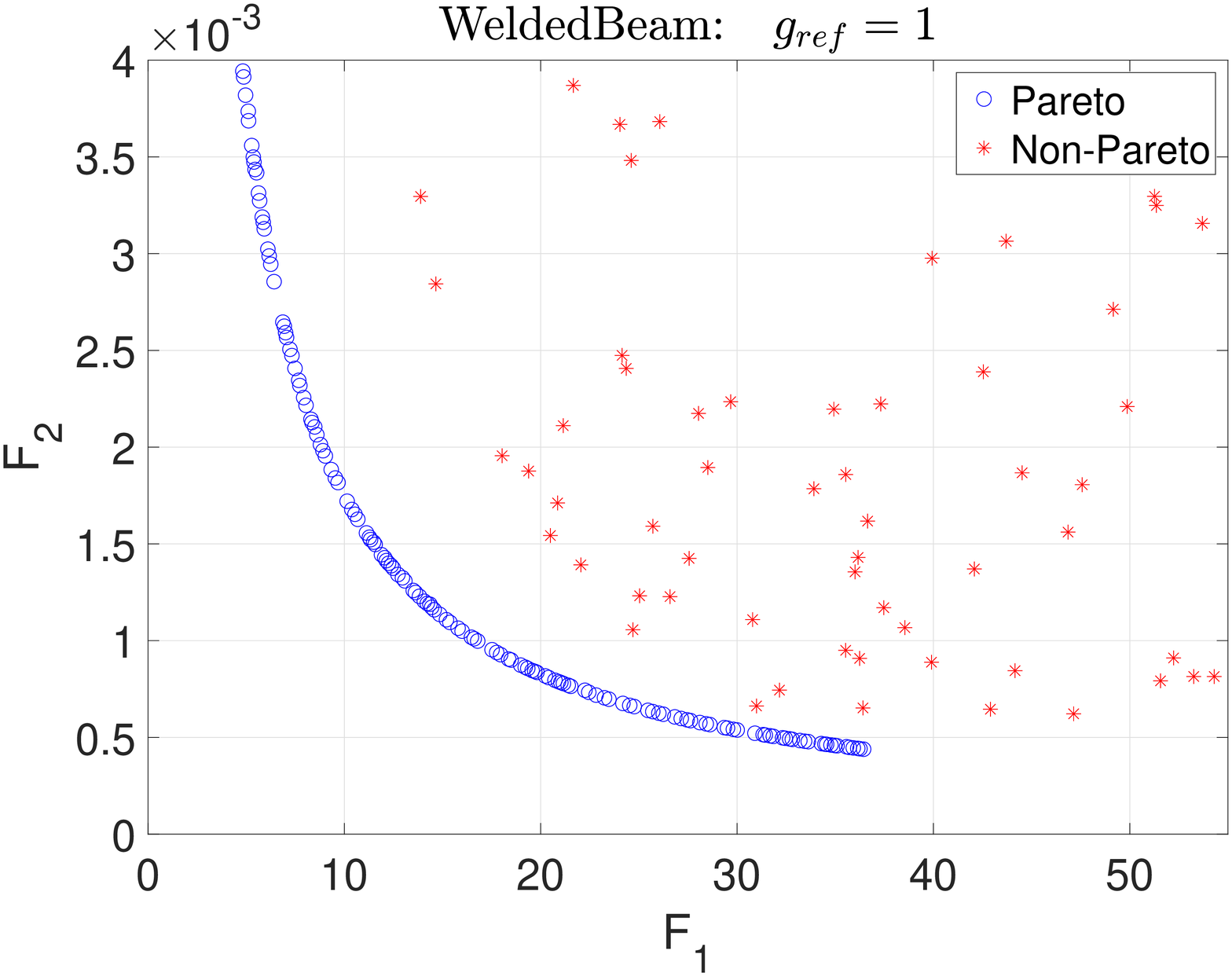}%
    \caption{F-space plot with $g_{ref} = 1$.}
    \label{fig:welded_beam_g1}
    \end{subfigure}%
    ~
    \begin{subfigure}[b]{0.4\textwidth}
    \centering
    \includegraphics[width = \linewidth]{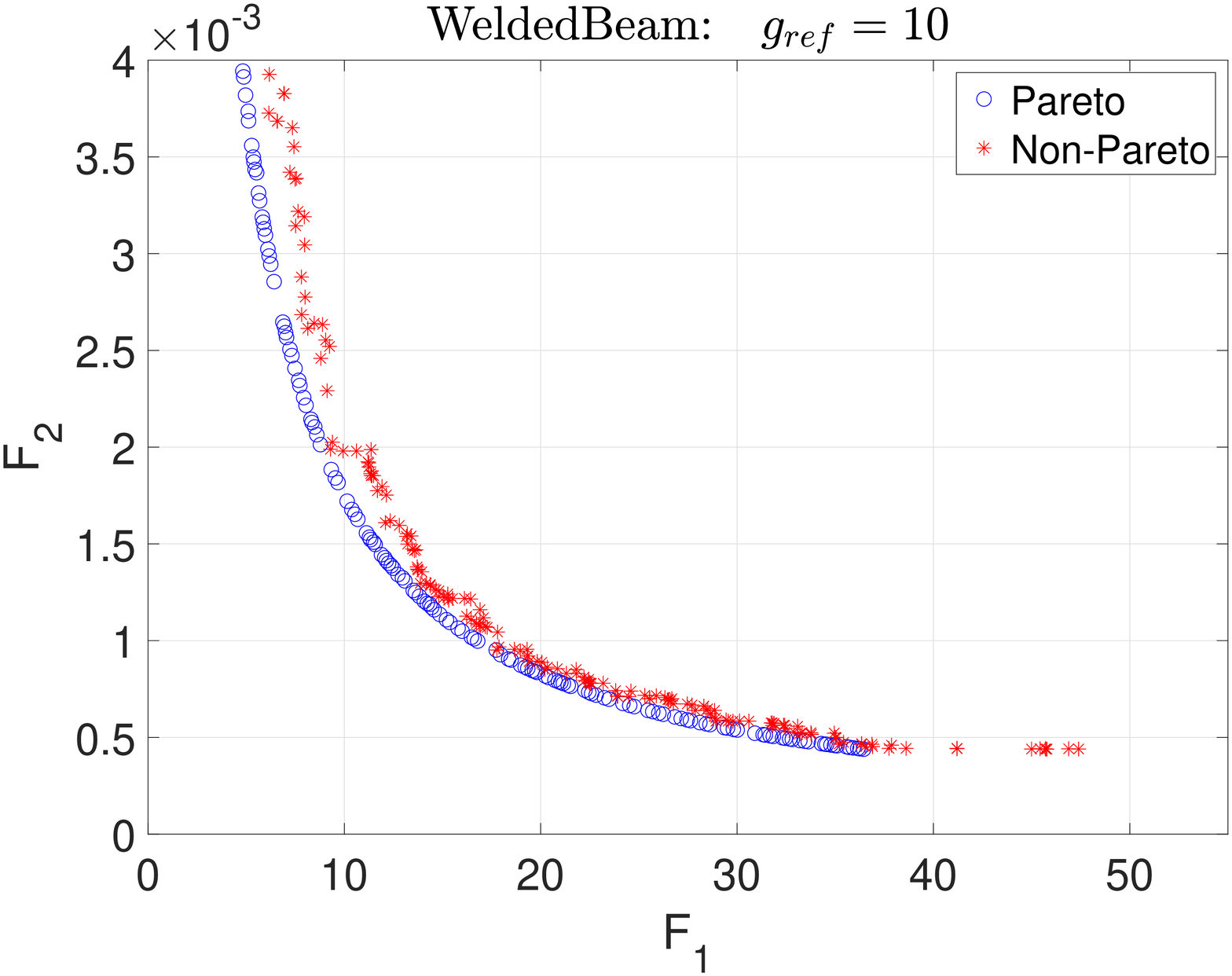}%
    \caption{F-space plot with $g_{ref} = 10$.}
    \label{fig:welded_beam_g10}
    \end{subfigure}%
    \caption{Welded Beam Design Problem data visualization. $\tau_{rank} = 3$ is kept fixed. Problem at (b) is more difficult to solve than at (a).}
    \label{fig:welded_beam_data}
\end{figure*}

Visualization of this is provided in Fig.~\ref{fig:truss_data} and Fig.~\ref{fig:welded_beam_data} for truss and welded-beam problems respectively.

The parameter setting used to create datasets for ZDT1, ZDT2, ZDT3, Truss and Welded-beam problems are shown in Table~\ref{tab:parameter_multi_obj}.
\begin{table}[hbt]
\setcellgapes{2pt}
\makegapedcells
    \centering
    \caption{Parameter Settings for generating datasets for multi-objective problems. For Truss and Welded Beam (WB), different experiments are performed with different values of $g_{ref}$ and $\tau_{rank}$. These values are provided in the caption of result tables.}
    \begin{tabular}{|c|c|c|c|c|c|c|}\hline
       {\bf Problem}  & {\bf Pop-Size} ($N$) & $g_{max}$ & $g_{ref}$ & $\tau_{rank}$ & $N_A$ & $N_B$ \\\hline
         ZDT-1 & 100 & 200 & 200 & 10 & 100 & 5000\\\hline
         ZDT-2 & 100 & 200 & 200 & 5 & 100 & 5000\\\hline
         ZDT-3 & 100 & 200 & 200 & 10 & 100 & 5000\\\hline
         Truss-2D & 200 & 1000 & 1 & -- & 200 & 200\\\hline
         WB & 200 & 1000 & -- & 3 & 200 & 200\\\hline
    \end{tabular}
    
    \label{tab:parameter_multi_obj}
\end{table}

\rebuttalii{
\subsection{m-ZDT, m-DTLZ Problems}
These are the modified versions of ZDT and DLTZ problems \cite{deb2002fast, deb2002scalable}. For m-ZDT problems, the \emph{g-function} is modified to the following:}

\begin{align}
    g_{zdt}(x_2,\ldots,x_n) = & 1 + \frac{18}{n-1} \sum_{\mbox{$i=2$ and even}}^n(x_i + x_{i+1} - 1)^2.  
\end{align}
\rebuttalii{Many two-variable relationships ($x_i + x_{i+1} = 1$) must be set to be on the Pareto set.} 

\rebuttalii{For m-DTLZ problems, the \emph{g-function} for $\mathbf{x_m}$ variables is}

\begin{align}
    g_{dtlz}(\mathbf{x_m}) = & 100 \times \sum_{\mbox{$x_i\in \mathbf{x_m}$ and $i$ is even}}^n(x_i + x_{i+1} - 1)^2.
\end{align}

\rebuttalii{Pareto points for m-ZDT and m-DTLZ problems are generated by using the exact analytical expression of Pareto-set (locations where $g = 0$). The non-Pareto set is generated by using two parameters $\sigma_{spread}$ and $\sigma_{offset}$. To compute the location of a point $\mathbf{x_{np}}$ belonging to non-Pareto set from the location of the  point $\mathbf{x_p}$ on the Pareto set, following equation is used:}

\begin{align}
    x_{np}^{(i)} = & x_p^{(i)} + r_1(\sigma_{offset} + r_2\sigma_{spread})\\
    \text{where ~} & r_1 \in {-1,1}, \quad r_2 \in [0,1].
    \label{eq:m_zdt_dtlz_np}
\end{align}

\rebuttalii{Here, $r_1$ and $r_2$ are randomly generated for each $x_{np}^{(i)}$. For m-ZDT problems, $i = 1, 2, \dots n$, while for m-DTLZ problems $x_{np}^{(i)}, x_p^{(i)} \in \mathbf{x_m}$. Parameter setting for generating m-ZDT and m-DTLZ datasets is provided in Table~\ref{tab:m_zdt_dtlz_params}. For 30 variable problems, all $x_i \in \mathbf{x_m}$ are changed according to Eq.~\ref{eq:m_zdt_dtlz_np} for generate non-pareto points. However, for 500 vars problems, only first 28 of variables in $\mathbf{x_m}$ are modified according to Eq.~\ref{eq:m_zdt_dtlz_np} to generate non-pareto data-points.}

\begin{table}[hbtp]
\caption{\rebuttalii{Parameter setting to generate datasets for m-ZDT and m-DTLZ problems. We generate 1000 datapoints for each class.}}
    \centering
    \begin{tabular}{|c|c|c|c|}\hline
       \textbf{Prob.}  &  $\mathbf{n_{vars}}$ & $\mathbf{\sigma_{spead}}$ & $\mathbf{\sigma_{offset}}$\\\hline
       m-ZDT1 & 30 & 0.3 & 0.1 \\\hline
       m-ZDT1 & 500 & 0.3 & 0.1 \\\hline
       m-ZDT2 & 30 & 0.3 & 0.1 \\\hline
       m-ZDT2 & 500 & 0.3 & 0.1 \\\hline
       
       m-DTLZ1 & 30 & 0.05 & 0 \\\hline
       m-DTLZ1 & 500 & 0.05 & 0 \\\hline
       m-DTLZ2 & 30 & 0.05 & 0 \\\hline
       m-DTLZ2 & 500 & 0.05 & 0 \\\hline
       
    \end{tabular}
    \label{tab:m_zdt_dtlz_params}
\end{table}

\begin{figure*}[hbtp]
    \centering
    \begin{subfigure}[b]{0.45\textwidth}
    \includegraphics[width = \textwidth]{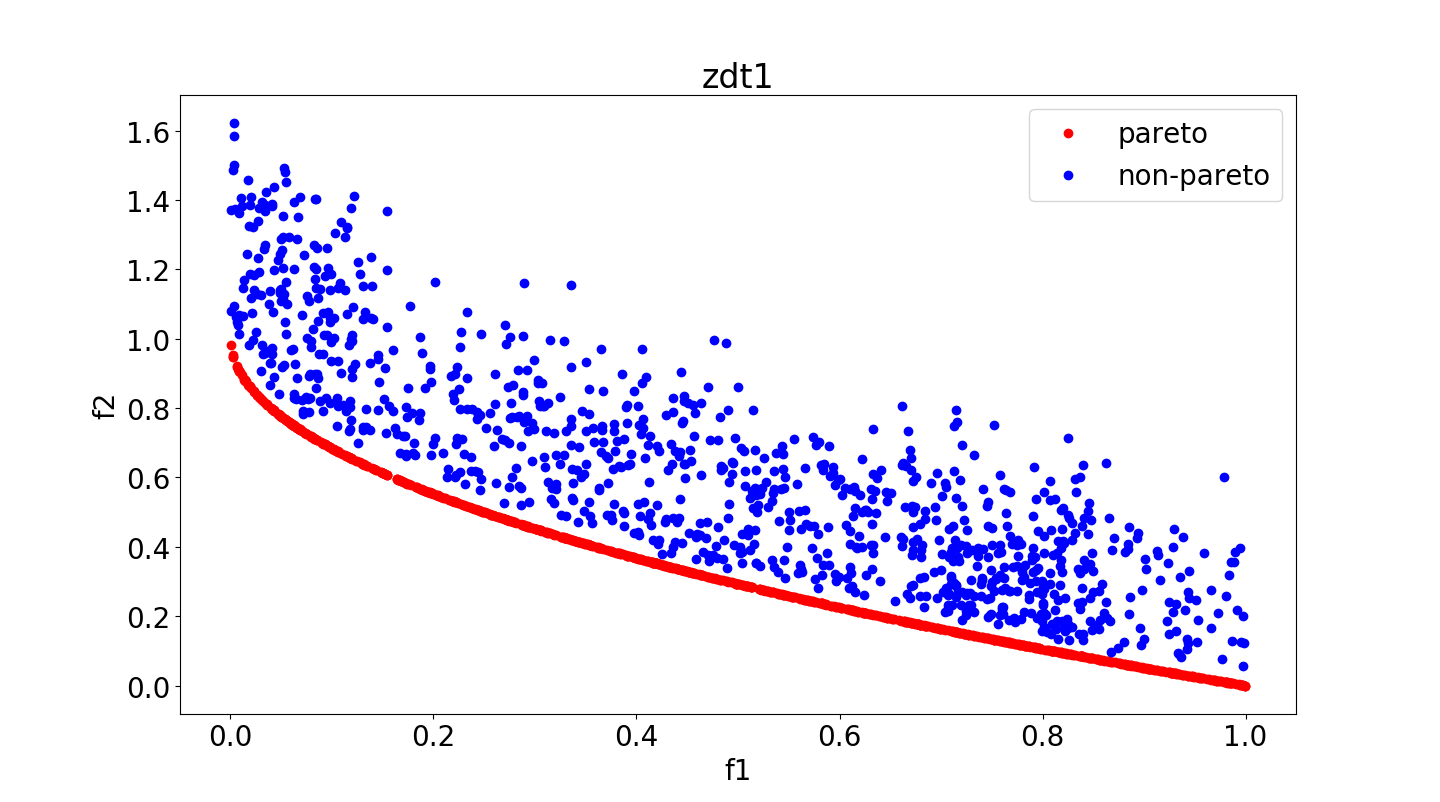}
    \caption{\rebuttalii{m-ZDT1, 30 vars}}
    \end{subfigure}
    ~
    \begin{subfigure}[b]{0.45\textwidth}
    \includegraphics[width = \textwidth]{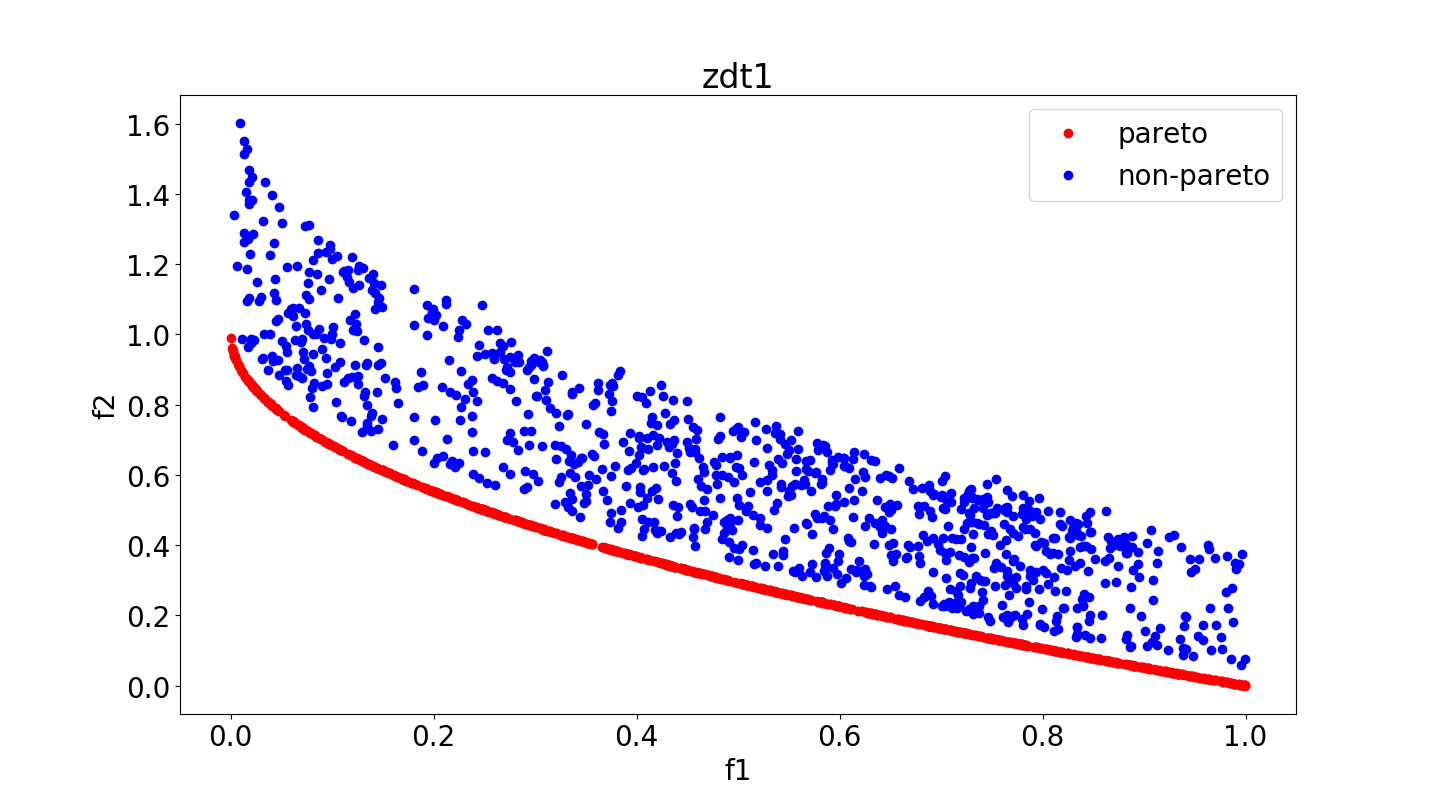}
    \caption{\rebuttalii{m-ZDT1, 500 vars}}
    \end{subfigure}
    \\
    \begin{subfigure}[b]{0.45\textwidth}
    \includegraphics[width = \textwidth]{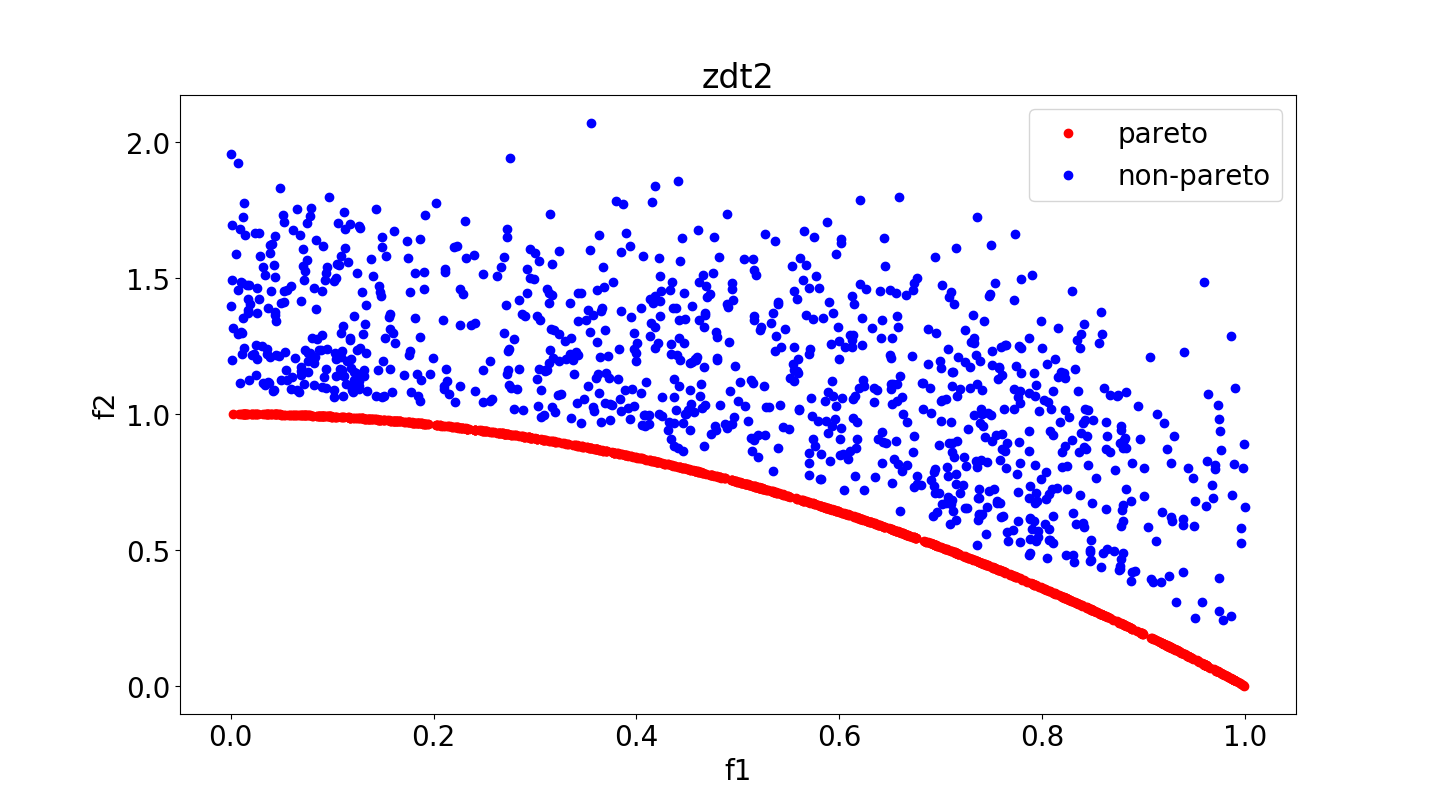}
    \caption{\rebuttalii{m-ZDT2, 30 vars}}
    \end{subfigure}
    ~
    \begin{subfigure}[b]{0.45\textwidth}
    \includegraphics[width = \textwidth]{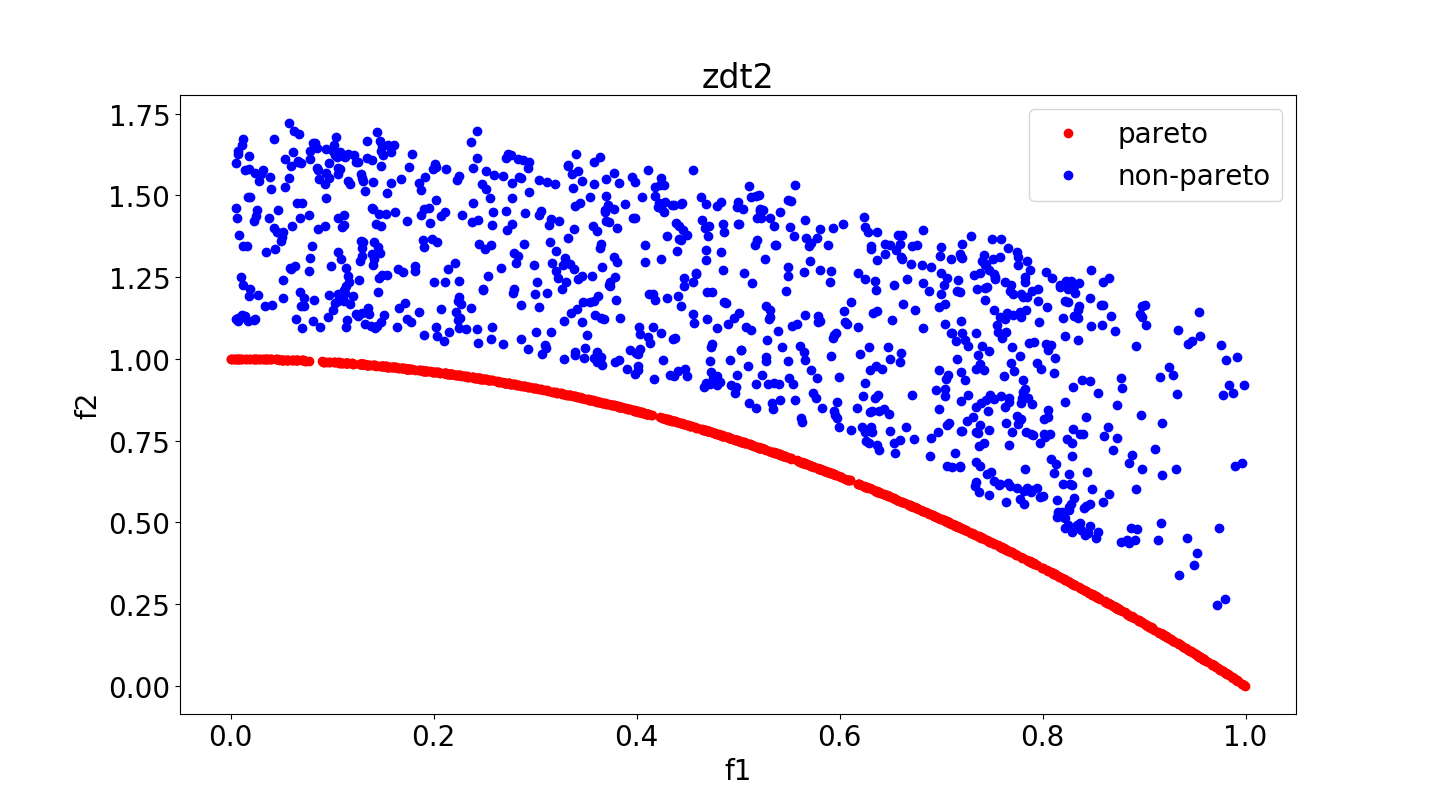}
    \caption{\rebuttalii{m-ZDT2, 500 vars}}
    \end{subfigure}
    \caption{\rebuttalii{m-ZDT Datasets.}}
    \label{fig:m_zdt_datasets}
\end{figure*}

\begin{figure*}[hbtp]
    \centering
    \begin{subfigure}[b]{0.45\textwidth}
    \includegraphics[width = \textwidth]{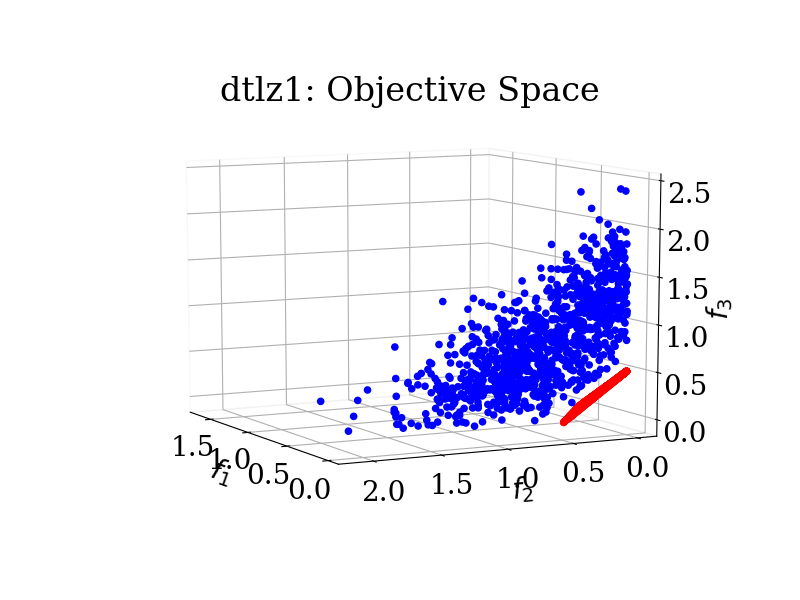}
    \caption{\rebuttalii{m-DTLZ1, 30 vars}}
    \end{subfigure}
    ~
    \begin{subfigure}[b]{0.45\textwidth}
    \includegraphics[width = \textwidth]{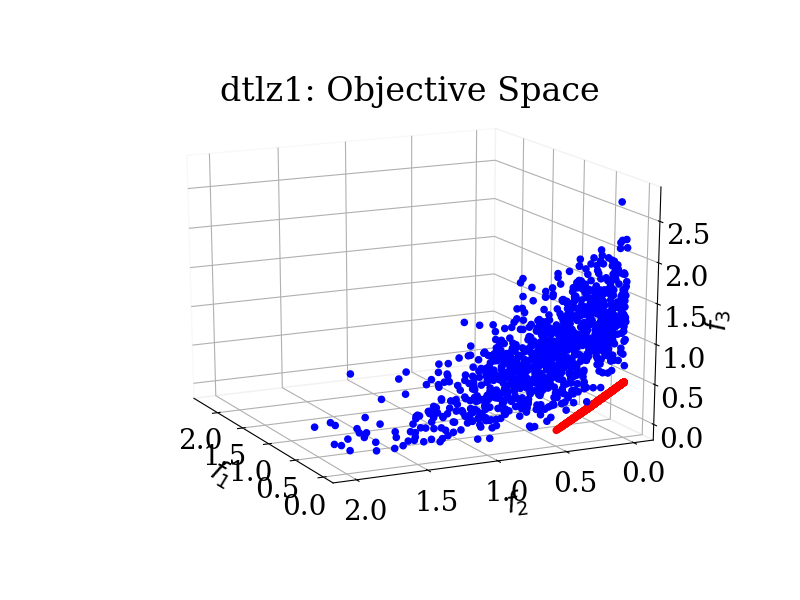}
    \caption{\rebuttalii{m-DTLZ1, 500 vars}}
    \end{subfigure}
    \\
    \begin{subfigure}[b]{0.45\textwidth}
    \includegraphics[width = \textwidth]{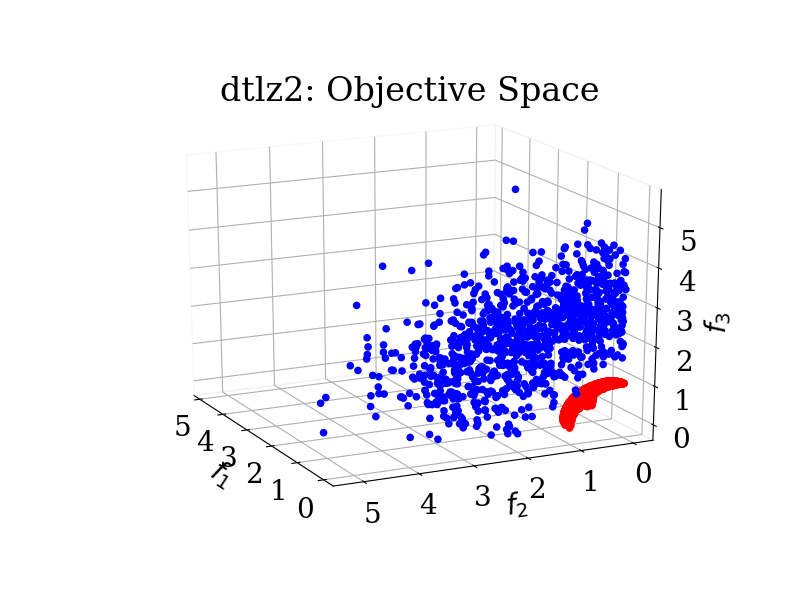}
    \caption{\rebuttalii{m-DTLZ2, 30 vars}}
    \end{subfigure}
    ~
    \begin{subfigure}[b]{0.45\textwidth}
    \includegraphics[width = \textwidth]{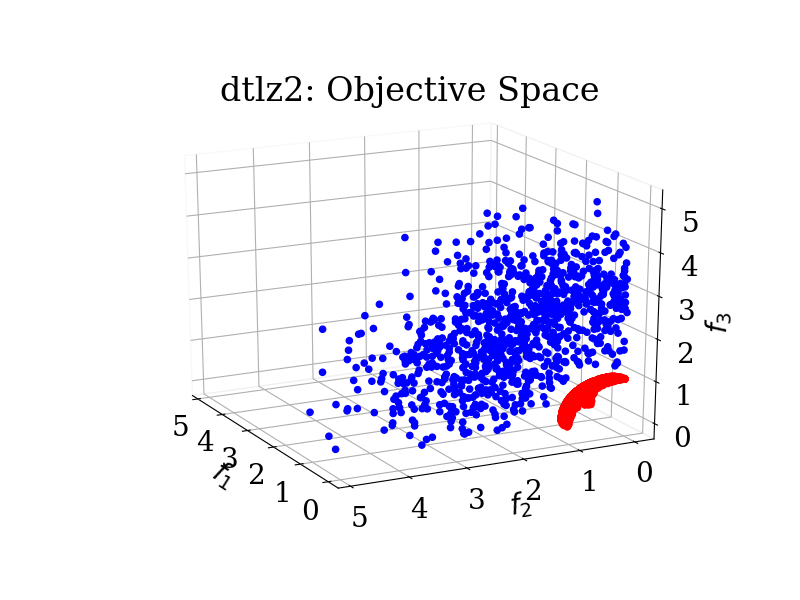}
    \caption{\rebuttalii{m-DTLZ2, 500 vars}}
    \end{subfigure}
    \caption{\rebuttalii{m-DTLZ Datasets.}}
    \label{fig:m_dtlz_datasets}
\end{figure*}

\rebuttalii{Visualization of Datasets for m-ZDT and m-DTLZ problems in provided in Figure~\ref{fig:m_zdt_datasets} and Figure~\ref{fig:m_dtlz_datasets}.}

\section{Duplicate Update Operator for Upper Level GA}
\label{sec:duplicate_repair}
After creating the offspring population, we attempt to eliminate duplicate population members. For each duplicate found in the child population, the block-matrix $\mathbf{B}$ of that individual is randomly mutated using the following equation
\begin{equation}
    \mathbf{B}(i_r,j_r) = \mathbf{E}(k_r),
\end{equation}
where $i_r, j_r$ and $k_r$ are randomly chosen integers within specified limits. 
This process is repeated for all members of child population, until each member is unique within the child population. This operation allows to maintain diversity and encourages the search to generate multiple novel and optimized equations of the split-rule.

\section{Ablation Studies and Comparison}
\label{sec:ablation_studies}
In this section, we test the efficacy of lower-level and upper-level optimization algorithm by applying them on four customized problems (DS1-DS4) which are shown pictorially in Fig.~\ref{fig:customized_datasets}. Their performances are compared against two standard classifiers: CART and support-vector-machines (SVM)\footnote{Here, the support vector machine is applied without any kernal-trick.}.

\subsection{Ablation Studies on Lower Level GA}
Since the lower level GA (LLGA) focuses at determining coefficients $w_i$ and bias $\theta_i$ of linear split-rule in \emph{B-space}, DS1 and DS2 data-sets are used to guage its efficacy. 
The block matrix $\mathbf{B}$ is fixed to $2\times2$ identity matrix and the modulus-flag $m$ is set to 0. Hence, the split-rule (classifier) to be evolved is 
\[f(\mathbf{x}) = \theta_1 + w_1x_1 + w_2x_2,\]
where $w_1, w_2$ and $\theta_1$ are to be determined by the LLGA.

\begin{figure}[hbt]
    \centering
    \begin{subfigure}{0.48\linewidth}
        \centering
        \includegraphics[width = \textwidth]{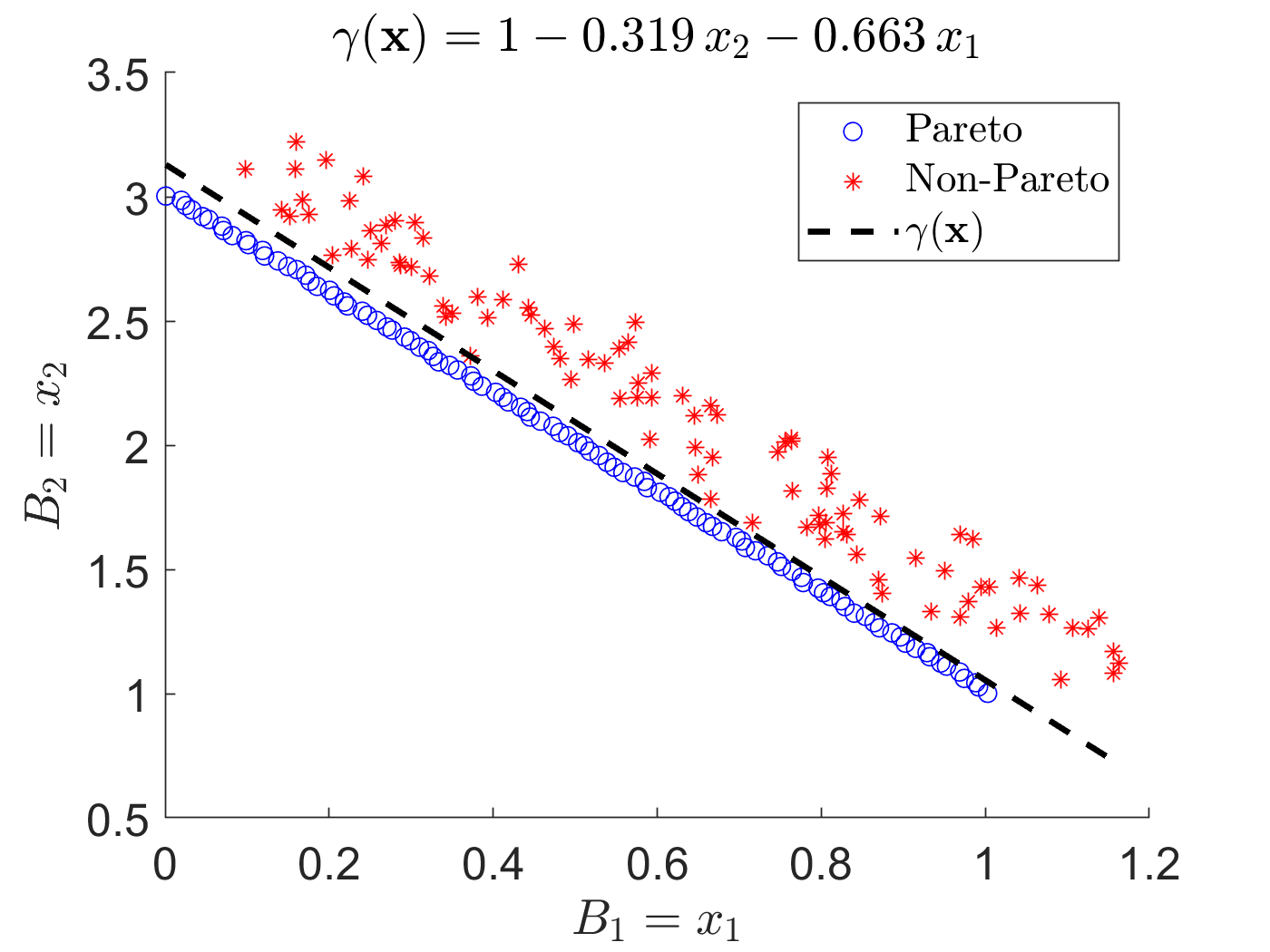}
        \caption{LLGA: $(0\%, 0\%)$; $(0\%, 1\%)$}
    \end{subfigure}%
    \hfill
    \begin{subfigure}{0.48\linewidth}
        \centering
        \includegraphics[width = \textwidth]{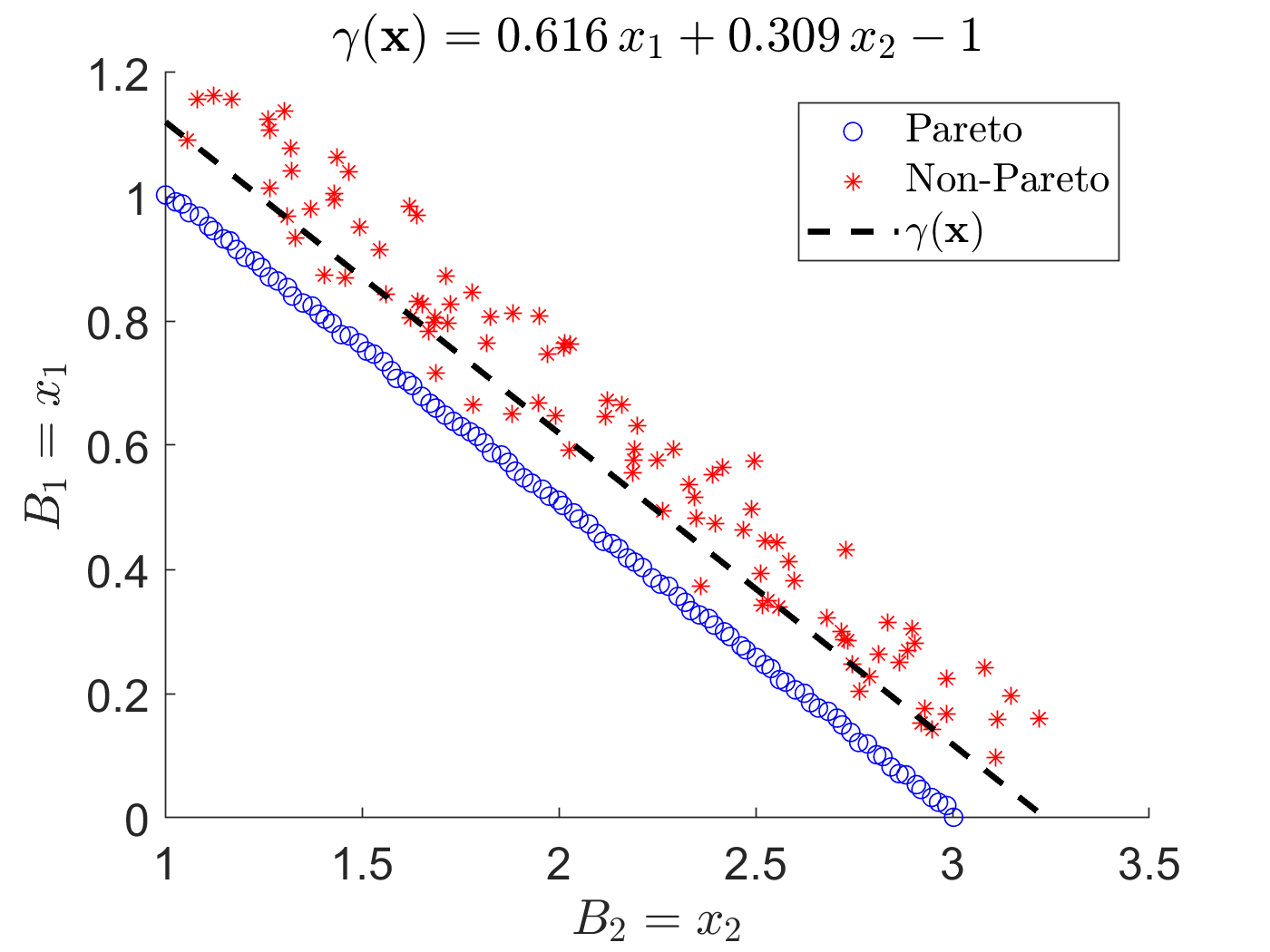}
        \caption{SVM: $(0\%, 17\%)$; $(0\%, 10\%)$}
    \end{subfigure}
    \caption{Results on DS1 dataset to benchmark LLGA. Numbers in first parenthesis indicate \emph{Type-1} and \emph{Type-2} error on training data and the numbers in second parenthesis indicate \emph{Type-1} and \emph{Type-2} error on testing data.}
    \label{fig:results_DS1_lower_level}
\end{figure}

\begin{figure*}[hbt]
    \centering
    \begin{subfigure}{0.33\textwidth}
        \centering
        \includegraphics[width = \textwidth]{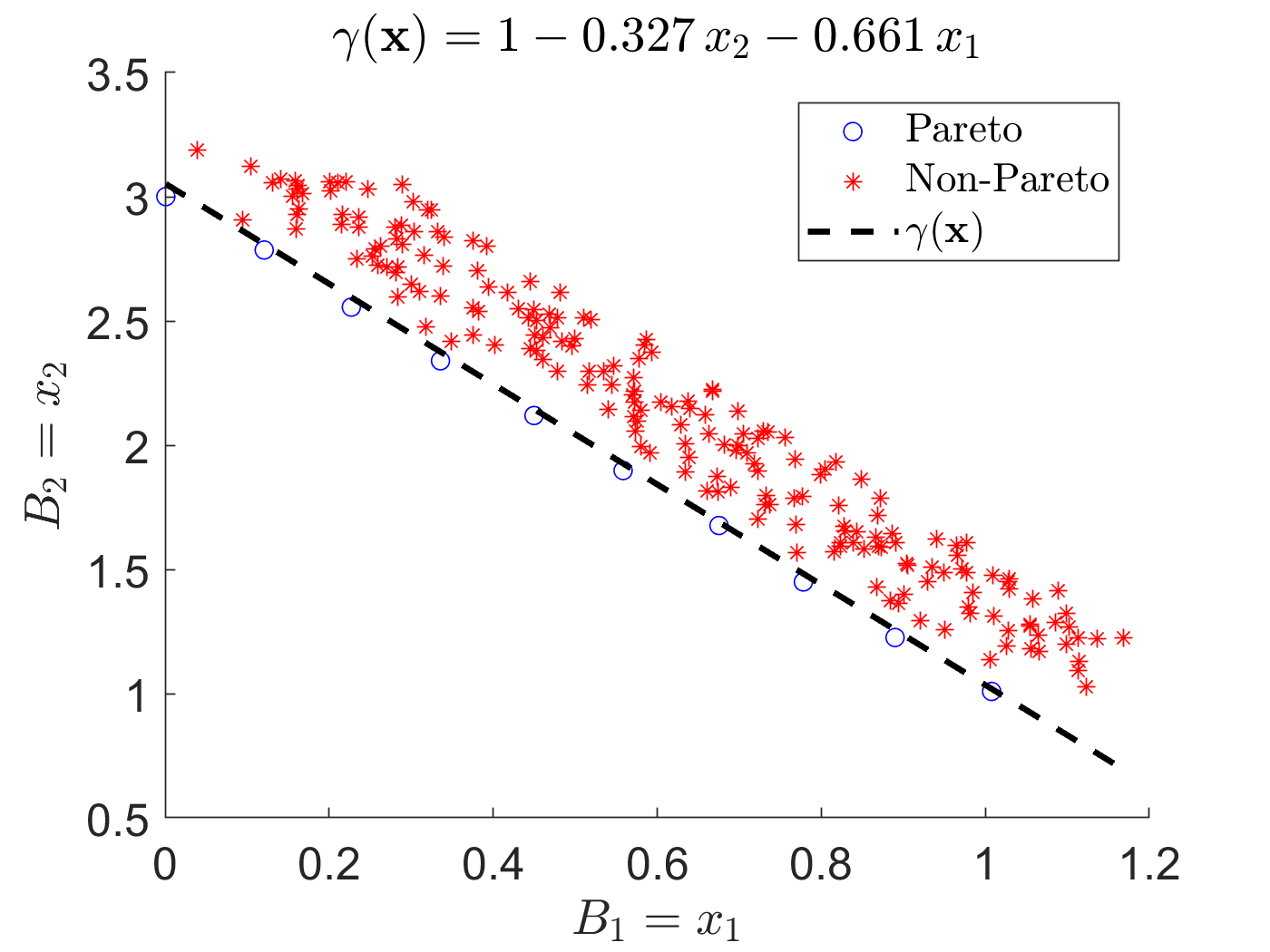}
        \caption{LLGA : $(0\%, 0\%)$; $(0\%,0\%)$}
        \label{fig:DS2_rga}
    \end{subfigure}%
   \hfill
    \begin{subfigure}{0.33\textwidth}
        \centering
        \includegraphics[width = \textwidth]{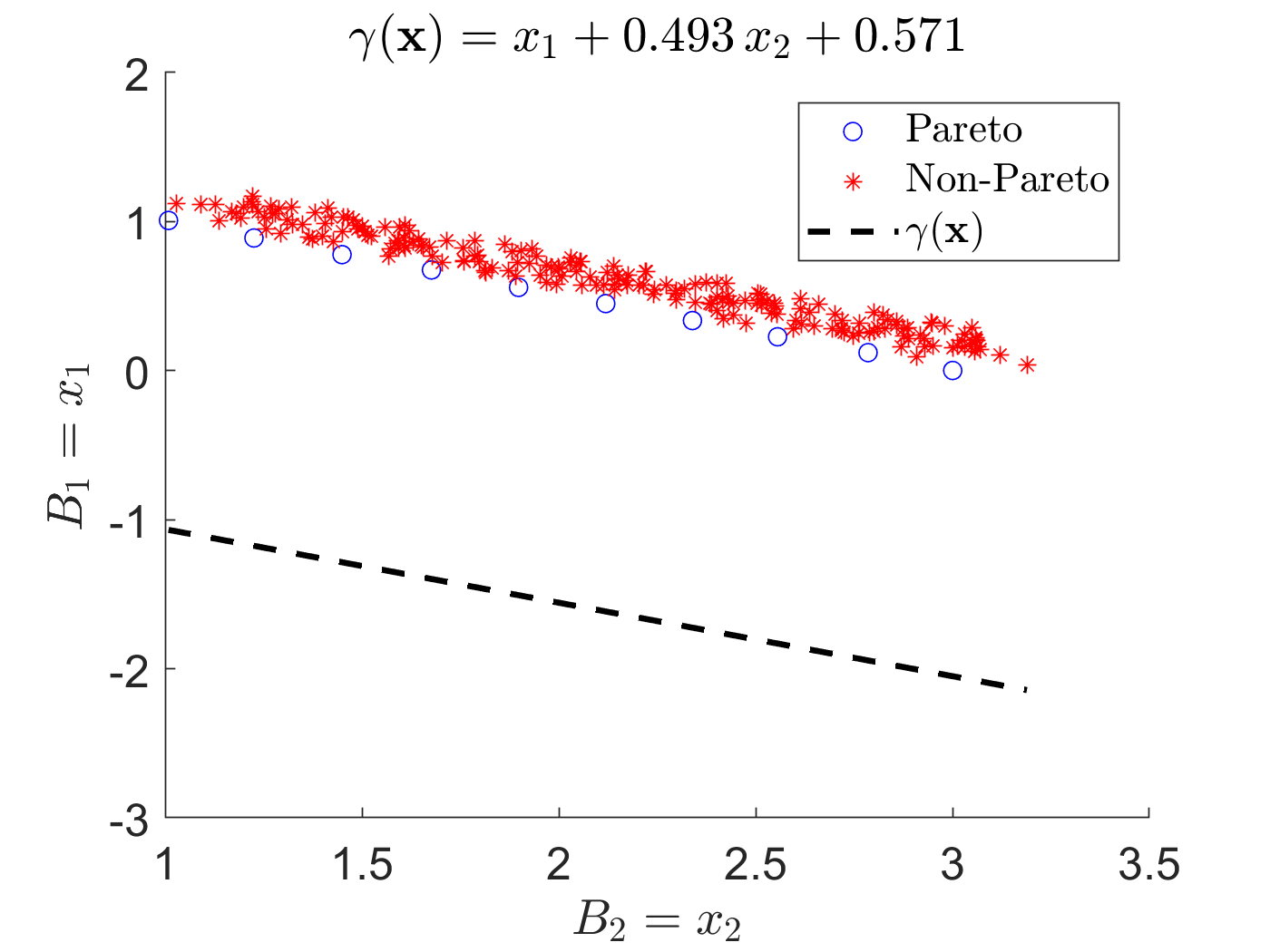}
        \caption{SVM: $(100\%,0\%)$; $(100\%,0\%)$}
        \label{fig:DS2_svm_only}
    \end{subfigure}%
   \hfill
    \begin{subfigure}{0.33\textwidth}
        \centering
        \includegraphics[width = \textwidth]{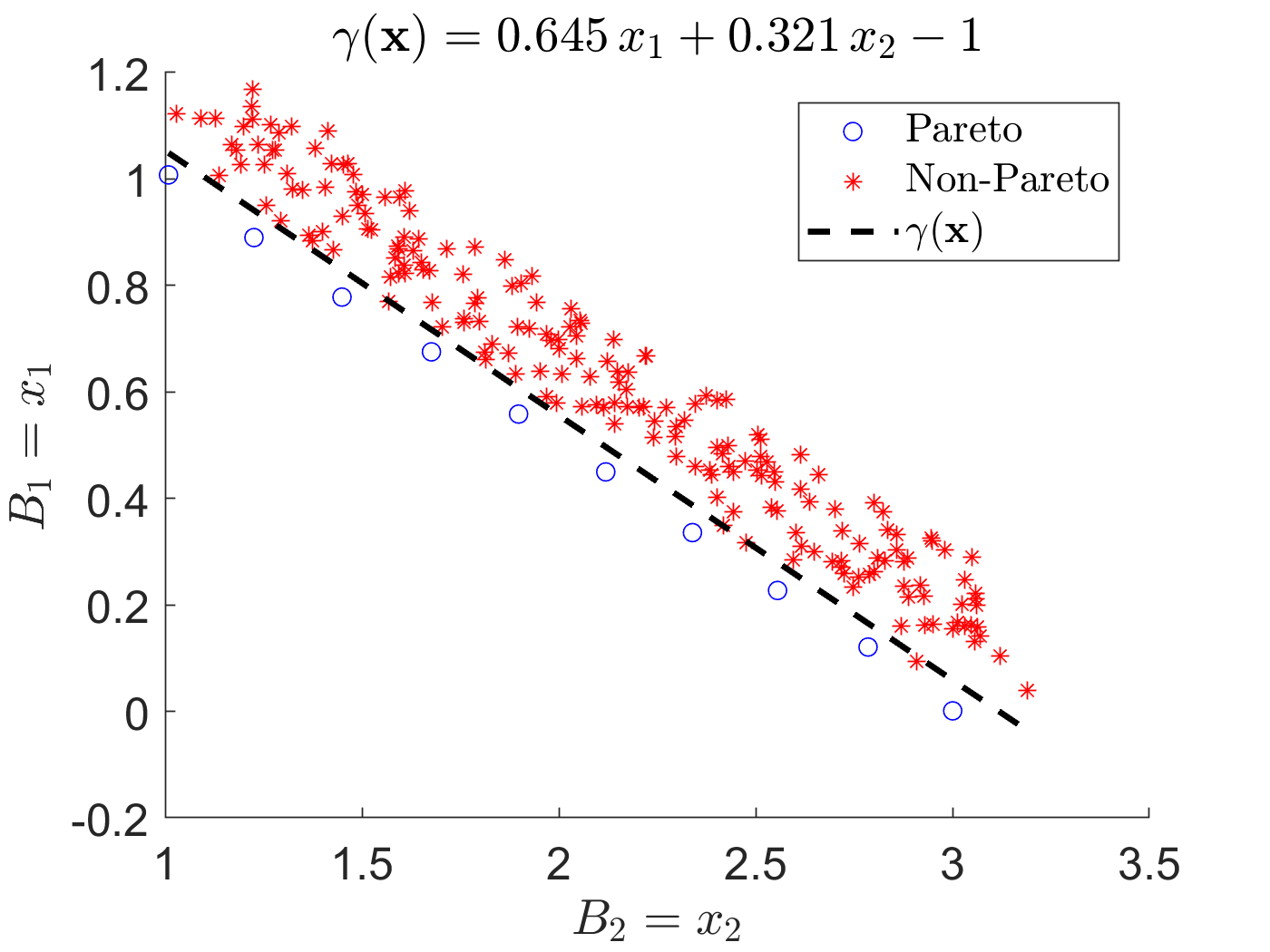}
        \caption{SVM$+$SMOTE: $(0\%,1.5\%)$; $(0\%,1.5\%)$}
        \label{fig:DS2_svm_smote}
    \end{subfigure}
    \caption{Results on DS2 data-set to benchmark LLGA.}
    \label{fig:results_DS2_lower_level}
\end{figure*}

The classifier generated by our LLGA is compared with that obtained using the standard SVM algorithm (without any kernel-trick) on DS1 data-set. Since DS2 dataset is unbalanced, SMOTE algorithm \cite{chawla2002smote} is first applied to over-sample data-points of the minority class before generating the classifier using SVM. 
For the sake of completeness, classifier generated by SVM on DS2 dataset, without any oversampling, is also compared against the ones obtained using LLGA and SMOTE+SVM. The results are shown in Fig.~\ref{fig:results_DS1_lower_level} and Fig.~\ref{fig:results_DS2_lower_level}, respectively, for DS1 and DS2 data-sets. Type-1 and Type-2 errors are reported for each experiment\footnote{Type-1 error indicates the percentage of data-points belonging to \emph{Class-1} getting classified as \emph{Class-2}. Type-2 error indicates the percentage of points belonging to \emph{Class-2} getting classified as \emph{Class-1}).}.

It is clearly evident from the results shown in Fig.~\ref{fig:results_DS1_lower_level} that the proposed customized LLGA is more efficient than SVM in arriving at the desired split-rule as a classifier. The LLGA is able to find the decision boundary with $100\%$ prediction accuracy on training and testing data-sets. However, the classifier generated by SVM has an overlap with the cloud of data-points belonging to the scattered class and there-by resulting into the accuracy of $89\%$ on the testing data-set.

On the DS2 dataset, the SVM is needed to rely on SMOTE to synthetically generate data-points in order to achieve respectable performance as can be seen from Fig.~\ref{fig:DS2_svm_only} and \ref{fig:DS2_svm_smote}. However, LLGA performs better without SMOTE, as shown in Fig.~\ref{fig:DS2_rga}. 

Ablation study conducted above on LLGA clearly indicates that the customized optimization algorithm developed for estimating weights $\mathbf{w}$ and bias $\mathbf{\Theta}$ in the expression of a split-rule is reliable and efficient and could easily outperform the classical SVM algorithm.

\subsection{Ablation Studies on Proposed Bilevel GA}
\begin{figure}[hbt]
    \centering
    \begin{subfigure}{0.48\linewidth}
        \hspace*{-3mm}\includegraphics[width = 1.1\textwidth]{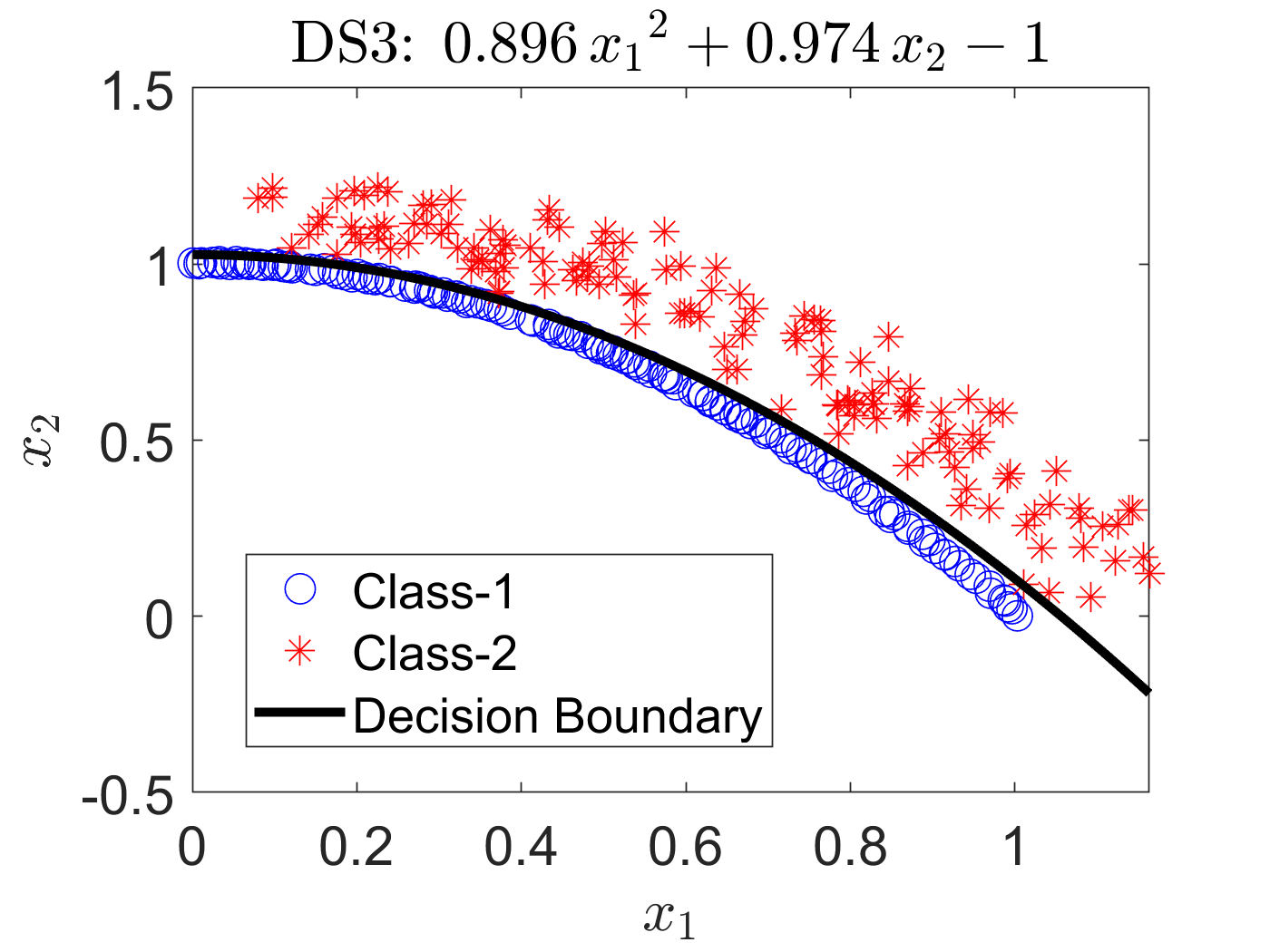}
        \caption{DS3: $(0\%, 0\%)$; $(0\%, 0\%)$}
        \label{fig:DS3_decicion_boundary_plot}
    \end{subfigure}%
    \hfill
    \begin{subfigure}{0.48\linewidth}
        \hspace*{-3mm}\includegraphics[width = 1.1\textwidth]{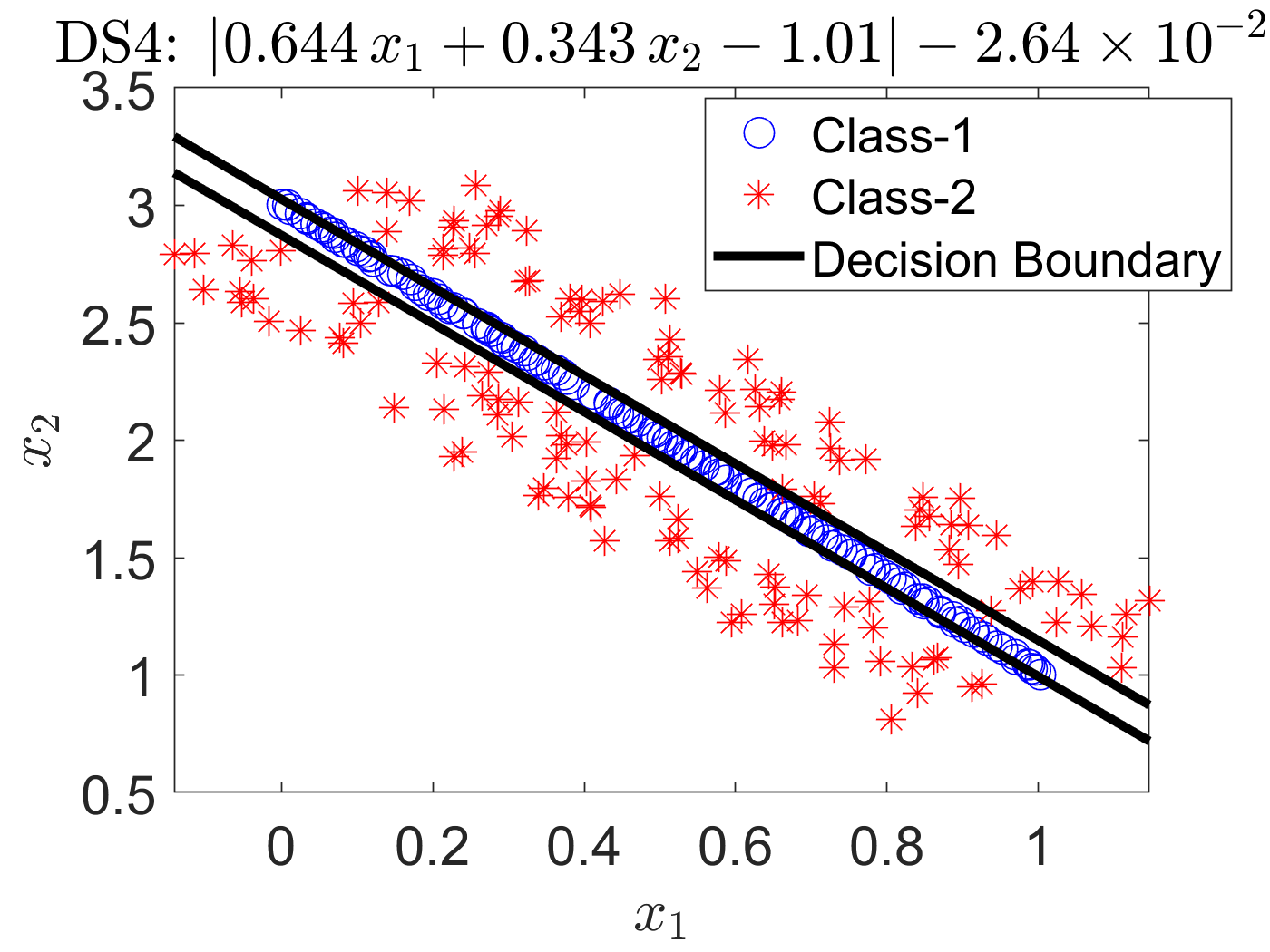}
        \caption{DS4: $(0\%, 0\%)$; $(0\%, 0\%)$}
        \label{fig:DS4_decicion_boundary_plot}
    \end{subfigure}
    \caption{Bilevel GA results on DS3 and DS4.}
    \label{fig:ablation_upper_level}
\end{figure}

As mentioned before, the upper level of our bilevel algorithm aims at determining optimal power-law structures (i.e. $B_i$s) and the value of modulus flag $m$. Experiments are performed on DS3 and DS4 datasets to guage the efficacy of upper level (and thus the overall \emph{bilevel}) GA. No prior information about the optimal values of block matrix $\mathbf{B}$ and modulus flag $m$ is supplied. Thus, the structure of the equation of the split-rule is unknown to the algorithm. 
Results of these experiments are shown in Figs.~\ref{fig:DS3_decicion_boundary_plot} and \ref{fig:DS4_decicion_boundary_plot}.

As can be seen from Fig.~\ref{fig:DS3_decicion_boundary_plot} that the proposed bilevel algorithm is able to find classifiers which are able to classify the data-sets with no Type-I or Type-II error.   
Results on DS4 data-set validated the importance of involving modulus-flag $m$ as a search variable for upper level GA in addition to the block-matrix $\mathbf{B}$. The algorithm is able to converge at the optimal tree topology (Fig.~\ref{fig:DS4_decicion_boundary_plot}) with $100\%$ classification accuracy. 
It is to note here that the algorithm had no prior knowledge about optimal block matrix $\mathbf{B}$ of exponents $b_{ij}$. This observation suggests that the upper level GA is successfully able to navigate through the search space of block-matrices $\mathbf{B}$ and modulus-flag $m$ (which is a binary variable) to arrive at the optimal power-laws and split-rule.

\begin{figure}[hbtp]
    \centering
    \includegraphics[width=0.9\linewidth]{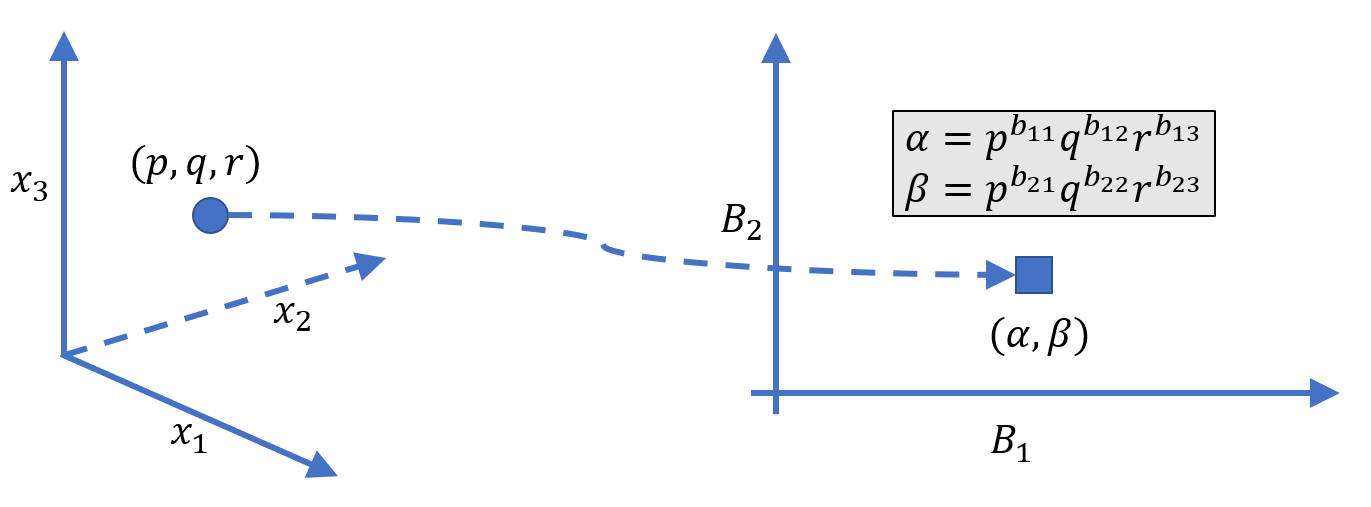}
    \caption{\rebuttalii{Feature Transformation. A point in three-dimensional X-space is mapped to a point in a two-dimensional B-space for which $B_i = x_1^{b_{i1}}x_2^{b_{i2}}x_3^{b_{i3}}$.}}
    \label{fig:feature_transform}
\end{figure}

\rebuttalii{
\subsection{Visualization of Split Rule: X-Space and B-Space}
A comprehensible visualization of the decision boundary is possible if the dimension of the feature space is up to three. However, if the dimension $d$ of the original feature space (or, \emph{X-space}) is larger than three, the decision-boundary can be visualized on the \emph{transformed-feature-space} (or \emph{B-space}). In our experiments, the maximum number of allowable power-law-rules ($p$) is fixed to three (i.e. $p = 3$). Thus, any data-point $\mathbf{x}$ in a $d$-dimensional \emph{X-space} can be mapped to a three-dimensional \emph{B-space}. A conceptual illustration of this mapping is provided in Fig.~\ref{fig:feature_transform},}
\rebuttalii{where, for the sake of simplicity, a three-dimensional \emph{X-space} ($x_1$ to $x_3$) is mapped to a two-dimensional B-space using two power-law-rules ($B_1$ and $B_2$). }

\rebuttalii{It is to note here that the power-law rules $B_i$'s are not known a priori. The proposed bilevel optimization method determines them so that the data becomes linearly separable in B-space.} 

\section{Additional Results}
\begin{figure*}[hbt!]
    \begin{subfigure}[b]{0.45\textwidth}
        \centering
        \includegraphics[width = \textwidth]{figures/ds3_x_space.png}
        \caption{Bilevel GA results: $(0\%,0\%)$ and $(1\%,0\%)$.}
        \label{fig:DS3_overall_plot}
    \end{subfigure}%
   \hfill
   \centering
    \begin{subfigure}[b]{0.45\textwidth}
       \vspace{10pt}
        \centering
        \includegraphics[width = \textwidth]{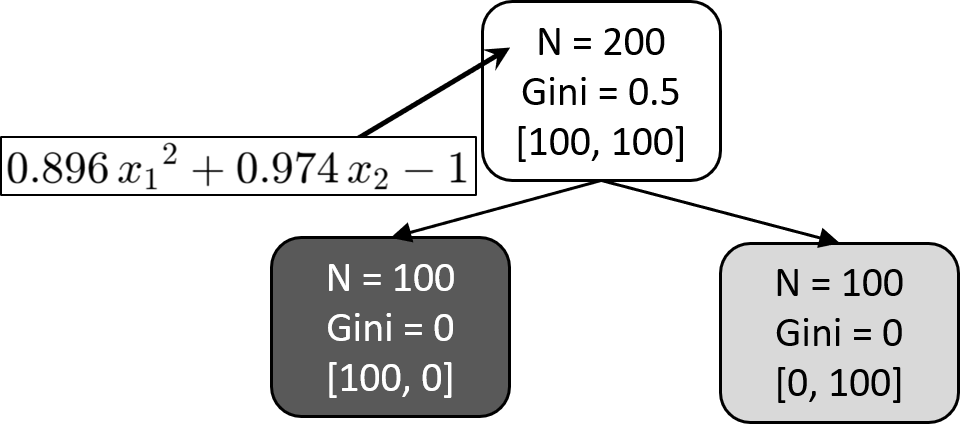}
        \caption{Obtained NLDT for DS3 problem -- One rule.}
        \label{fig:DS3_overall_tree}
    \end{subfigure}%
    \hfill
    \begin{subfigure}{\textwidth}
        \centering
        \vspace{15pt}
        \includegraphics[width = \textwidth]{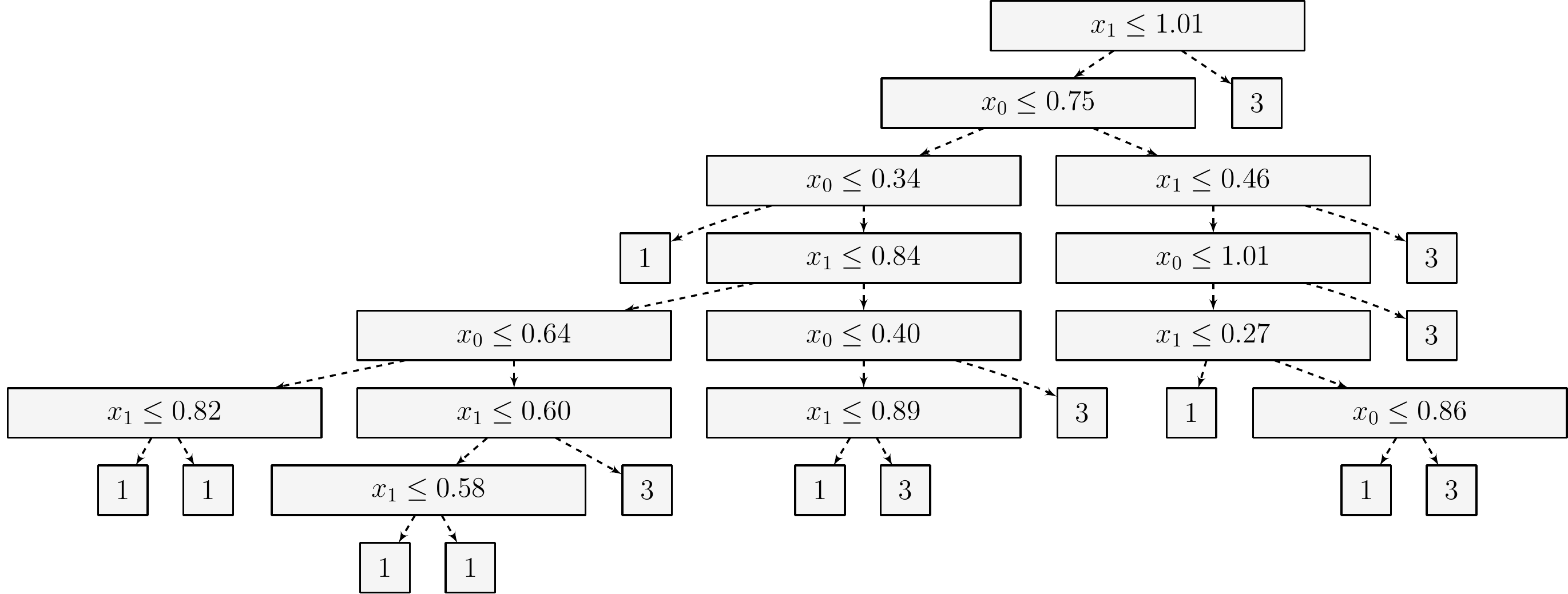}
        \caption{CART results: $(0\%,4\%)$ and $(4\%,3\%)$ -- 13 rules.}
        \label{fig:DS3_CART_tree}
    \end{subfigure}%
    \caption{Results on DS3 dataset to benchmark the overall bilevel algorithm. {\color{black}{CART is difficult to interpret, while NLDT classifier is easy to interpret.}}}
    \label{fig:result_DS3_overall}
\end{figure*}

\begin{figure*}[hbt!]
    \centering
        \begin{subfigure}[b]{0.45\textwidth}
        \centering
        \includegraphics[width = \textwidth]{figures/ds4_x_space.png}
        \caption{Bilevel GA results: $(0\%,0\%)$ and $(0\%,0\%)$ -- No error.}
        \label{fig:DS4_bilevel_plot}
    \end{subfigure}%
    \hfill
    \begin{subfigure}[b]{0.45\textwidth}
        \centering
        \vspace{10pt}
        \includegraphics[width = \textwidth]{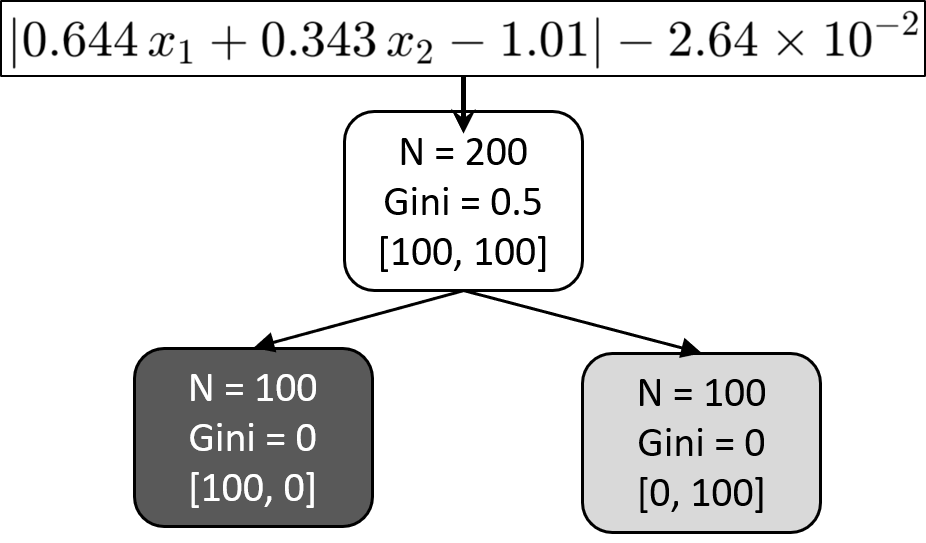}
        \caption{Obtained NLDT for DS4 problem -- One rule.}
        \label{fig:DS4_bilevel_tree}
    \end{subfigure}%
    \hfill
    \begin{subfigure}{\textwidth}
        \centering
        \vspace{17pt}
        \includegraphics[width = \textwidth]{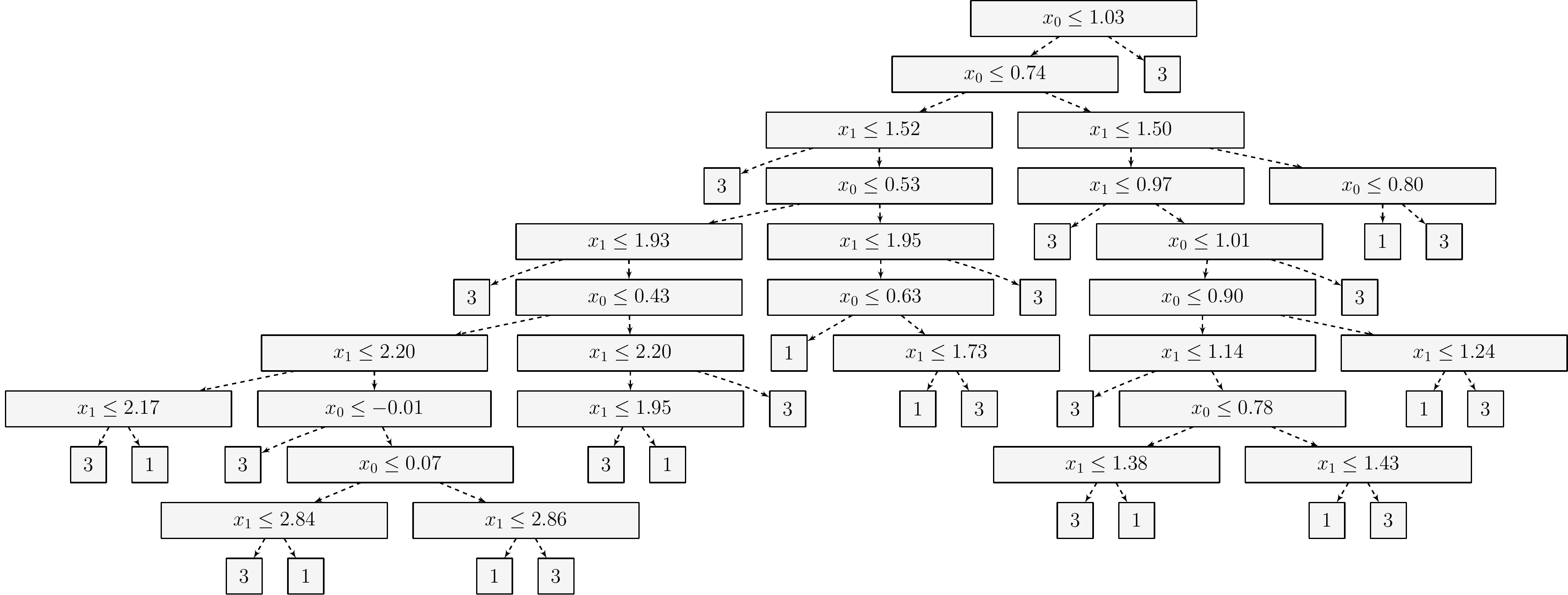}
        \caption{CART results: $(4\%,3\%)$ and $(5\%, 18\%)$, 27 rules.}
        \label{fig:Ds4_CART_tree}
    \end{subfigure}%
    \caption{Results on DS4 dataset to benchmark the overall bilevel algorithm. {\color{black}{CART is difficult to interpret, while NLDT classifier is easy to interpret.}}}
    \label{fig:result_DS4_overall}
\end{figure*}

Here we provide some additional results. Fig.~\ref{fig:result_DS3_overall} and Fig.~\ref{fig:result_DS4_overall} shows plots of one of the decision boundary (classifier) obtained using our bilevel-algorithm and its comparison against decision-tree generated using traditional CART algorithm on \emph{DS3} and \emph{DS4} datasets.

\subsection{2D Truss Problem}
In table~\ref{tab:Truss_rank6_Vs_rank9}, comparison of accuracy-scores and complexity obtained for datasets generated using $\tau_{rank} = 6$ and $\tau_{rank} = 9$ for bi-objective truss design problem are provided.

\begin{table*}[hbt]
\setcellgapes{2pt}
\makegapedcells
    \centering
    \caption{2D Truss problem with $\tau_{rank}=6$ and $\tau_{rank}=9$.}
    \begin{tabular}{|c|c|c|c|c|c|}\hline
    $\tau_{rank}$  &  \textbf{Training Accuracy} & \textbf{Testing Accuracy} & \textbf{\#Rules} & \textbf{n-vars} & \textbf{${\bf F_U}$/Rule}\\\hline
    6 &  $ 99.86 \pm 0.36 $ & $ 99.6 \pm 0.58 $ & $ 1.20 \pm 0.53 $ & $ 2.78 \pm 0.41 $ & $ 2.92 \pm 0.52 $ \\\hline
    9 &  $ 100\pm 0$ & $ 99.92 \pm 0.19 $ & $ 1.36 \pm 0.48 $ & $ 2.14 \pm 0.35 $ & $ 2.26 \pm 0.52 $\\\hline
    \end{tabular}
    \label{tab:Truss_rank6_Vs_rank9}
\end{table*}

\rebuttalii{Since the dataset generated with $\tau_{rank} = 6$ is relatively difficult, we conduct further analysis on it and compare the result obtained using our NLDT approach with that of CART and SVM. Compilation of these results is shown in Table~\ref{tab:truss2d_6_big}}.

\begin{table*}[hbt]
\setcellgapes{2pt}
\makegapedcells
    \centering
    \caption{\rebuttalii{Results on Truss-2D with $\tau_{rank}=6$ and $g_{ref} = 1$.}}
    >{\color{black}\begin{tabular}{|c|c|c|cHH|HHc|Hc|c|}\hline
     \multirow{2}{*}{\textbf{Method}}  &  \multirow{2}{*}{\textbf{Training Accuracy}} & \multirow{2}{*}{\textbf{Testing Accuracy}} & \multirow{2}{*}{\textbf{p-value}} & \textbf{Training} & \textbf{Training} & \textbf{Testing Error} & \textbf{Testing Error} & \multirow{2}{*}{\textbf{\#Rules}} & \multirow{2}{*}{\textbf{n-vars}} & \multirow{2}{*}{\textbf{${\bf F_U}$/Rule}}  & \multirow{2}{*}{\textbf{Rule Length}}\\
    &  &  &  & {\bf Type 1} & {\bf Type 2} & {\bf Type 1} & {\bf Type 2} & & & & \\\hline
    \rebuttalii{NLDT} & $99.77 \pm 0.72 $ & ${\bf 99.54} \pm {\bf 0.75}$ & -- & $0.18 \pm 0.90 $ & $0.29 \pm 0.97 $ & $0.30 \pm 1.09 $ & $0.62 \pm 0.95$ & $1.2 \pm 0.5$ & $2.8 \pm 0.4$ & $2.9 \pm 0.5$ & ${\bf 3.3} \pm {\bf 0.9}$\\\hline

    CART & $99.34 \pm 0.32 $ & $98.33 \pm 1.10 $ & 6.04e-08 & $0.58 \pm 0.49 $ & $0.75 \pm 0.60 $ & $1.87 \pm 1.41 $ & $1.48 \pm 1.47$ & $11.06 \pm 3.15$ & $1$ & ${\bf 1}$ & $11.06 \pm 3.15$\\\hline
    SVM & $99.66 \pm 0.15 $ & $\mathit{99.46 \pm 0.50} $ & {\it 0.135} & $0\pm 0$ & $0.68 \pm 0.30 $ & $0\pm 0$ & $1.08 \pm 1.00$ & ${\bf 1}$ & $3$ & $62.5 \pm 2.9$ & $62.5 \pm 2.9$\\\hline
    \end{tabular}}
    \label{tab:truss2d_6_big}
\end{table*}

\subsection{Welded Beam Design Resutls}
\rebuttalii{Additional results for Welded beam design problem are shown in Table~\ref{tab:welded_beam_gref_1_big}. Here, two sets of experiments are conducted for two different values of $g_{ref}$, keeping the value of $\tau_{rank}$ fixed to three. A higher value of $g_{ref}$ results in good and bad data-points too close to each other, thereby making it difficult for the classification algorithm to determine a suitable classifier. This can be validated from the results presented in Table~\ref{tab:welded_beam_gref_1_big}. However, bilevel NLDTs have produced one rule with about 2 to 4 variable appearances compared to 39.5 to 126 (on average) for SVM classifiers. CART does well in this problem with, on average, 2.94 to 8.42 rules.
}

\begin{table*}[hbt]
\setcellgapes{2pt}
\makegapedcells
    \centering
    \caption{\rebuttalii{Results on welded beam design with $g_{ref}= 1$ and $g_{ref}= 10$. $\tau_{rank}=3$ is kept fixed.}}
    >{\color{black}
    \begin{tabular}{|c|c|c|c|cHH|HHc|Hc|c|}\hline
    \multirow{1}{*}{\textbf{$g_{ref}$}} &
     \multirow{1}{*}{\textbf{Method}}  &  \multirow{1}{*}{\textbf{Training Accuracy}} & \multirow{1}{*}{\textbf{Testing Accuracy}} & \multirow{1}{*}{\textbf{p-value}} & \textbf{Training} & \textbf{Training} & \textbf{Testing Error} & \textbf{Testing Error} & \multirow{1}{*}{\textbf{\#Rules}} & \multirow{1}{*}{\textbf{n-vars}} & \multirow{1}{*}{\textbf{${\bf F_U}$/Rule}}  & \multirow{1}{*}{\textbf{Rule Length}}\\ 
\hline
\multirow{3}{*}{1} &
    \rebuttalii{NLDT} & $99.98 \pm 0.06 $ & ${\it 99.38 \pm 0.49} $ & {\it 0.158} & $0\pm 0$ & $0.04 \pm 0.12 $ & $0.37 \pm 0.76 $ & $0.88 \pm 0.90$ & ${\bf 1.0} \pm {\bf 0.0}$ & $2.5 \pm 0.6$ & $2.6 \pm 0.7$ & ${\bf 2.6 \pm 0.7}$\\ \cline{2-13}
  &  CART & $99.97 \pm 0.07 $ & $\mathbf{99.50 \pm 0.59} $ & -- & $0\pm 0$ & $0.06 \pm 0.13 $ & $0.23 \pm 0.84 $ & $0.77 \pm 0.77$ & $2.94 \pm 0.71$ & $1$ & ${\bf 1}$ & $2.94 \pm 0.71$\\ \cline{2-13}
  &  SVM & $99.37 \pm 0.17 $ & ${\it 99.40 \pm 0.40 }$ & {\it 0.331} & $0\pm 0$ & $1.26 \pm 0.33 $ & $0\pm 0$ & $1.20 \pm 0.79$ & ${\bf 1}$ & $4$ & $39.5 \pm 2.5$ & $39.5 \pm 2.5$\\\hline
 \multirow{3}{*}{10} &
    \rebuttalii{NLDT} & $99.39 \pm 0.38 $ & $98.58 \pm 1.13 $ & 3.42e-02 & $0.26 \pm 0.61 $ & $0.96 \pm 0.63 $ & $0.82 \pm 1.36 $ & $2.02 \pm 1.73$ & ${\bf 1.0} \pm {\bf 0.0}$ & $3.1 \pm 0.6$ & $3.9 \pm 1.0$ & ${\bf 3.9 \pm 1.0}$\\ \cline{2-13}
  &  CART & $99.46 \pm 0.27 $ & $97.72 \pm 1.04 $ & 4.63e-08 & $0.44 \pm 0.54 $ & $0.65 \pm 0.51 $ & $1.98 \pm 1.35 $ & $2.58 \pm 1.75$ & $8.42 \pm 1.42$ & $1$ & ${\bf 1}$ & $8.42 \pm 1.42$ \\ \cline{2-13}
  &  SVM & $99.46 \pm 0.19 $ & ${\bf 98.97 \pm 0.54 }$ & -- & $0.23 \pm 0.23 $ & $0.86 \pm 0.31 $ & $0.77 \pm 0.77 $ & $1.28 \pm 0.86$ & ${\bf 1}$ & $4$ & $126.0 \pm 6.8$ & $126.0 \pm 6.8$\\\hline
    \end{tabular}}
    \label{tab:welded_beam_gref_1_big}
\end{table*}

\subsection{ZDT Problems}
\begin{figure*}
    \centering
    \begin{subfigure}{0.28\textwidth}
        \centering
        \includegraphics[width = \textwidth]{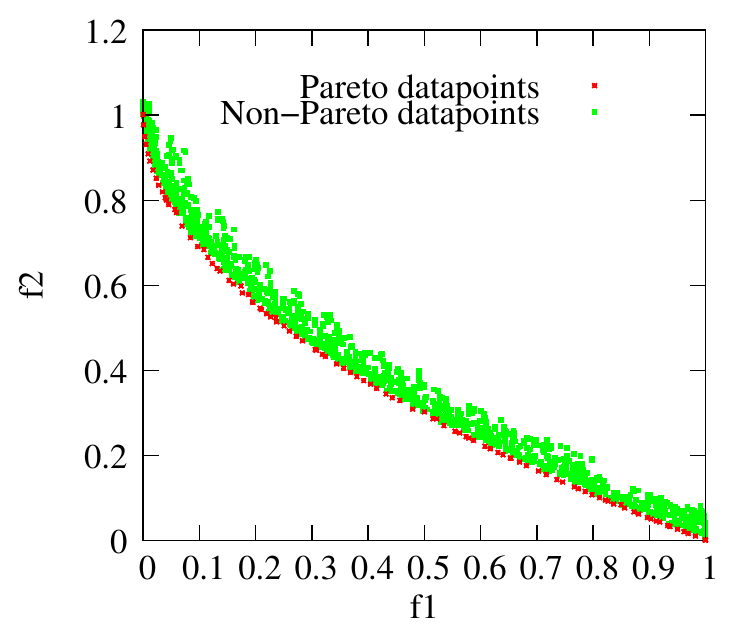}
        \caption{ZDT1.}
    \end{subfigure}
    \begin{subfigure}{0.28\textwidth}
        \centering
        \includegraphics[width = \textwidth]{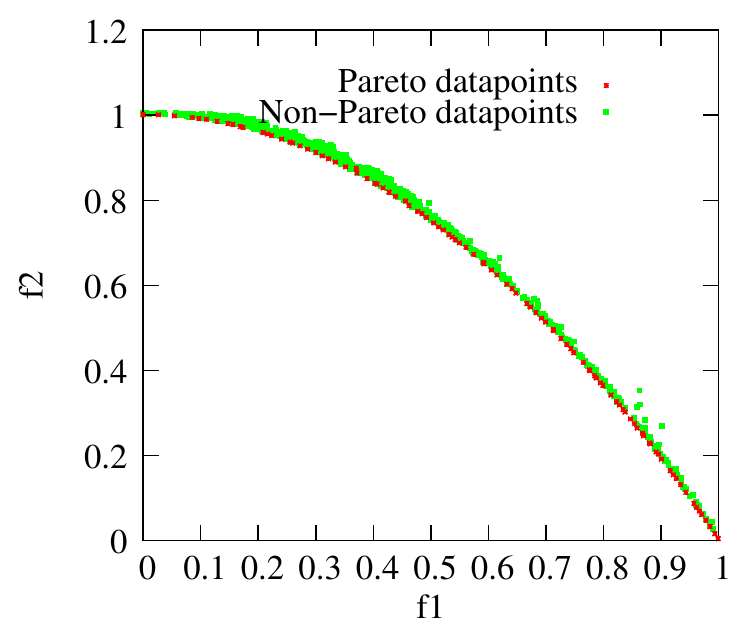}
        \caption{ZDT2.}
    \end{subfigure}
    \begin{subfigure}{0.28\textwidth}
        \centering
        \includegraphics[width = \textwidth]{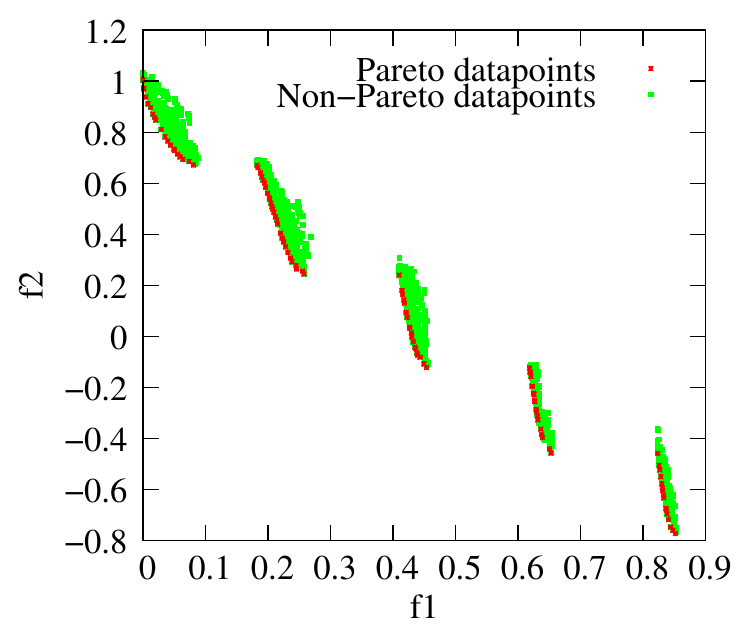}
        \caption{ZDT3.}
    \end{subfigure}
    \caption{There is no visible separation of the Pareto and non-Pareto datapoints, making the classification task difficult.}
    \label{fig:zdt_datasets}
\end{figure*}

ZDT problems are two-objective optimization problems and their formulations can be found in \cite{deb2002fast}. The dataset created was highly biased with $100$-datapoints in \emph{Pareto}-class and $5,000$-datapoints in \emph{non-Pareto-class}. Visualization of these datasets is provided in Figure~\ref{fig:zdt_datasets}. As can be seen from the figure \ref{fig:zdt_datasets}, \rebuttal{the classification task was deliberately made more difficult} by tightening the gap between \emph{Pareto} and \emph{non-Pareto class.}

Due to highly biased distribution, a naive classifier predicting all points as \emph{non-Pareto} would have the classification accuracy of $98.04\%$ ($=5000/(5000 + 100)$). Hence, the pruning implementation with $\tau_{prune} = 3\%$ will prune the entire tree, resulting into a tree with only one split-rule. Thus, we investigate the performance of the unpruned-NLDT and the prune-NLDT with $\tau_{prune} = 0.5\%$. The compilation of results on ZDT problems is provided in Table~\ref{tab:zdt_big}.
\begin{table*}[hbt]
\setcellgapes{2pt}
\makegapedcells
    \centering
    \caption{Results on multi-objective ZDT problems. Visualization of split-rules is provided in supplementary.}
    \begin{tabular}{|c|c|c|c|cHH|HHc|Hc|c|}\hline
    \multirow{2}{*}{\textbf{Problem}} & \multirow{2}{*}{\textbf{Method}}  &  \multirow{2}{*}{\textbf{Training Accuracy}} & \multirow{2}{*}{\textbf{Testing Accuracy}} & \multirow{2}{*}{\textbf{p-value}} & \textbf{Training} & \textbf{Training} & \textbf{Testing Error} & \textbf{Testing Error} & \multirow{2}{*}{\textbf{\#Rules}} & \multirow{2}{*}{\textbf{n-vars}} & \multirow{2}{*}{\textbf{${\bf F_U}$/Rule}}  & \multirow{2}{*}{\textbf{Rule Length}}\\
    & &  &  &  & {\bf Type 1} & {\bf Type 2} & {\bf Type 1} & {\bf Type 2} & & & & \\\hline
    \multirow{4}{*}{ZDT1} &
     \rebuttalii{NLDT} & $99.97 \pm 0.04 $ & ${\it 99.74 \pm 0.16 }$ & {\it 5.65e-01} & $1.34 \pm 1.81 $ & $0.00 \pm 0.01 $ & $8.97 \pm 6.59 $ & $0.08 \pm 0.09$ & $3.3 \pm 1.0$ & $3.5 \pm 0.8$ & $2.6 \pm 1.0$ & $7.9 \pm 1.9$\\\cline{2-13}

    & \rebuttalii{NLDT-pruned} & $99.80 \pm 0.05 $ & ${\it 99.66 \pm 0.18 }$ & {\it 9.36e-02} & $3.46 \pm 2.92 $ & $0.13 \pm 0.06 $ & $7.74 \pm 5.72 $ & $0.18 \pm 0.14$ & ${\bf 1.0} \pm {\bf 0.0}$ & $3.5 \pm 0.8$ & $3.6 \pm 0.9$ & ${\bf 3.6 \pm 0.9}$\\\cline{2-13}

    & CART & $99.95 \pm 0.04 $ & $99.59 \pm 0.15 $ & 1.10e-05 & $1.03 \pm 1.08 $ & $0.03 \pm 0.03 $ & $12.32 \pm 6.67 $ & $0.16 \pm 0.11$ & $9.0 \pm 2.1$ & $1$ & ${\bf 1}$ & $9.0 \pm 2.1$\\\cline{2-13}
    & SVM & $99.76 \pm 0.05 $ & ${\bf 99.72 \pm 0.11 }$ & -- & $4.94 \pm 1.59 $ & $0.14 \pm 0.04 $ & $7.87 \pm 3.92 $ & $0.12 \pm 0.09$ & ${\bf 1}$ & $30$ & $178.3 \pm 13.8$ & $178.3 \pm 13.8 $ \\\hline\hline
   
   \multirow{4}{*}{ZDT2} &  \rebuttalii{NLDT} & $99.86 \pm 0.08 $ & ${\it 99.20 \pm 0.23 }$ & {\it 9.41e-01} & $6.31 \pm 4.10 $ & $0.02 \pm 0.02 $ & $27.35 \pm 9.45 $ & $0.25 \pm 0.16$ & $5.7 \pm 1.7$ & $15.0 \pm 4.0$ & $6.2 \pm 1.5$ & $33.7 \pm 5.4$\\\cline{2-13}
    & \rebuttalii{NLDT-pruned} & $99.54 \pm 0.13 $ & $99.08 \pm 0.27 $ & 1.40e-02 & $11.17 \pm 7.00 $ & $0.25 \pm 0.16 $ & $23.94 \pm 10.63 $ & $0.44 \pm 0.24$ & $1.6 \pm 1.0$ & $15.0 \pm 4.0$ & $15.9 \pm 5.7$ & $21.6 \pm 4.3$\\\cline{2-13}

    & CART & $99.85 \pm 0.07 $ & $99.04 \pm 0.27 $ & 2.58e-04 & $5.43 \pm 3.41 $ & $0.05 \pm 0.04 $ & $32.00 \pm 13.51 $ & $0.32 \pm 0.17$ & $21.2 \pm 2.4$ & $1$ & ${\bf 1}$ & ${\bf 21.2 \pm 2.4}$\\\cline{2-13}
    & SVM & $99.29 \pm 0.07 $ & ${\bf 99.21 \pm 0.16 }$ & -- & $32.69 \pm 4.51 $ & $0.07 \pm 0.04 $ & $34.58 \pm 7.49 $ & $0.09 \pm 0.06$ & ${\bf 1}$ & $30$ & $141.9 \pm 32.7$ & $141.9 \pm 32.7 $ \\\hline\hline

    \multirow{4}{*}{ZDT3} & \rebuttalii{NLDT} & $99.95 \pm 0.06 $ & $99.71 \pm 0.17 $ & 9.71e-06 & $2.60 \pm 3.08 $ & $0.00 \pm 0.00 $ & $12.26 \pm 7.68 $ & $0.05 \pm 0.07$ & $4.4 \pm 1.2$ & $8.8 \pm 2.5$ & $3.8 \pm 1.8$ & $15.5 \pm 4.6$\\\cline{2-13}

    & \rebuttalii{NLDT-pruned} & $99.75 \pm 0.08 $ & ${\bf 99.60 \pm 0.19 }$ & -- & $8.40 \pm 4.24 $ & $0.08 \pm 0.08 $ & $13.61 \pm 8.63 $ & $0.13 \pm 0.14$ & $1.1 \pm 0.3$ & $8.8 \pm 2.5$ & $9.4 \pm 3.3$ & ${\bf 9.7 \pm 3.5}$\\\cline{2-13}

    & CART & $99.90 \pm 0.06 $ & $99.47 \pm 0.20 $ & 2.30e-03 & $2.83 \pm 2.15 $ & $0.05 \pm 0.05 $ & $17.29 \pm 8.08 $ & $0.19 \pm 0.13$ & $13.5 \pm 2.2$ & $1$ & ${\bf 1}$ & $13.5 \pm 2.2$\\\cline{2-13}

    & SVM & $98.18 \pm 0.19 $ & $98.02 \pm 0.11 $ & 7.28e-10 & $92.94 \pm 9.77 $ & $0.00 \pm 0.00 $ & $97.61 \pm 5.52 $ & $0.00 \pm 0.00$ & ${\bf 1}$ & $30$ & $286.7 \pm 33.2$ & $286.7 \pm 33.2 $ \\\hline

    \end{tabular}
    \label{tab:zdt_big}
\end{table*}
Clearly, the pruned NLDT requires one or two rules with fewer variables (or features) per rule than CART (on an average 3.6 versus 9.0 on ZDT1 and 9.7 versus 286.7 on ZDT3), but achieves a similar classification accuracy. The difficulty with CART solutions is that they are too deep to be easily interpretable. Although theoretical Pareto-optimal points for ZDT problems can be identified by checking $x_i=0$ for $i\geq 2$, NSGA-II, being a stochastic optimization algorithm, is not expected to find the exact Pareto-optimal solutions, thereby making classification methods to find a classifier with more than one participating variables.

\begin{figure*}[hbt]
    \centering
    \includegraphics[width = 0.5\linewidth]{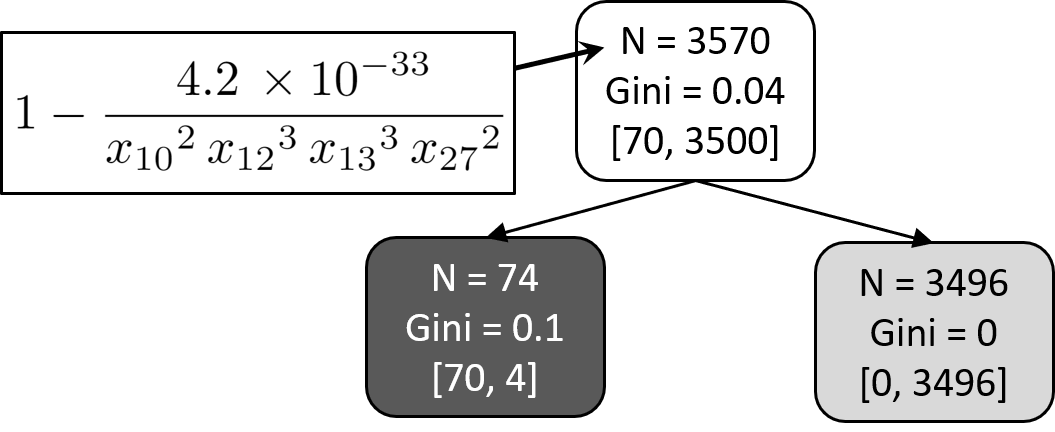}
    \caption{NLDT for ZDT-1. It has the testing accuracy of $99.93\%$ and involves $4$ variables in the split-rule.}
    \label{fig:zdt1_pruned_tree}
\end{figure*}

\begin{figure*}[hbt]
    \centering
    \includegraphics[width = \linewidth]{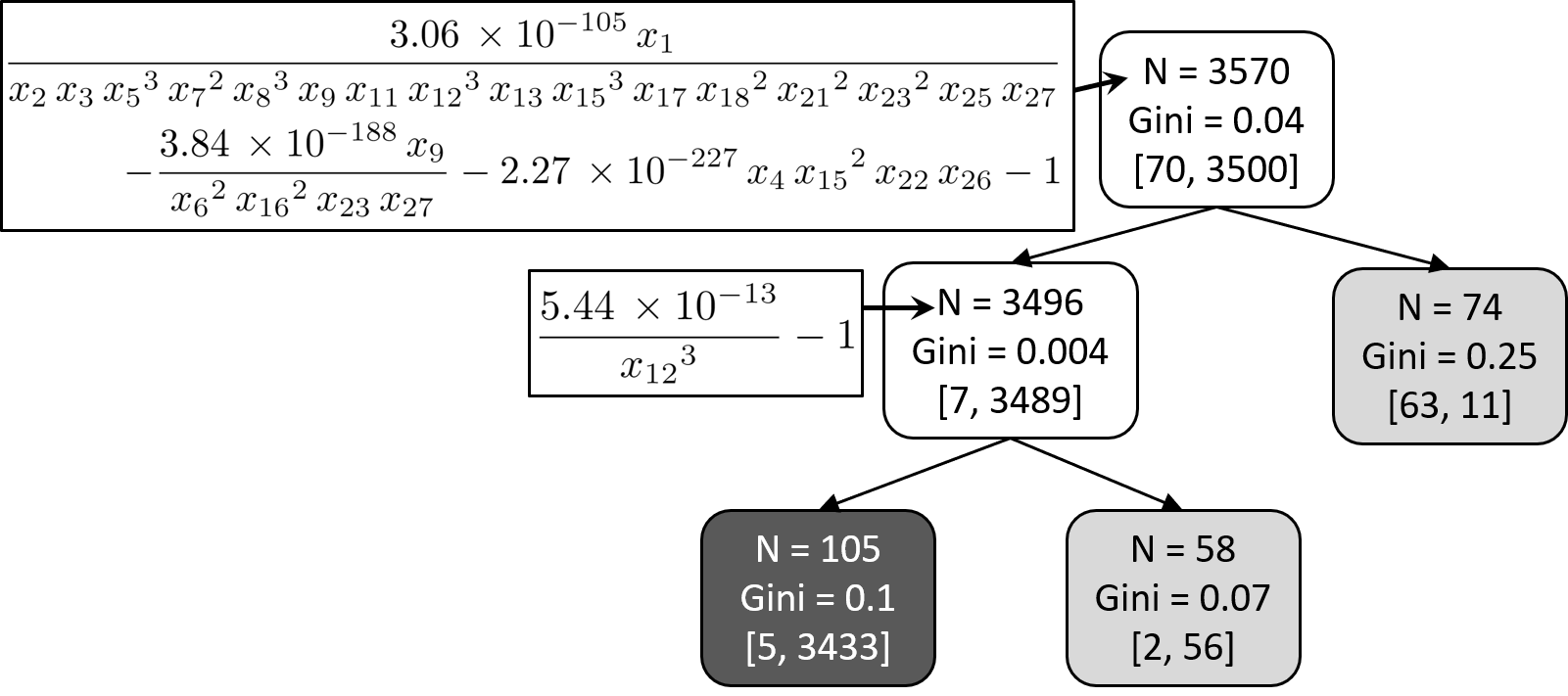}
    \caption{NLDT for ZDT-2. It has the testing accuracy of $99.54\%$ and involves $22$ variables in equation of the split-rule at the root node.}
    \label{fig:zdt2_pruned_tree}
\end{figure*}

\begin{figure*}[hbt]
    \centering
    \includegraphics[width = 0.7\linewidth]{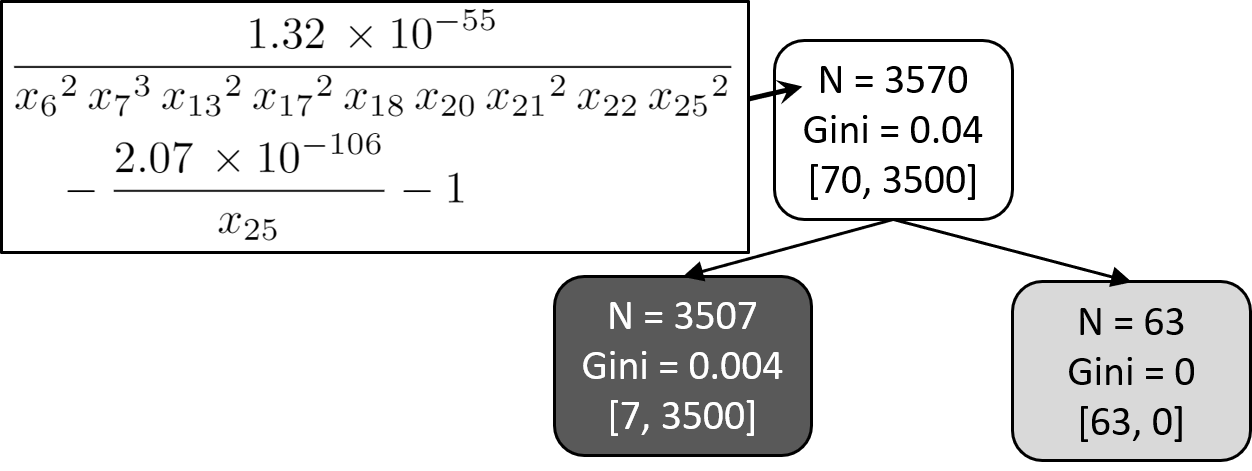}
    \caption{NLDT for ZDT-3. It has the testing accuracy of $100\%$ and involves $9$ variables in equation of the split-rule.}
    \label{fig:zdt3_pruned_tree}
\end{figure*}

Visualization of NLDT obtained using our bilevel algorithm on ZDT problems is provided in Figures \ref{fig:zdt1_pruned_tree}, \ref{fig:zdt2_pruned_tree} and \ref{fig:zdt3_pruned_tree}.

\subsection{\rebuttalii{m-ZDT and m-DTLZ problems}}
\begin{figure*}[hbt]
\centering
\begin{subfigure}[b]{0.49\textwidth}
\includegraphics[width = \textwidth]{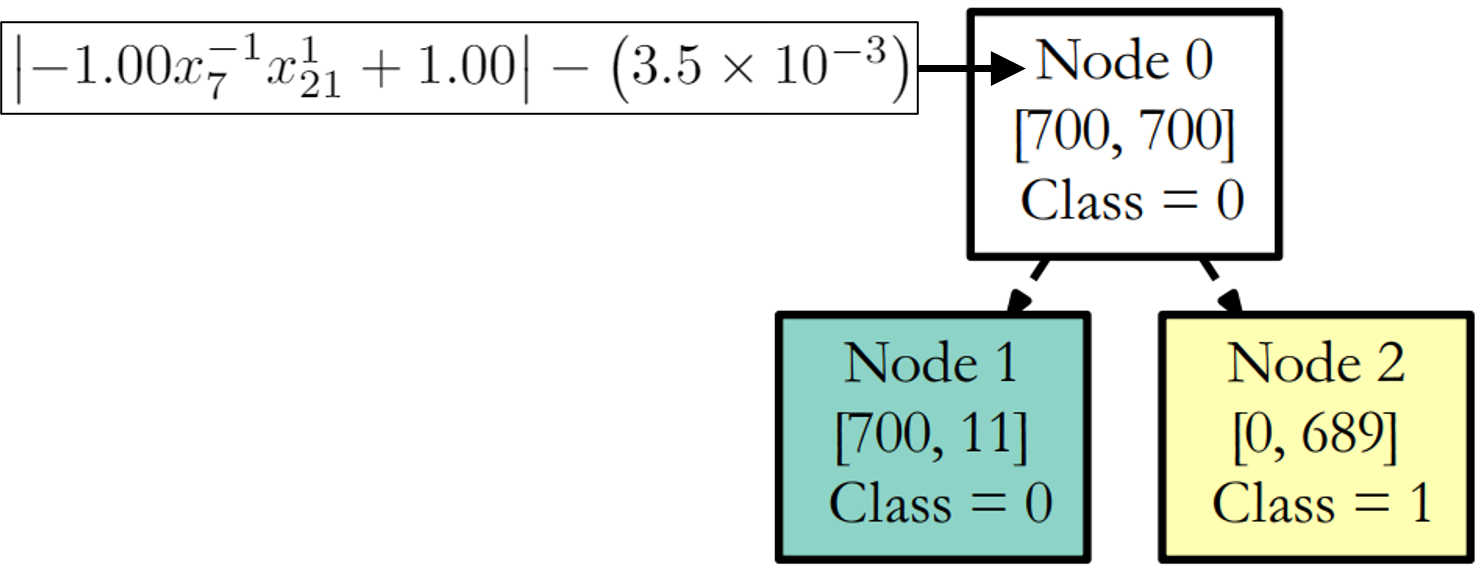}
\caption{\rebuttalii{m-ZDT1 30 vars}}
\end{subfigure}
~
\begin{subfigure}[b]{0.49\textwidth}
\includegraphics[width = \textwidth]{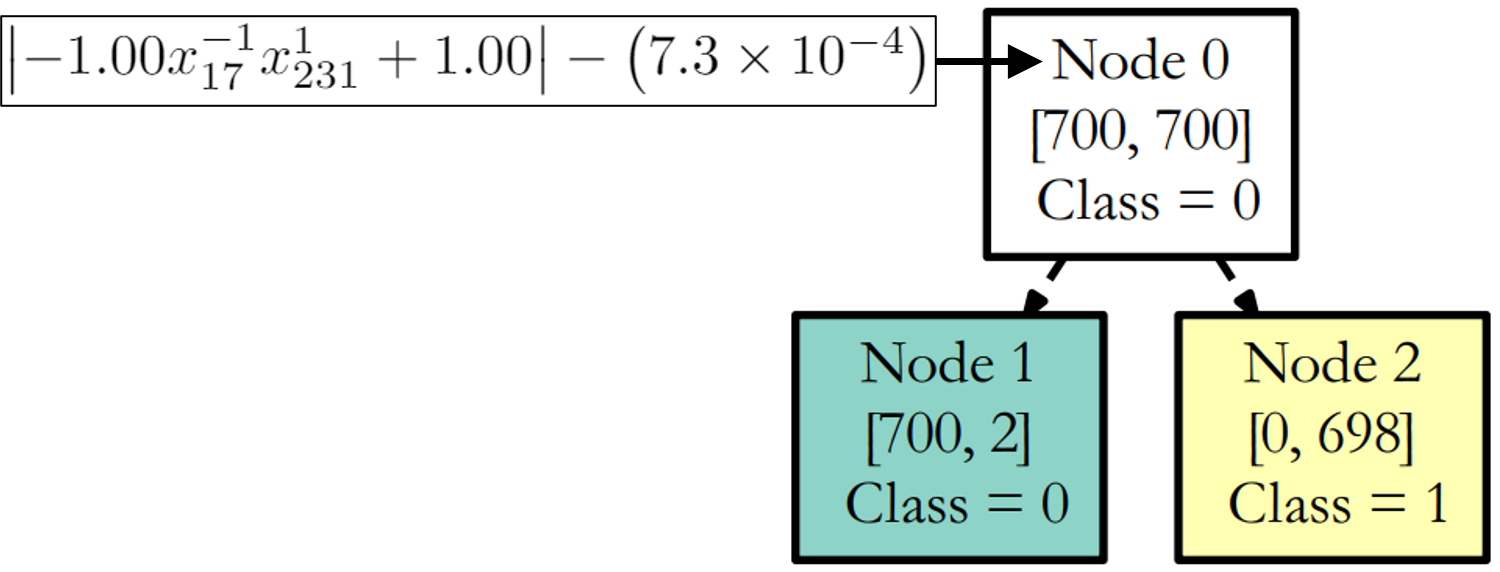}
\caption{\rebuttalii{m-ZDT1 500 vars}}
\end{subfigure}
\\
\begin{subfigure}[b]{0.6\textwidth}
\includegraphics[width = \textwidth]{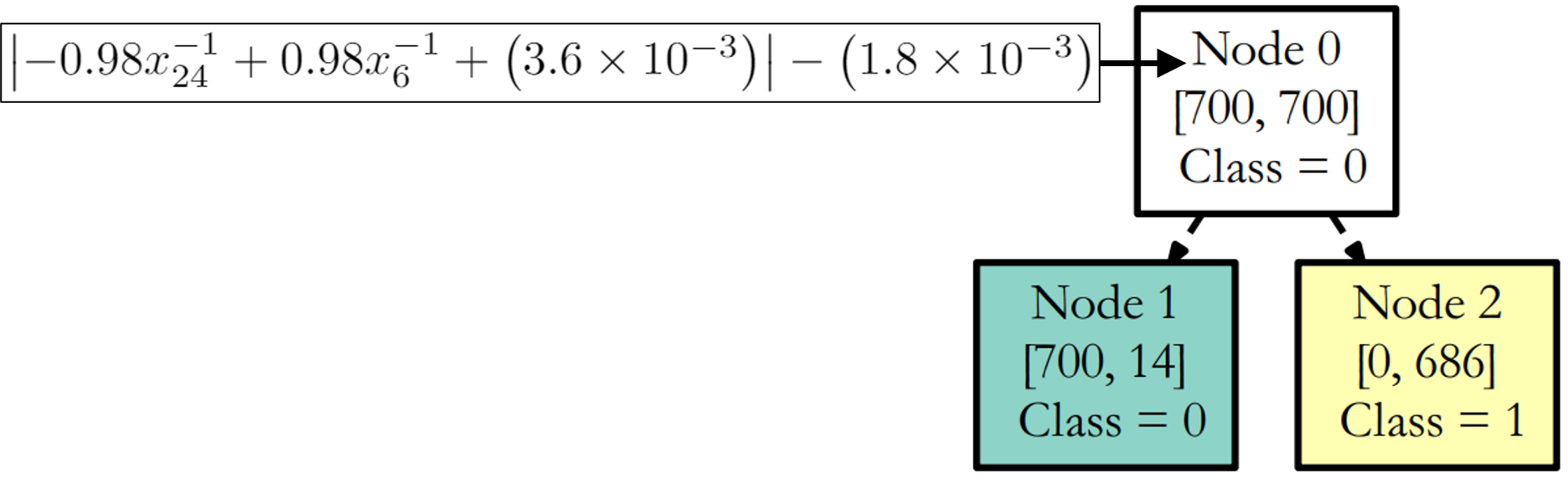}
\caption{\rebuttalii{m-ZDT2 30 vars}}
\end{subfigure}
~
\begin{subfigure}[b]{0.49\textwidth}
\includegraphics[width = \textwidth]{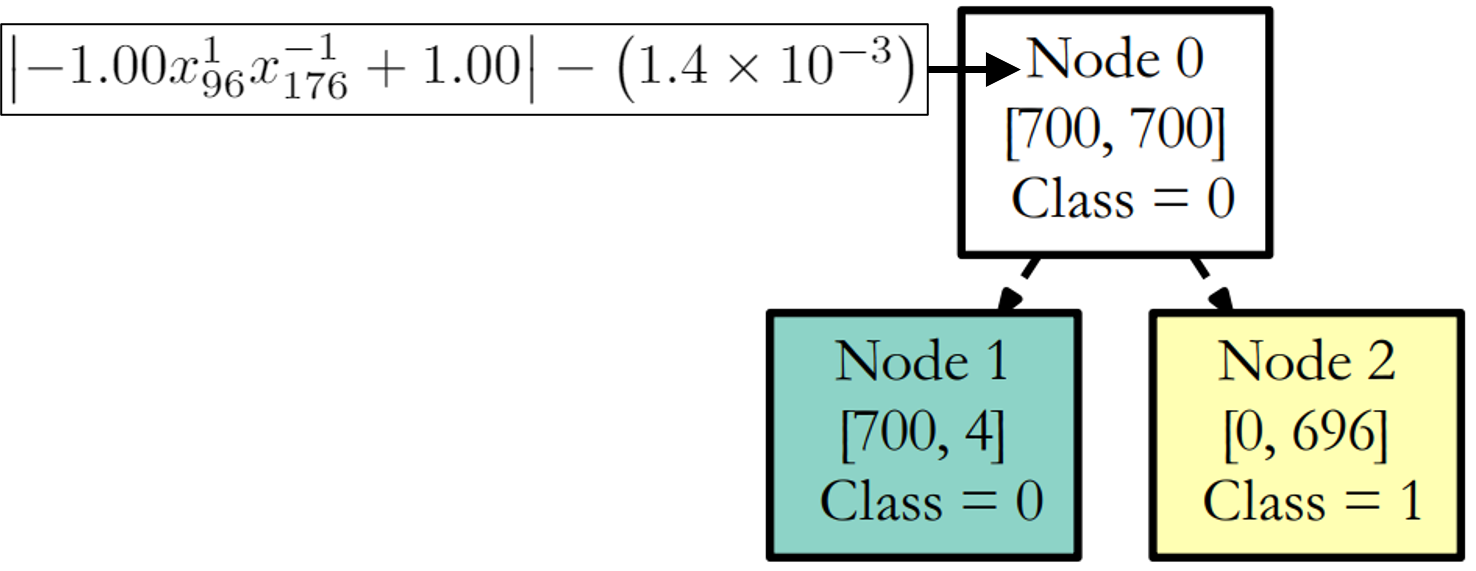}
\caption{\rebuttalii{m-ZDT2 500 vars}}
\end{subfigure}
\caption{\rebuttalii{NLDTs for m-ZDT problems. Class=0 indicates the Pareto dataset. Even for 500-var problems, only a two-variable interaction is enough to make a highly-accurate classification.}}
\label{fig:m_zdt_trees}
\end{figure*}

\begin{figure*}[hbt]
\centering
\begin{subfigure}[b]{0.53\textwidth}
\includegraphics[width = \textwidth]{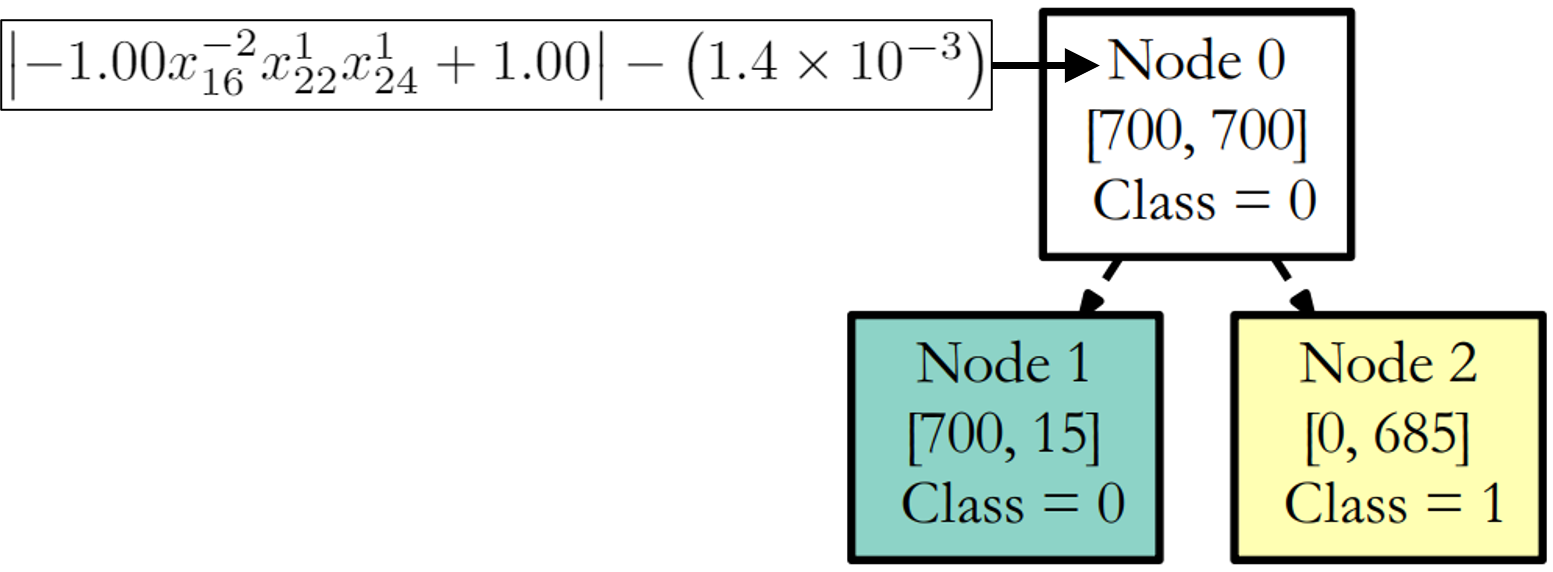}
\caption{\rebuttalii{m-DTLZ1 30 vars}}
\end{subfigure}
~
\begin{subfigure}[b]{0.63\textwidth}
\includegraphics[width = \textwidth]{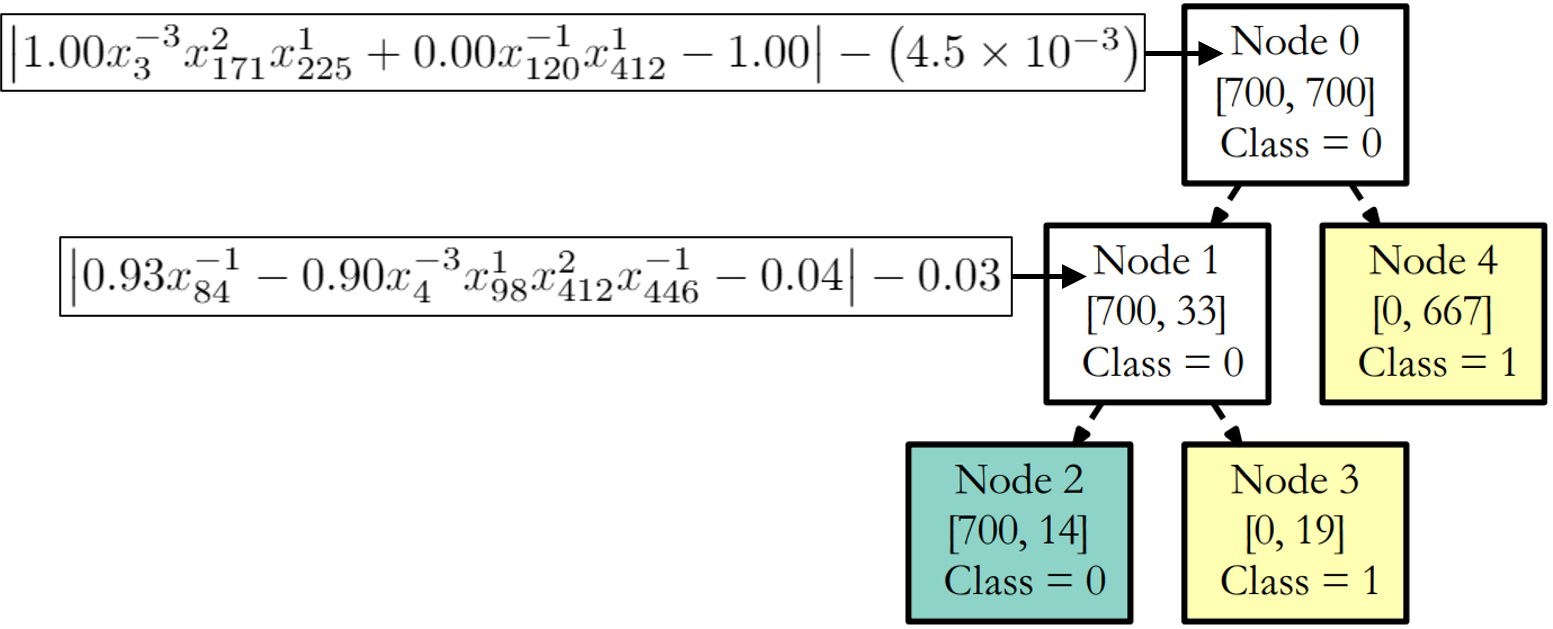}
\caption{\rebuttalii{m-DTLZ1 500 vars}}
\end{subfigure}
\\
\begin{subfigure}[b]{0.75\textwidth}
\includegraphics[width = \textwidth]{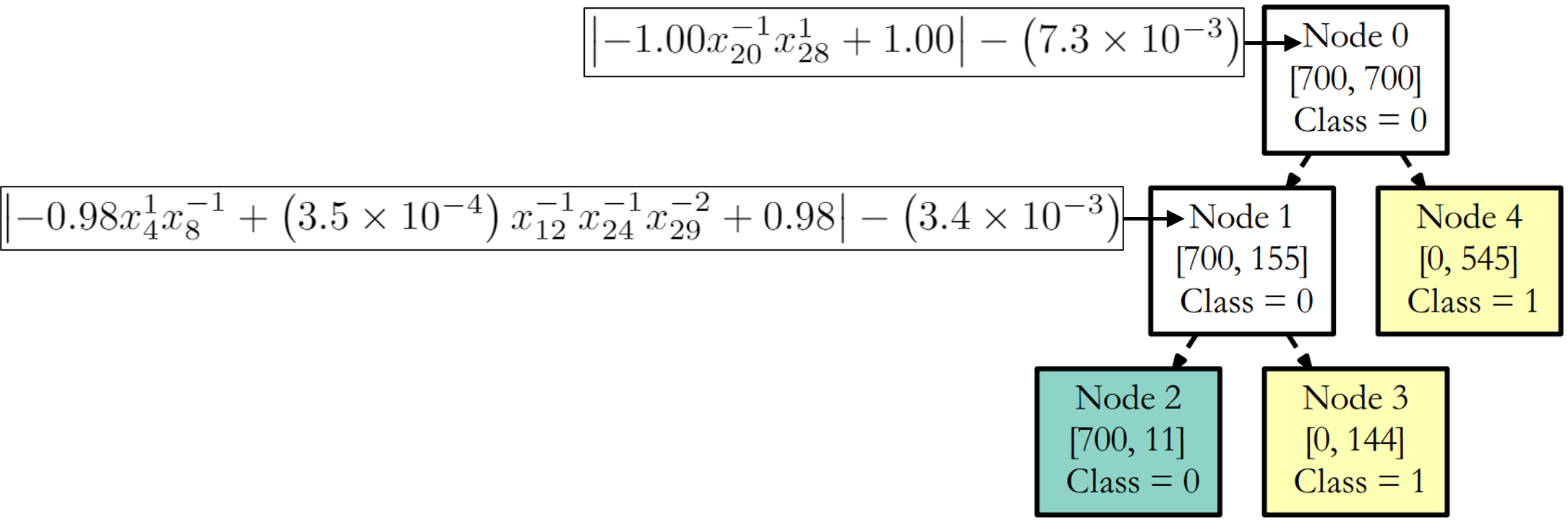}
\caption{\rebuttalii{m-DTLZ2 30 vars}}
\end{subfigure}
~
\begin{subfigure}[b]{0.6\textwidth}
\includegraphics[width = \textwidth]{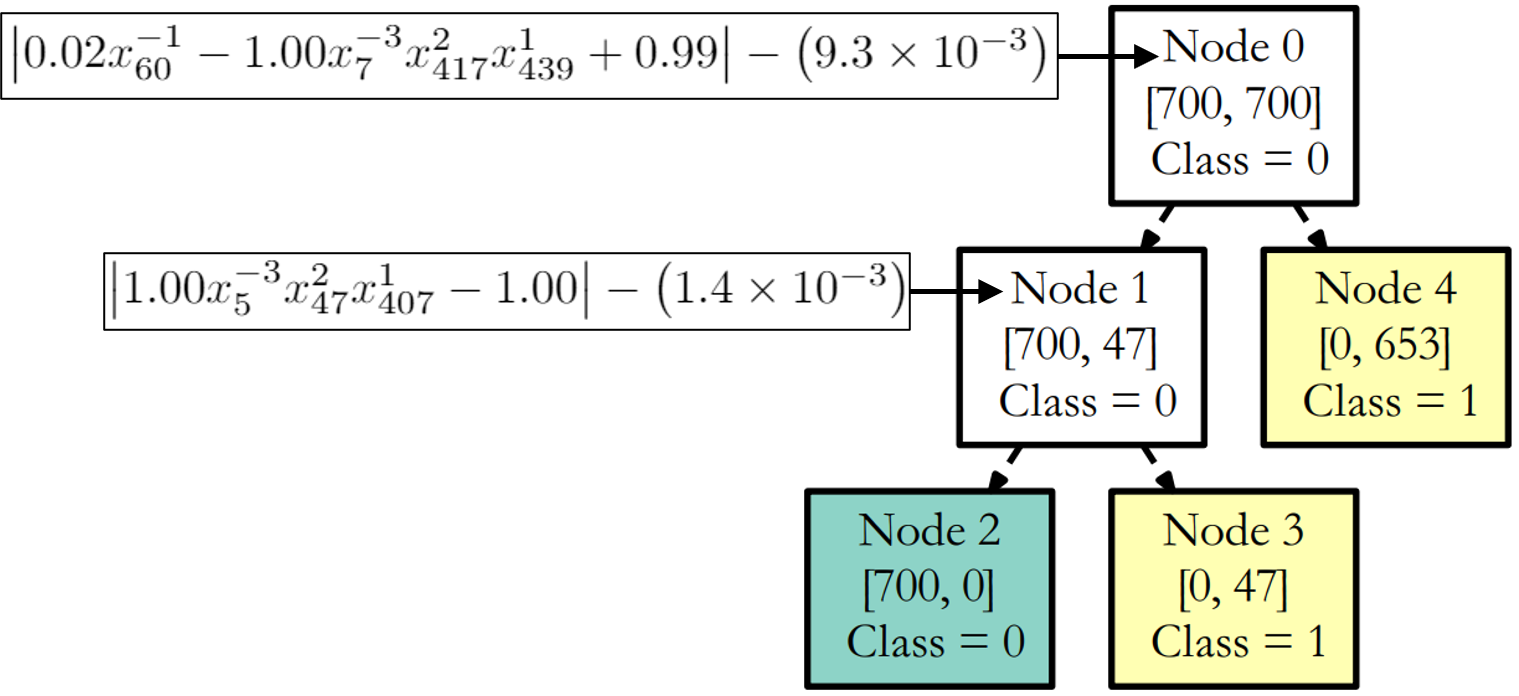}
\caption{\rebuttalii{m-DTLZ2 500 vars}}
\end{subfigure}
\caption{\rebuttalii{NLDTs for m-DTLZ problems. Class=0 indicates the Pareto dataset.}}
\label{fig:m_dtlz_trees}
\end{figure*}

\rebuttalii{Visualization of some NLDTs obtained for m-ZDT and m-DTLZ problems are shown in Figure~\ref{fig:m_zdt_trees} and \ref{fig:m_dtlz_trees}, respectively. For m-ZDT problems, a relationship with two consecutive variables would constitute a relationship present in the Pareto set, but our approach finds a different two-variable interaction which has been found to be present in the dataset and was enough to classify Pareto versus non-Pareto data. For m-DTLZ problems, the relationships are a little more complex. Interestingly, for 500-var problem, only seven variables are needed in a nonlinear fashion to classify two classes of data.}

\section{Influence of Parameter $\tau_I$}
\begin{figure*}[hbt]
    \centering
    \includegraphics[width = \linewidth]{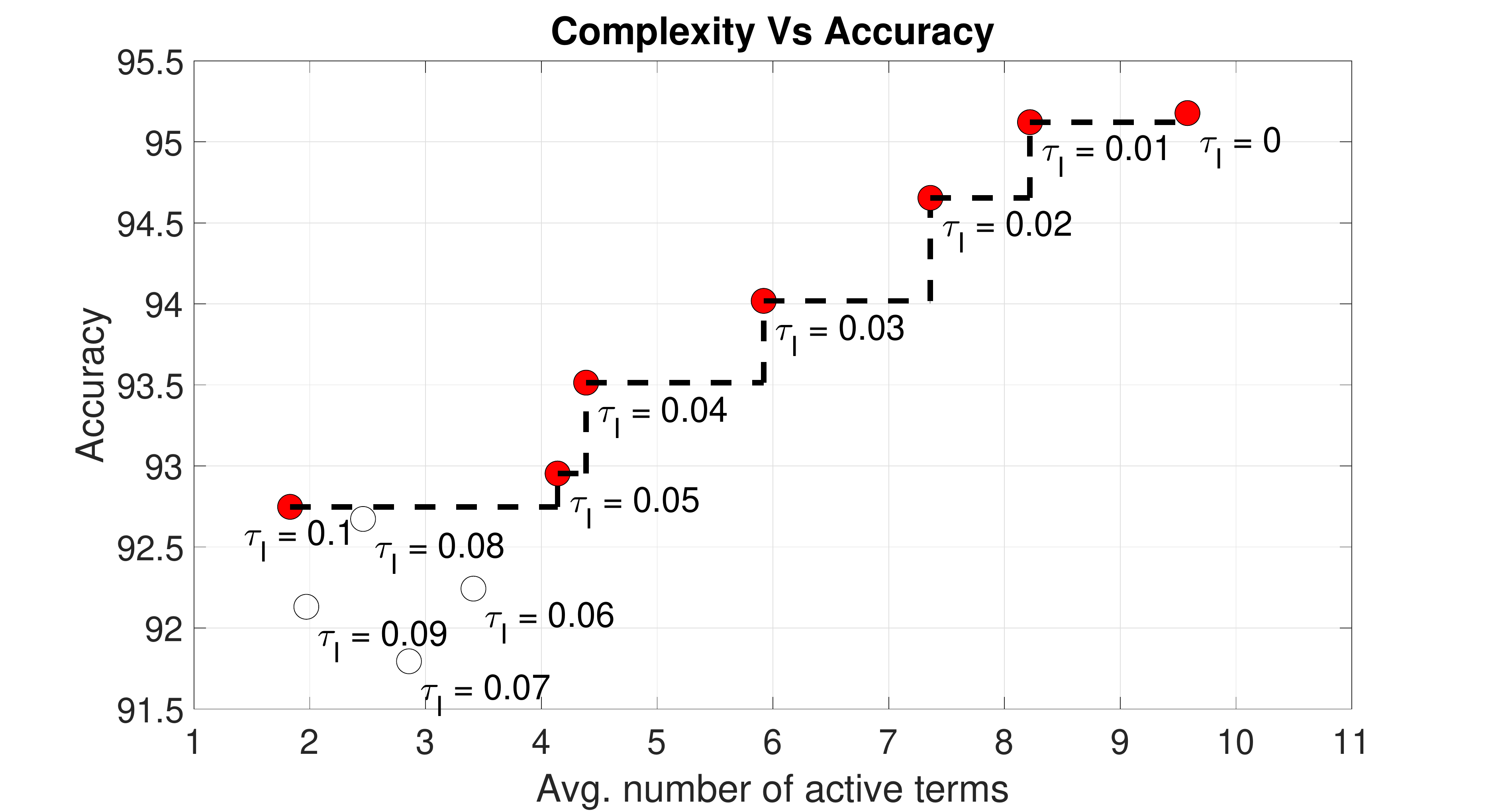}
    \caption{\rebuttal{The effect of various values of $\tau_I$ on the complexity (horizontal axis) and accurary of classifier (vertical axis) is shown here for 36 dimensional two-class real-world auto-industry problem. The value of $\tau_I$ in the range from 0.03 to 0.05 seems to be reliable for this two-class classification problem.}}
    \label{fig:tauI_complexity_accuracy}
\end{figure*}

\rebuttal{In this section, we study the influence of parameter $\tau_I$ on the complexity of the split-rule and the accuracy of overall decision tree. As can be seen from Figure~\ref{fig:tauI_complexity_accuracy}, value of $\tau_I = 0.05$ seemed to be reasonable enough for a two-class classification dataset.}

\section{\rebuttalii{Multi-Class Iris Dataset Results}}
\rebuttalii{Results obtained on three class \emph{iris} dataset using NLDT, CART and SVM are shown in Table~\ref{tab:iris_dataset_result}. From the results it is clear that the proposed approach can be extended to multi-class problems as well. In our future study, we will extend our work to multi-class problems with more complicated decision boundaries than the standard \emph{iris} dataset.}
\begin{table*}[hbtp]
\setcellgapes{2pt}
\makegapedcells
    \centering
    \caption{\rebuttalii{Results on multi-class iris dataset. NLDT requires fewer rules and achieves more accuracy than CART and NLDT requires much smaller rule length (number of features in all rules) than SVM, making an excellent compromise of the two existing approaches.}}
    {\color{black}
    \begin{tabular}{|c|c|c|H|c|c|c|}\hline
    \textbf{Method} & \textbf{Training Acc} & \textbf{Testing Acc.} & \textbf{p-value} & \textbf{\# Rules} & \textbf{$F_U$/Rule} & \textbf{Rule Length}\\\hline
    NLDT & {\bf 98.59 $\pm$  0.69} & $ 94.80 \pm 4.14 $ & NA & $ 2.00 \pm 0.00 $ & $ 1.96 \pm 0.57 $ & $ 3.92 \pm 1.15 $ \\\hline
    CART & $ 96.57 \pm  1.03 $ & $ 93.69 \pm 2.75 $ & NA & $ 3.80 \pm 0.49 $ & {\bf 1.00 $\pm$ 0.00} & {\bf 3.80 $\pm$ 0.49} \\\hline
    SVM & $ 96.76 \pm  1.17 $ & {\bf 96.13 $\pm$ 2.51} & NA & {\bf 1.00 $\pm$ 0.00} & $ 47.94 \pm 2.17 $ & $ 47.94 \pm 2.17 $ \\\hline
    \end{tabular}}
    \label{tab:iris_dataset_result}
\end{table*}

\section{\rebuttalii{Summary}}
\rebuttalii{Results of the main paper and additional results presented in this document amply indicate that the proposed bilevel-based NLDT approach is a viable strategy for classification tasks with fewer rules than a standard decision tree (CART, for example) produces and is more simplistic than a single complex rule that a SVM procedure produces. Hence, our nonlinear rules with restricted complexity are more humanly interpretable and useful not only for a mere classification task, but also for a better understanding of the classifier.}


\end{document}